\documentclass{article} %
\usepackage{iclr2026_conference,times}

\usepackage{amsmath,amsfonts,bm}
\def\eqref#1{equation~\ref{#1}}

\def\1{\bm{1}}

\DeclareMathAlphabet{\mathsfit}{\encodingdefault}{\sfdefault}{m}{sl}
\SetMathAlphabet{\mathsfit}{bold}{\encodingdefault}{\sfdefault}{bx}{n}

\definecolor{lightblue1}{rgb}{0.97, 0.985, 1} 
\definecolor{lightblue2}{rgb}{0.92, 0.965, 1} 
\definecolor{lightblue3}{rgb}{0.84, 0.93, 1}
\definecolor{lightblue4}{rgb}{0.74, 0.87, 1}
\definecolor{lightblue5}{rgb}{0.64, 0.81, 1}
\definecolor{lightblue6}{rgb}{0.54, 0.75, 1}

\definecolor{lightgreen1}{rgb}{0.97, 1.00, 0.97}
\definecolor{lightgreen2}{rgb}{0.92, 0.98, 0.92}
\definecolor{lightgreen3}{rgb}{0.84, 0.95, 0.84}
\definecolor{lightgreen4}{rgb}{0.74, 0.91, 0.74}
\definecolor{lightgreen5}{rgb}{0.64, 0.86, 0.64}
\definecolor{lightgreen6}{rgb}{0.54, 0.81, 0.54}

\definecolor{lightorange1}{rgb}{1.00, 0.98, 0.95}
\definecolor{lightorange2}{rgb}{1.00, 0.95, 0.85}
\definecolor{lightorange3}{rgb}{1.00, 0.90, 0.70}
\definecolor{lightorange4}{rgb}{1.00, 0.85, 0.55}
\definecolor{lightorange5}{rgb}{1.00, 0.80, 0.40}
\definecolor{lightorange6}{rgb}{1.00, 0.75, 0.30}

\definecolor{lightpurple1}{rgb}{0.985, 0.97, 1.00}
\definecolor{lightpurple2}{rgb}{0.96, 0.92, 1.00}
\definecolor{lightpurple3}{rgb}{0.93, 0.84, 1.00}
\definecolor{lightpurple4}{rgb}{0.87, 0.74, 1.00}
\definecolor{lightpurple5}{rgb}{0.81, 0.64, 1.00}
\definecolor{lightpurple6}{rgb}{0.75, 0.54, 1.00}

\definecolor{lightred1}{rgb}{1.00, 0.97, 0.97}
\definecolor{lightred2}{rgb}{1.00, 0.92, 0.92}
\definecolor{lightred3}{rgb}{1.00, 0.84, 0.84}
\definecolor{lightred4}{rgb}{1.00, 0.74, 0.74}
\definecolor{lightred5}{rgb}{1.00, 0.64, 0.64}
\definecolor{lightred6}{rgb}{1.00, 0.54, 0.54}

\definecolor{lightcyan1}{rgb}{0.97, 1.00, 1.00}
\definecolor{lightcyan2}{rgb}{0.92, 0.98, 0.98}
\definecolor{lightcyan3}{rgb}{0.84, 0.95, 0.96}
\definecolor{lightcyan4}{rgb}{0.74, 0.91, 0.94}
\definecolor{lightcyan5}{rgb}{0.64, 0.87, 0.92}
\definecolor{lightcyan6}{rgb}{0.54, 0.83, 0.90}

\usepackage{xcolor,colortbl}
\definecolor{Gray}{gray}{0.85}
\definecolor{LightCyan}{rgb}{0.88,1,1}
\definecolor{greyC}{RGB}{180,180,180}
\definecolor{greyL}{RGB}{235,235,235}
\definecolor{citeColor}{RGB}{0,20,115}
\usepackage{hyperref}
\hypersetup{colorlinks,linkcolor={red},citecolor={citeColor},urlcolor={citeColor}}
\definecolor{shadecolor}{rgb}{0.92,0.92,0.92}

\usepackage{graphicx}
\usepackage{xspace}
\usepackage{subcaption}
\usepackage{hyperref}
\usepackage{url}
\newcommand{\method}{TFPI}
\newcommand{\thinkfree}{\textit{ThinkingFree}}
\usepackage{booktabs}
\usepackage{multirow}
\usepackage{cleveref}
\crefname{template}{Template}{Template}
\newtheorem{template}{Template}

\usepackage{enumitem} 
\usepackage[most]{tcolorbox}
\usepackage{pifont}
\usepackage{ulem}
\usepackage{authblk}
\definecolor{rliableblue}{RGB}{0, 102, 204} %
\usepackage{xcolor}

\tcbset{
    myblueboxstyle/.style={
        colback=lightblue3, %
        colframe=lightblue5, %
        coltitle=black, %
        fonttitle=\bfseries, %
        boxrule=1pt, %
        width=\linewidth, %
        left=2pt, %
        right=2pt, %
        top=2pt, %
        bottom=2pt, %
        sharp corners, %
        boxsep=2pt
    }
}

\newcommand{\github}{\raisebox{-1.5pt}{\includegraphics[height=1.05em]{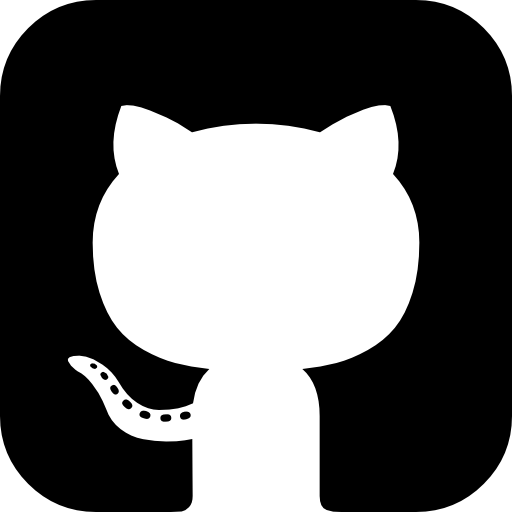}}\xspace}
\newcommand{\dataset}{\raisebox{-1.5pt}{\includegraphics[height=1.05em]{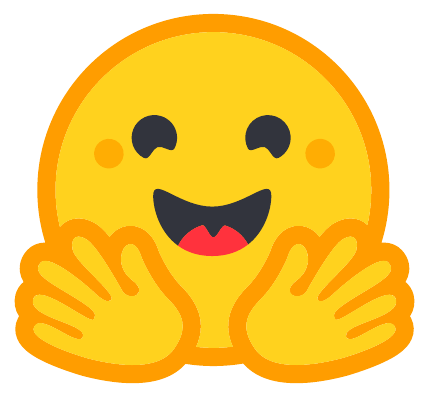}}\xspace}
\NewDocumentCommand{\xx}
{ mO{} }{\textcolor{blue}{\textsuperscript{\textit{todo}}\textsf{\textbf{\small[#1]}}}}
\newcommand{\revise}[1]{\textcolor{black}{#1}}
\title{Thinking-Free Policy Initialization Makes Distilled Reasoning Models More Effective and Efficient Reasoners}

\author{
\small{
\textbf{Xin Xu}$^{1,2}$, 
\textbf{Clive Bai}$^{2}$, 
\textbf{Kai Yang}$^{2}$, 
\textbf{Tianhao Chen}$^{1}$, 
\textbf{Yang Wang}$^{5}$, 
\textbf{Saiyong Yang}$^{2,\dagger}$, 
\textbf{Can Yang}$^{1,3,4,\dagger}$\\
$^1$Department of Mathematics, HKUST
$^2$HY, Tencent \\
$^3$Big Data Bio-Intelligence Lab, HKUST 
$^4$State Key Laboratory of Nervous System Disorders, HKUST \\
$^5$Department of Mathematics, HKU \\
\github \textbf{Github Repo}: \href{https://github.com/Tencent-Hunyuan/Thinking-Free_Policy_Initialization}{[GitHub Page]} 
\dataset \textbf{Models}: \href{https://huggingface.co/collections/xx18/tfpi}{[Huggingface Models]}
}
}

\iclrfinalcopy %
\begin{document}

\maketitle
\renewcommand*{\thefootnote}{\fnsymbol{footnote}}
\footnotetext{$\dagger$ Corresponding Authors.}
\begin{abstract}

Reinforcement Learning with Verifiable Reward (RLVR) effectively solves complex tasks but demands extremely long context lengths during training, leading to substantial computational costs.
While multi-stage training can partially mitigate this, starting with overly short contexts often causes irreversible performance degradation, ultimately failing to reduce overall training compute significantly.
In this paper, we introduce \textbf{T}hinking-\textbf{F}ree \textbf{P}olicy \textbf{I}nitialization (\textbf{{\method}}), a simple yet effective adaptation to RLVR that bridges long Chain-of-Thought (CoT) distillation and standard RLVR.  
{\method} employs a simple \thinkfree{} operation, explicitly discarding the thinking content via a direct \texttt{\textless/think\textgreater} append, to reduce token usage during inference.  
Training with \thinkfree{}-adapted inputs improves performance and lowers token consumption, even in the original slow-thinking mode.
Extensive experiments across various benchmarks have shown that {\method} accelerates RL convergence, achieves a higher performance ceiling, and yields more token-efficient reasoning models without specialized rewards or complex training designs.
With {\method} only, we train a 4B model to reach 89.0\% accuracy on AIME24 and 65.5\% on LiveCodeBench using less than 4K H20 hours.

\end{abstract}

\section{Introduction}

Reasoning is a fundamental aspect of human cognition, and equipping artificial intelligence (AI) with strong reasoning capabilities is critical for its deployment and applications~\citep{morris2023levels, huang2024olympicarena, xu2025ugmathbench}.
Progress in pretraining~\citep{shao2024grpo,yang2024qwen25math,chen2025gpas}, supervised fine-tuning (SFT)~\citep{metamath2023yu,xu2024egsm,dartmath2024tong, ye2025limo, muennighoff2025s1}, rigorous evaluation~\citep{rein2024gpqa,phan2025humanity,xu2025ugphysics}, reinforcement learning (RL)~\citep{jaech2024openaio1, guo2025deepseekr1} has significantly enhanced the reasoning abilities of large language models (LLMs).
Among these, RL with verifiable rewards (RLVR) stands out as an effective approach that enables large language models (LLMs) to generate long Chains-of-Thought (CoT)~\citep{CoT2022Wei} spontaneously, and empowers them with unprecedented performance on challenging reasoning tasks.
Thus, these RLVR-trained LLMs are termed as long-CoT LLMs, slow-thinking LLMs, or large reasoning models (LRMs).

Compared with initializing from a base LLM, starting from an SFT-distilled LRM typically yields better results and accelerates convergence in RLVR~\citep{guo2025deepseekr1,Polaris2025}.
However, SFT-distilled LRMs often produce excessively long responses during the rollout stage of RLVR, which necessitates a large training context length for RLVR.
Using such large training contexts also incurs substantial computational costs.
A common mitigation strategy is multistage RLVR, which begins with a relatively ``short'' context and gradually increases the training length~\citep{deepscaler2025,Polaris2025}.
Nonetheless, \citet{zeng2025glm4_5} argue that multistage training might cause irreversible performance degradation.
Moreover, even multistage training demands significant computational resources.
For instance, training a 4B model while progressively increasing the maximum context length from 40K to 48K to 52K tokens requires approximately 8K H800 GPU hours~\citep{Polaris2025}.
These limitations underscore the need for more efficient RLVR training methods.

\begin{figure}
    \centering
    \includegraphics[width=0.9\linewidth]{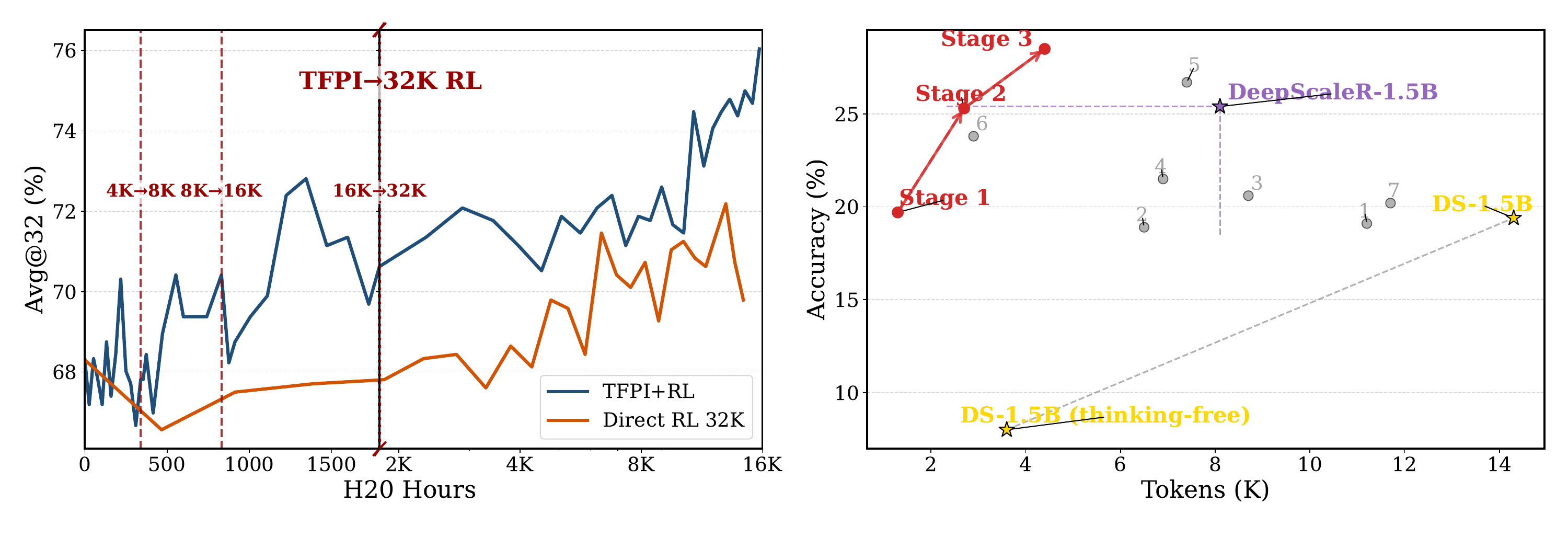}
    \vspace{-1em}
    \caption{Our proposed {\method} accelerates the convergence of RLVR to a higher performance ceiling (left) and yields more token-efficient reasoning models (right).  
    \textbf{Left:} \texttt{avg@32} versus training compute, measured in H20 hours. 
    ``Direct RL'' refers to directly training \texttt{Qwen3-4B} with a 32K context window using DAPO, while ``{\method} + RL'' denotes running 32K-context DAPO after initialization with our 3-stage {\method}. 
    The x-axis for {\method} uses a linear scale during the {\method} phase, followed by a logarithmic scale, with the transition indicated by a black vertical line.  
    \textbf{Right:} Average accuracy on 4 reasoning datasets (AIME24/25, Beyond AIME, and GPQA) versus average output tokens. Points in the upper-left region indicate better performance. Baseline names and their corresponding numbers are listed in Table~\ref{tab:results_tokens}. Red dots denote different stages of our {\method}.}
    \label{fig:overview}
    \vspace{-0.5em}
\end{figure}

In this work, we introduce \textbf{T}hinking-\textbf{F}ree \textbf{P}olicy \textbf{I}nitialization (\textbf{{\method}}), a simple yet effective adaptation to RLVR that bridges long Chain-of-Thought (CoT) distillation and standard RLVR.
We first observe that a \thinkfree{} operation—explicitly discarding the thinking content via a direct \texttt{\textless/think\textgreater} append—can substantially reduce token usage during inference (Section~\ref{subsec:thing-free reasoning}).  
Training with \thinkfree{}‑adapted inputs improves performance and lowers token consumption, even when evaluated in the original slow‑thinking mode (Section~\ref{subsec:thinking-free training}).  
We then formally define {\method} in Section~\ref{subsec:tpfi definition}.  
As illustrated in Figure~\ref{fig:overview}, {\method} accelerates RL convergence, achieves a higher performance ceiling, and produces more token‑efficient reasoning models without requiring specialized rewards or complex training designs (Section~\ref{sec:exp}).

Our contributions are as follows:  
\ding{182} We find that \thinkfree{} can substantially reduce inference costs for distilled LRMs, and that training with \thinkfree{}‑adapted inputs enhances slow‑thinking capability.
\ding{183} We propose {\method}, a fast, low‑cost initialization phase for long‑CoT RL that accelerates RL convergence and \revise{exhibits transfer} across domains, even when trained solely on mathematics.  
\ding{184} We show that long‑CoT RL following {\method} achieves a higher performance ceiling while producing more token‑efficient LRMs without the need for specialized rewards or complex training designs, offering an effective and efficient route to training high‑performing LRMs. 
\ding{185} We provide both behavioral and parameter‑level analyses and reveal that {\method} not only preserves the slow‑thinking reasoning pattern but also enables substantially faster rollouts in subsequent long‑CoT RL stages.

\section{Preliminary}
\textbf{Notation}.
In this paper, we define an LLM parameterized by $\theta$ as a policy $\pi_\theta$.  
Let $x$ denote a query and $\mathcal{D}$ the set of queries.  
Given a response $y$ to a query $x$, its likelihood under the policy $\pi_\theta$ is expressed as  
$\pi_\theta (y \mid x) = \prod_{t=1}^{|y|} \pi_\theta \left( y_t \mid x, y_{<t} \right),$
where $|y|$ denotes the number of tokens in $y$.  
A query–response pair $(x, y)$ is scored by a rule-based outcome reward $r(x, y) \in \{0, 1\}$, indicating whether the response $y$ aligns with the ground truth of $x$.

\textbf{\revise{RLVR Algorithms}}.
\revise{
RLVR algorithms are central to the recent success of LRMs, with prominent examples including Proximal Policy Optimization (PPO)~\citep{schulman2017ppo} and Group Relative Policy Optimization (GRPO)~\citep{shao2024grpo}.
PPO constrains updates to remain close to the previous policy $\pi_{\theta_\text{old}}$ via clipping.
GRPO bypasses the need for the value model by computing relative advantages among responses to the same query.
A widely adopted GRPO variant, DAPO~\citep{yu2025dapo}, adds clip-higher, dynamic sampling, and token-level policy-gradient losses to stabilize training.
The objective is given by $\mathcal{J}_{\text{DAPO}}(\theta) = \mathbb{E}_{x \sim \mathcal{D} } \left[ \mathcal{J}_{\text{DAPO}}(\theta, x) \right],$
{\small
\begin{align}
\mathcal{J}_{\text{DAPO}}(\theta, x) = & 
\Bigg[
\frac{1}{\sum_{i=1}^{G} |y_i|} 
\sum_{i=1}^{G} \sum_{t=1}^{|y_i|} 
\min \Big( 
r_{i,t}(\theta) \,\widehat{A}_{i,t},\, 
 \mathrm{clip}\big(r_{i,t}(\theta),
1 - \varepsilon_{\text{low}},  1 + \varepsilon_{\text{high}} \big) \,\widehat{A}_{i,t} 
\Big)
\Bigg], \nonumber \\ 
& \text{s.t.} \quad 0 < \left| \left\{ y_i \mid \texttt{is\_equivalent}(y_i, x) \right\} \right| < G.
\label{eq: dapo objective}
\end{align}
}Here, ${\{y_i\}_{i=1}^G \sim \pi_{\theta_\text{old}}( \cdot | x) }$ and $G$ is the number of generated responses to each query $x$ (i.e., the group size) and 
the importance ratio $r_{i,t}(\theta)$ and advantage $\widehat{A}_{i,t}$ of token $y_{i,t}$ are:
{\small
\begin{align}
    r_{i,t}(\theta)=\frac{ \pi_{\theta} (y_{i,t} | x, y_{i,<t}) }{ \pi_{\theta_\text{old}} (y_{i,t} | x,y_{i,<t})},\quad
    \widehat{A}_{i,t} = \widehat{A}_{i} = \frac{r(x, y_i) - \mathrm{mean} \left( \{ r(x, y_i) \}_{i=1}^G \right) }{ \mathrm{std} \left( \{ r(x, y_i) \}_{i=1}^G \right) },
    \label{eq: adv-ratio}
\end{align}
}respectively, where all the tokens in $y_i$ share the same advantage as $\widehat{A}_{i}$.
}

A concise overview of PPO and GRPO is provided in Appendix~\ref{app:preliminary}.
Across all experiments, we adopt DAPO for fair comparison.
In principle, any RLVR method can be applied to our {\method} stage.

\section{Methodology}

\subsection{Thinking-Free Mode Enables More Efficient Reasoning}\label{subsec:thing-free reasoning}

To generate a response $y \sim \pi_\theta(\cdot \mid x)$ for a query $x$ using an SFT-distilled LRM $\pi_\theta$, the query is typically formatted with a chat template.  
\cref{template:thinking} illustrates the template adopted by the Qwen model family~\citep{yang2025qwen3}.  
We define \thinkfree{} as an operator that transforms an input query $x$ into a modified query $x' = \thinkfree{}(x)$, in which the thinking content is explicitly omitted.  
Under this transformation, response generation follows $y \sim \pi_\theta(\cdot \mid x')$.  
This mechanism provides explicit control over whether reasoning content is present or absent in the generated output (see \cref{template:thinking free}).  
Additional examples using other chat templates are provided in Appendix~\ref{app:think-free operation}.  
Hereinafter, we use $x$ to denote a query formatted with the thinking template (e.g., \cref{template:thinking}), and $x'$ or $\thinkfree(x)$ to denote the corresponding thinking-free version (e.g., \cref{template:thinking free}).

\begin{tcolorbox}[colback=rliableblue!10!white,colframe=black,boxrule=1pt,boxsep=2pt,top=3pt,bottom=3pt,left=2pt,right=2pt]
\begin{template}[\textbf{\emph{Thinking Mode}}]
\label{template:thinking}
$<$\textbar im\_start\textbar$>$system\textbackslash nPlease reason step by step, and put your final answer within \textbackslash \textbackslash boxed\{\}.$<$\textbar im\_end\textbar$>$\textbackslash n$<$\textbar im\_start \textbar$>$user\textbackslash n\textbf{\{question (x)\}}\\$<$\textbar im\_end\textbar$>$\textbackslash n$<$\textbar im\_start\textbar$>$assistant\textbackslash n\looseness=-1
\end{template}
\vspace{0.05em}
\begin{template}[\textbf{\emph{Thinking-Free Mode}}]
\label{template:thinking free}
$<$\textbar im\_start\textbar$>$system\textbackslash nPlease reason step by step, and put your final answer within \textbackslash \textbackslash boxed\{\}.$<$\textbar im\_end\textbar$>$\textbackslash n$<$\textbar im\_start \textbar$>$user\textbackslash n\textbf{\{question (x)\}}\\$<$\textbar im\_end\textbar$>$\textbackslash n$<$\textbar im\_start\textbar$>$assistant\textbackslash n\looseness=-1 \textcolor{red}{\textless think\textgreater \textbackslash n\looseness=-1 \textbackslash n\looseness=-1\textless/think\textgreater}
\end{template}
\end{tcolorbox}

\begin{figure}[t]
    \centering
    \includegraphics[width=0.45\linewidth]{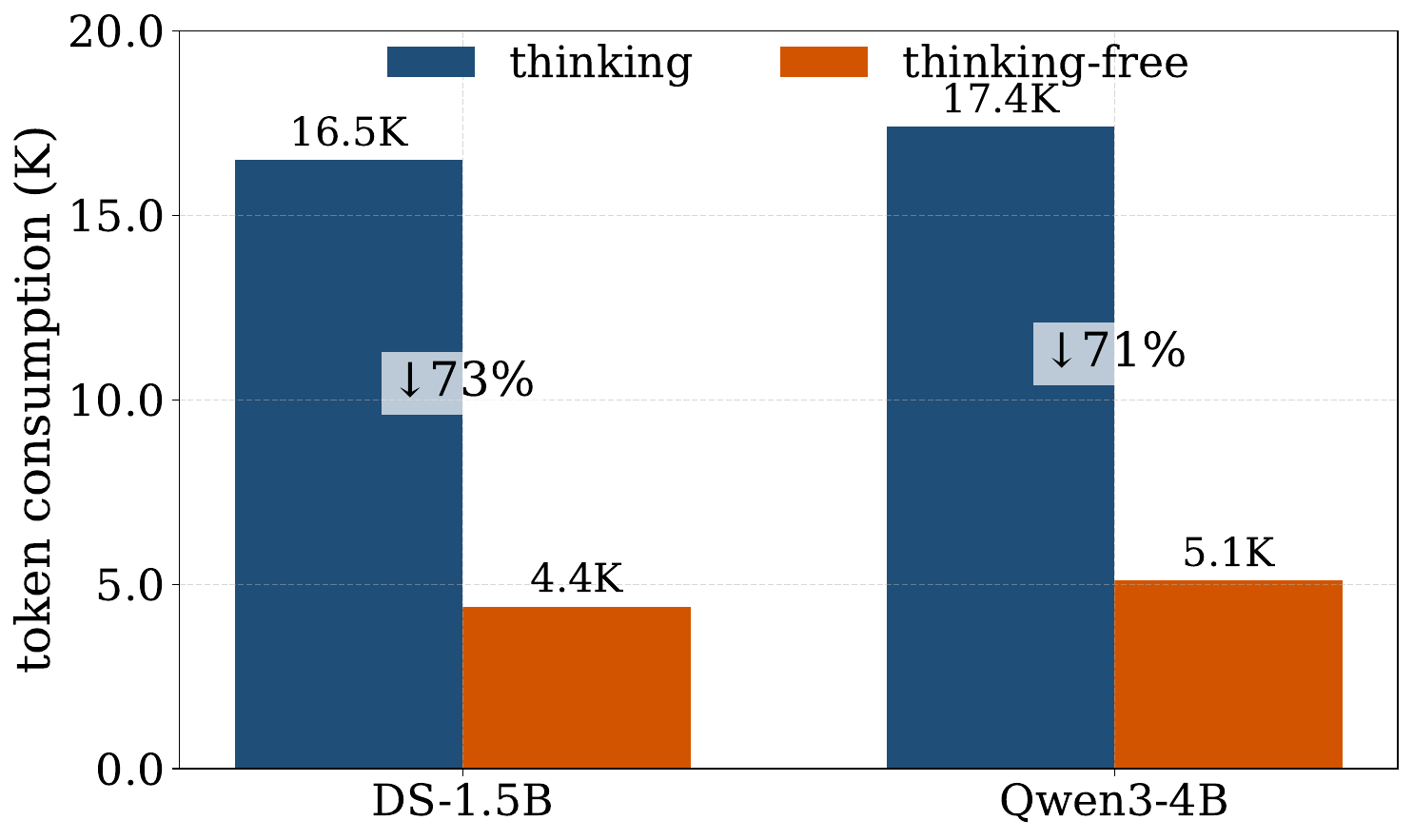}
    \includegraphics[width=0.45\linewidth]{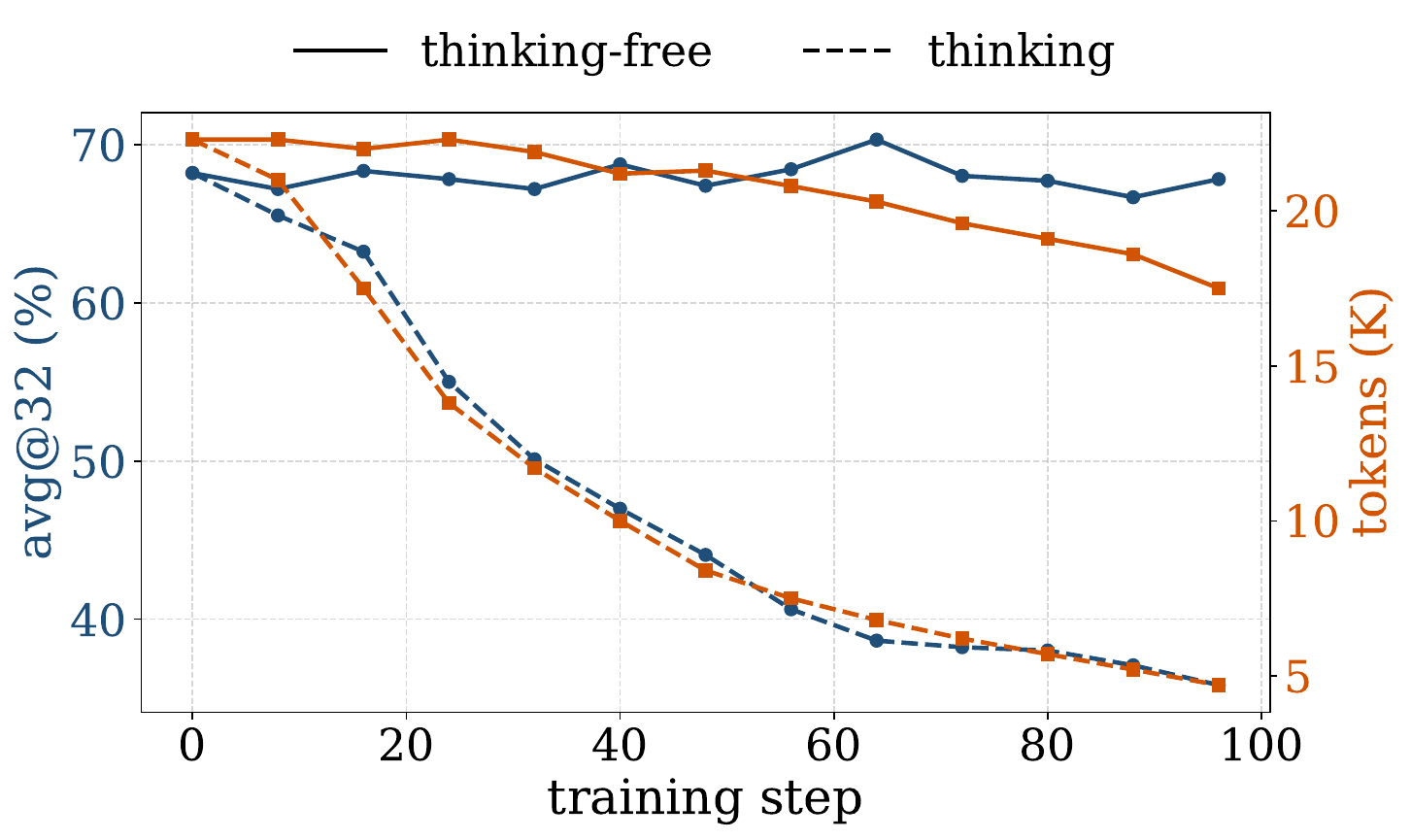}
    \vspace{-1em}
    \caption{Results of the meta-experiment on the \thinkfree{} operation. 
\textbf{Left:} Average output tokens in thinking mode and \thinkfree{} mode on AIME25. 
\textbf{Right:} Evolution of \texttt{avg@32} and average output tokens on AIME24 with thinking-mode evaluation over training steps \revise{under 4K training response length (both training in thinking mode and \thinkfree{} mode)}.}
    \label{fig:meta-exp}
    \vspace{-0.5em}
\end{figure}

\textbf{During inference, converting $x$ into its thinking-free version $x'$ can substantially reduce token consumption.} 
To assess how this transformation affects the reasoning capability of SFT-distilled LRMs, we evaluate \texttt{DeepSeek-Distilled-Qwen-1.5B} (abbreviated as \texttt{DS-1.5B}) and \texttt{Qwen3-4B} on the AIME25.
Detailed experimental setup is delayed to Appendix~\ref{app:meta-exp-token}.
As shown in Figure~\ref{fig:meta-exp} (Left), applying the \thinkfree{} reduces the number of output tokens by more than 70\% for both models.  
It is worth noting that \texttt{Qwen3-4B} is a fast–slow fusion model, whereas \texttt{DS-1.5B} is an SFT-distilled long-CoT model; nevertheless, both exhibit the same trend.

\subsection{Thinking-Free Training Is Beneficial to Slow-Thinking}\label{subsec:thinking-free training}

To train an SFT-distilled LRM, the training response length should not be too short~\citep{setlur2025e3}, as this is detrimental to testing performance~\citep{Polaris2025}. 
As evidenced by the dotted line in Figure~\ref{fig:meta-exp} (Right), training \texttt{Qwen-3-4B} with a 4K response length using DAPO indeed leads to a substantial drop in performance on AIME25. 
Given that using the \thinkfree{} variant for inference can significantly reduce token consumption, we pose an audacious question: can we apply the \thinkfree{} operation to all input queries during the rollout stage of RLVR? 
Moreover, will this approach be beneficial to preserving the original slow-thinking capability of the trained LLM?

Surprisingly, \textbf{RL trained with \thinkfree{} rollout can slightly improve accuracy and reduce token consumption when evaluated in thinking mode, even with very short training context lengths. }
Figure~\ref{fig:meta-exp} (Right) illustrates the training dynamics of \texttt{Qwen-3-4B} trained with a 4K response length under the \thinkfree{} transformation of queries (detailed settings are in Appendix~\ref{app:meta-exp-train}). Even with a 4K output length, \thinkfree{} RL increases the accuracy on AIME25 in thinking mode by approximately 2\% , while reducing output tokens by around 20\%. 
In contrast, standard RLVR with a 4K length reduces \texttt{avg@32} by more than 40\%. 
These results suggest that applying \thinkfree{} during rollout can yield steady improvements with minimal training compute.

\subsection{Thinking Free Policy Initialization}\label{subsec:tpfi definition}

\citet{hu2025prepretaining} introduce a pre-pretraining stage using formal language to accelerate convergence of pretraining, while \citet{wang2025octothinker} propose a mid-training stage between pretraining and RL-zero to facilitate RLVR.  
Inspired by these works, we ask: \textit{can a dedicated stage for SFT-distilled LRMs improve the efficiency and effectiveness of standard RLVR scaling?} 
\thinkfree{} RL, which enhances slow-thinking within short training context windows, requires substantially less compute than standard RLVR.  
Initializing RLVR with a \thinkfree{} RL policy could therefore reduce rollout tokens while achieving stronger downstream performance.  
We term this step as \textbf{T}hinking \textbf{F}ree \textbf{P}olicy \textbf{I}nitialization (\textbf{\method{}}), a stage preceding standard RLVR for SFT-distilled LRMs that aims to lower rollout costs, raise the ceiling of reasoning ability, and accelerate convergence.

Consider the RLVR objective $\mathcal{J}_\text{RLVR}(\theta) = \mathbb{E}_{x \sim \mathcal{D}} \left[ \mathcal{J}_\text{RLVR}(\theta, x) \right],$
where $\mathcal{J}_\text{RLVR}(\theta, x)$ denotes the per-example objective of any RLVR algorithm (e.g., DAPO in \cref{eq: dapo objective} or GRPO in \cref{equ:grpo}).  
For the \method{} stage, we use a modified objective: $\mathcal{J}_\text{{\method}}(\theta) = \mathbb{E}_{x \sim \mathcal{D}} \left[ \mathcal{J}_\text{RLVR}(\theta, {\color{red}x'}) \right],$
where ${\color{red}x' = \thinkfree(x)}$.  
In the rollout stage of {\method}, $G$ responses are generated conditioned on $x'$: $\{ y_i \}_{i=1}^G \sim \pi_{\theta_\text{old}}(\cdot \mid {\color{red}x'}).$
The importance ratios and advantages (\cref{eq: adv-ratio}) are then adapted as:  
{\small
\begin{align}
    r_{i,t}(\theta) = \frac{\pi_{\theta}(y_{i,t} \mid {\color{red}x'}, y_{i,<t})}{\pi_{\theta_\text{old}}(y_{i,t} \mid {\color{red}x'}, y_{i,<t})}, \quad
    \widehat{A}_{i,t} = \widehat{A}_{i} = \frac{r({\color{red}x'}, y_i) - \mathrm{mean}\!\left( \{ r({\color{red}x'}, y_j) \}_{j=1}^G \right)}{\mathrm{std}\!\left( \{ r({\color{red}x'}, y_j) \}_{j=1}^G \right)},
    \label{eq:adv-ratio-tpi}
\end{align}
}where \revise{\cref{template:thinking} and \cref{template:thinking free} showcase the correspondence between $x$ and $x'$} and $r(x', y) = r(x, y)$ since the \thinkfree{} operator does not alter the ground-truth answer of the original problem.  
In our experiments, we instantiate RLVR with DAPO, i.e., $\mathcal{J}_\text{RLVR}(\theta) = \mathcal{J}_\text{DAPO}(\theta)$ (see \cref{eq: dapo objective}).
\revise{Note that Heuristics for RLVR (e.g., multi-stage RL, data curricula) can be directly applied to {\method}.}

\section{Experiments}\label{sec:exp}

\subsection{Experimental Settings}\label{subsec:exp_setting}

In this section, we provide a brief overview of the key experimental setup, including training procedures, baselines, and evaluation details. Additional information can be found in Appendix~\ref{app:exp-setup}.

\textbf{Training Details}.
We build on the VeRL codebase~\citep{sheng2024verl} with RLVR, following the DAPO recipe~\citep{yu2025dapo}, which enables dynamic sampling and clipping higher.
All methods use the same hyperparameters (batch size $256$, learning rate $1\!\times\!10^{-6}$, no warm-up) and rollout settings (temperature $1$, \texttt{topp} $1$, \texttt{topk} $-1$, 8 rollouts/problem). 
Training is conducted on \texttt{DS-1.5B}, \texttt{Qwen3-4B}, and \texttt{DS-7B} using Polaris-53K~\citep{Polaris2025}. 
Direct RLVR uses a maximum output length of 16K for \texttt{DS-1.5B}/\texttt{DS-7B} and 32K for \texttt{Qwen3-4B}, while {\method} adopts multi-stage training: $2$\text{K}$\!\rightarrow\!4$\text{K}$\!\rightarrow\!8$\text{K} for \texttt{DS-1.5B}/\texttt{DS-7B} and $4$\text{K}$\!\rightarrow\!8$\text{K}$\!\rightarrow\!16$\text{K} for \texttt{Qwen3-4B}.

\textbf{Baselines}.
We compare {\method} with direct RLVR training from an SFT-distilled LRM (``Direct RL'') under matched total training compute (Table~\ref{tab:main_result}). 
To assess {\method} as a pre-RLVR stage, we run ``{\method} + RL'' with similar compute and compare against ``Direct RL'' (Table~\ref{tab:sota}), also including competitive LRMs of the same size, such as Polaris~\citep{Polaris2025}, DeepScaleR~\citep{deepscaler2025}, Skywork-OR1~\citep{he2025skyworkreasoner}, and so on. 
For efficiency analysis, we compare with several RL-based efficient reasoning baselines (Table~\ref{tab:results_tokens}) under the same evaluation for fair comparison, including TLMRE~\citep{arora2025TLMRE}, AdaptThink~\citep{zhang2025adaptthink}, AutoThink~\citep{tu2025AutoThink}, Laser~\citep{liu2025Laser}, L1Max~\citep{aggarwal2025L1}, and ThinkLess~\citep{fang2025thinkless}.

\textbf{Evaluation Details}.
Our evaluation benchmarks cover:
\ding{182} \textbf{Math Reasoning}: AIME24/25 and BeyondAIME~\citep{bytedance_seed_2025_beyondaime}.
\ding{183} \textbf{Multi-Task Reasoning}: GPQA-Diamond~\citep{rein2024gpqa}.
\ding{184} \textbf{Code Generation}: LiveCodeBench~\citep{jain2024livecodebench}.
\ding{185} \textbf{Instruction Following}: IFEval~\citep{zhou2023ifeval}.
Following \citet{guo2025deepseekr1}, we generate multiple outputs (ranging from 4 to 32 depending on the size of the test set) per problem and report \texttt{pass@1} accuracy.
Note that \ding{182} is in-domain evaluation, and \ding{183} \ding{184} \ding{185} are out-of-domain evaluation.
For IFEval, we report the \textit{strict prompt} accuracy. 
All evaluation scripts are adapted from the DeepscaleR codebase~\citep{deepscaler2025}.
\revise{Following the officially recommended decoding parameters\footnote{See \href{https://huggingface.co/Qwen/Qwen3-4B}{Qwen3-4B Hugging Face page} and \href{https://huggingface.co/deepseek-ai/DeepSeek-R1-Distill-Qwen-1.5B}{DS-1.5B Hugging Face page}.}, we set the temperature to 0.6 and top-p to 0.95 for the thinking mode, and $0.7, 0.8$ for the thinking-free mode, respectively. 
The maximum output length is 32k for all configurations, except for the \texttt{Qwen3-4B} thinking mode, where 48k is used to reduce the clipping ratio. 
Further details are provided in Appendix~\ref{app:evaluation}, and the effects of changing decoding parameters are presented in Appendix~\ref{app:decoding_para}.}

\subsection{{\method} Enhances the Slow-Thinking of Distilled Reasoning Models}
\begin{table}[!t]
    \centering
    \caption{Results of {\method} vs. direct RL across different benchmarks.  
``\texttt{Avg@k}'' denotes the average accuracy (\%) over $k$ random generations (i.e., \texttt{pass@1}).  
All models are evaluated in thinking mode.  
The total training compute for the 3 stages of {\method} equals that of ``Direct RL'' for fair comparison.
Darker colors in the cell background denote better results within each model group.}
\vspace{-1em}
    \label{tab:main_result}
    \vspace{2mm}
    \resizebox{0.95\textwidth}{!}{
    \begin{tabular}{l|ccc|c|c|c|c}
    \toprule[1.6pt]
        \multirow{3}{*}{\textbf{Models}} 
        & \multicolumn{3}{c|}{\textbf{Mathematics}} 
        & \textbf{Multi-Task} 
        & \textbf{Code} 
        & \textbf{Instruction} 
        & \textbf{Overall} \\ \cmidrule(lr){2-8}
        ~ & \textbf{AIME 24} 
           & \textbf{AIME 25} 
           & \textbf{Beyond AIME} 
           & \textbf{GPQA}
           & \textbf{LiveCode}
           & \textbf{IFEval}
           & \textbf{Overall} \\ 
        & \texttt{Avg@32} & \texttt{Avg@32} & \texttt{Avg@8} & \texttt{Avg@8} & \texttt{Avg@8} & \texttt{Avg@4} & \texttt{Avg.} \\ 
        \midrule
        \multicolumn{8}{c}{\texttt{\textbf{{DeepSeek-Distill-Qwen-1.5B}}}} \\
        \midrule
        Initial Model & \cellcolor{lightblue1}{29.6} & \cellcolor{lightblue1}{23.0} & \cellcolor{lightblue1}{8.7} & \cellcolor{lightblue1}{16.3} & \cellcolor{lightblue1}{17.7} & \cellcolor{lightblue1}{36.6} & \cellcolor{lightblue1}{22.0} \\ 
- Direct RL & \cellcolor{lightblue3}{33.9} & \cellcolor{lightblue3}{27.1} & \cellcolor{lightblue2}{11.9} & \cellcolor{lightblue2}{19.2} & \cellcolor{lightblue2}{18.2} & \cellcolor{lightblue4}{41.8} & \cellcolor{lightblue2}{25.3} \\
- {\method} stage 1 & \cellcolor{lightblue2}{32.1} & \cellcolor{lightblue2}{26.9} & \cellcolor{lightblue4}{13.9} & \cellcolor{lightblue3}{29.3} & \cellcolor{lightblue3}{18.4} & \cellcolor{lightblue1}{39.5} & \cellcolor{lightblue3}{26.7} \\
- {\method} stage 2 & \cellcolor{lightblue4}{36.8} & \cellcolor{lightblue4}{28.4} & \cellcolor{lightblue5}{14.5} & \cellcolor{lightblue4}{27.8} & \cellcolor{lightblue5}{20.5} & \cellcolor{lightblue5}{42.3} & \cellcolor{lightblue4}{28.4} \\
- {\method} stage 3 & \cellcolor{lightblue5}{40.1} & \cellcolor{lightblue5}{30.8} & \cellcolor{lightblue3}{13.8} & \cellcolor{lightblue5}{29.6} & \cellcolor{lightblue4}{19.9} & \cellcolor{lightblue3}{40.8} & \cellcolor{lightblue5}{29.2} \\
        \midrule
        \multicolumn{8}{c}{\texttt{\textbf{{Qwen3-4B}}}} \\
        \midrule
        Initial Model     & \cellcolor{lightgreen2}{73.6} & \cellcolor{lightgreen3}{68.3} & \cellcolor{lightgreen2}{43.4} & \cellcolor{lightgreen2}{56.8} & \cellcolor{lightgreen3}{54.9} & \cellcolor{lightgreen2}{64.9} & \cellcolor{lightgreen3}{60.3} \\
Direct RL         & \cellcolor{lightgreen3}{75.7} & \cellcolor{lightgreen1}{67.0} & \cellcolor{lightgreen3}{43.6} & \cellcolor{lightgreen1}{56.3} & \cellcolor{lightgreen1}{52.5} & \cellcolor{lightgreen4}{66.0} & \cellcolor{lightgreen2}{60.2} \\
{\method} stage 1 & \cellcolor{lightgreen1}{75.2} & \cellcolor{lightgreen2}{67.8} & \cellcolor{lightgreen1}{42.4} & \cellcolor{lightgreen4}{57.9} & \cellcolor{lightgreen4}{55.3} & \cellcolor{lightgreen4}{66.0} & \cellcolor{lightgreen4}{60.8} \\
{\method} stage 2 & \cellcolor{lightgreen4}{76.0} & \cellcolor{lightgreen4}{68.2} & \cellcolor{lightgreen4}{44.7} & \cellcolor{lightgreen3}{57.8} & \cellcolor{lightgreen2}{54.8} & \cellcolor{lightgreen1}{64.8} & \cellcolor{lightgreen5}{61.0} \\
{\method} stage 3 & \cellcolor{lightgreen5}{79.9} & \cellcolor{lightgreen5}{70.6} & \cellcolor{lightgreen5}{46.7} & \cellcolor{lightgreen5}{58.5} & \cellcolor{lightgreen5}{57.0} & \cellcolor{lightgreen5}{70.2} & \cellcolor{lightgreen5}{63.8} \\
        \midrule
        \multicolumn{8}{c}{\texttt{\textbf{{DeepSeek-Distill-Qwen-7B}}}} \\
        \midrule
        Initial Model & \cellcolor{lightcyan1}{56.3} & \cellcolor{lightcyan1}{40.0} & \cellcolor{lightcyan1}{25.0} & \cellcolor{lightcyan1}{36.9} & \cellcolor{lightcyan2}{39.5} & \cellcolor{lightcyan1}{55.3} & \cellcolor{lightcyan1}{42.2} \\ 
- Direct RL & \cellcolor{lightcyan2}{57.9} & \cellcolor{lightcyan3}{40.4} & \cellcolor{lightcyan2}{26.6} & \cellcolor{lightcyan2}{38.0} & \cellcolor{lightcyan1}{38.3} & \cellcolor{lightcyan3}{56.7} & \cellcolor{lightcyan2}{43.0} \\
- {\method} stage 1 & \cellcolor{lightcyan3}{59.4} & \cellcolor{lightcyan3}{40.4} & \cellcolor{lightcyan3}{29.2} & \cellcolor{lightcyan5}{49.0} & \cellcolor{lightcyan3}{39.6} & \cellcolor{lightcyan2}{56.2} & \cellcolor{lightcyan3}{45.6} \\
- {\method} stage 2 & \cellcolor{lightcyan4}{61.8} & \cellcolor{lightcyan4}{43.9} & \cellcolor{lightcyan5}{31.5} & \cellcolor{lightcyan4}{48.0} & \cellcolor{lightcyan4}{42.0} & \cellcolor{lightcyan4}{57.1} & \cellcolor{lightcyan4}{47.4} \\
- {\method} stage 3 & \cellcolor{lightcyan5}{62.0} & \cellcolor{lightcyan5}{44.6} & \cellcolor{lightcyan4}{31.1} & \cellcolor{lightcyan3}{46.8} & \cellcolor{lightcyan5}{42.1} & \cellcolor{lightcyan5}{60.2} & \cellcolor{lightcyan5}{47.8} \\
        \bottomrule[1.6pt]
    \end{tabular}}
\vspace{-0.6em}
\end{table}

To evaluate the impact of {\method} on the slow-thinking capabilities, %
we present the results of {\method} versus ``Direct RL'' under the same training compute in Table~\ref{tab:main_result}.  
From the table, we have:  

\ding{182} \textbf{{\method} substantially enhances the slow-thinking capabilities of distilled LRMs even when trained with a small response length.}  
For example, on \texttt{DS-1.5B}, {\method} Stage~1 raises the overall average accuracy from $22.0\%$ to $26.7\%$ (+4.7\%) despite being restricted to a 2K training response length.  
Results of \texttt{DS-1.5B} continue to improve through stages~1 to~3 on AIME25, accuracy increases from $23.0$\% (initial model) to $26.9$\% after Stage~1, $28.4$\% after Stage~2, and $30.8$\% after Stage~3, representing a total gain of +7.8\%.  
Similar improvements are observed for \texttt{Qwen3-4B} and \texttt{DS-7B}.  
These findings indicate that {\method} enables effective training under low-cost settings (short context windows), and that combining this with multi-stage RL yields substantial accuracy gains.

\ding{183} \textbf{Compared with ``Direct RL'', {\method} delivers faster and larger accuracy improvements under the same training cost.}  
{\method} outperforms ``Direct RL'' in nearly all configurations.  
For example, \texttt{Qwen3-4B} with {\method} attains $63.8\%$ overall accuracy versus $60.2\%$ for ``Direct RL'' (+3.6\%), while \texttt{DS-7B} improves by +4.8\% ($47.8\%$ vs.\ $43.0\%$).  
Given that the equal training compute, %
these results imply that {\method} achieves convergence more efficiently than conventional long-CoT RL training.

\ding{184} \textbf{Improvements of {\method} \revise{indicate some degree of transfer beyond mathematics}.}
Although {\method} is trained solely on Polaris-53K (math-specific data), it demonstrates great out-of-domain improvements as well. %
For example, on \texttt{DeepSeek-Distill-Qwen-1.5B}, GPQA accuracy increases from $16.3\%$ to $29.6\%$, LiveCodeBench from $17.7\%$ to $19.9\%$, and IFEval from $36.6\%$ to $40.8\%$ after Stage~3.  
Notably, \textbf{improvements on mathematical benchmarks are often consistent across successive training stages, whereas other domains sometimes exhibit fluctuations} (e.g., GPQA for \texttt{DS-7B} and LiveCodeBench for \texttt{DS-1.5B}).  
This suggests that \textbf{incorporating more diverse training data spanning multiple domains could be helpful for {\method}}.

\subsection{{\method} as a Foundation for RLVR to Achieve Higher Performance}\label{subsec:sota}
\begin{table}[!t]
    \centering
    \caption{Results (\%) of RL after {\method} (``{\method}+RL'') vs. ``Direct RL'' across different benchmarks.  
``\texttt{Avg@k}'' denotes the average accuracy (\%) over $k$ random generations (i.e., \texttt{pass@1}).  
For LRMs marked with ``*'', results are taken from the corresponding reports (see Appendix~\ref{app:source}); results of 4B models are from our own runs with 48K response length.  
All models are evaluated in thinking mode.  
The total training compute for ``{\method}+RL'' is matched to that of ``Direct RL'' for fair comparison.
Darker colors in the cell background denote better results.}
    \vspace{-1em}
    \label{tab:sota}
    \vspace{2mm}
    \resizebox{0.95\textwidth}{!}{
    \begin{tabular}{l|ccc|c|c|c|c}
    \toprule[1.6pt]
        \multirow{3}{*}{\textbf{Models}} 
        & \multicolumn{3}{c|}{\textbf{Mathematics}} 
        & \textbf{Multi-Task} 
        & \textbf{Code} 
        & \textbf{Instruction} 
        & \textbf{Overall} \\ \cmidrule(lr){2-8}
        ~ & \textbf{AIME 24} 
           & \textbf{AIME 25} 
           & \textbf{Beyond AIME} 
           & \textbf{GPQA}
           & \textbf{LiveCode}
           & \textbf{IFEval}
           & \textbf{Overall} \\ 
        & \texttt{Avg@32} & \texttt{Avg@32} & \texttt{Avg@8} & \texttt{Avg@8} & \texttt{Avg@8} & \texttt{Avg@4} & \texttt{Avg.} \\ 
        \midrule
        DeepSeek R1$^*$ 
& \cellcolor{lightpurple4}{79.8} 
& \cellcolor{lightpurple1}{65.0} 
& \cellcolor{lightpurple1}{42.4} 
& \cellcolor{lightpurple4}{71.5} 
& \cellcolor{lightpurple4}{64.3} 
& \cellcolor{lightpurple4}{86.1} 
& \cellcolor{lightpurple4}{68.2} \\

Seed-1.5-Thinking$^*$ 
& \cellcolor{lightpurple5}{86.7} 
& \cellcolor{lightpurple4}{74.0} 
& \cellcolor{lightpurple4}{48.0} 
& \cellcolor{lightpurple5}{77.3} 
& \cellcolor{lightpurple5}{64.9} 
& \cellcolor{lightpurple5}{87.4} 
& \cellcolor{lightpurple6}{73.0} \\

Claude4 Opus Thinking$^*$ 
& - 
& \cellcolor{lightpurple3}{75.5} 
& - 
& \cellcolor{lightpurple6}{79.6} 
& \cellcolor{lightpurple1}{48.9} 
& \cellcolor{lightpurple6}{89.7} 
& - \\

Qwen3-235B-Thinking$^*$ 
& \cellcolor{lightpurple3}{85.7} 
& \cellcolor{lightpurple6}{81.5} 
& - 
& \cellcolor{lightpurple2}{71.1} 
& \cellcolor{lightpurple3}{55.7} 
& \cellcolor{lightpurple3}{83.4} 
& - \\

\midrule
Qwen3-4B Direct RL 
& 78.8 
& \cellcolor{lightpurple2}{71.5} 
& \cellcolor{lightpurple3}{46.4} 
& \cellcolor{lightpurple1}{56.2} 
& \cellcolor{lightpurple2}{54.3} 
& \cellcolor{lightpurple1}{65.1} 
& \cellcolor{lightpurple1}{62.0} \\

Qwen3-4B {\method} stage 3 
& \cellcolor{lightpurple2}{79.9} 
& \cellcolor{lightpurple1}{70.6} 
& \cellcolor{lightpurple2}{46.7} 
& \cellcolor{lightpurple2}{58.5} 
& \cellcolor{lightpurple4}{57.0} 
& \cellcolor{lightpurple2}{70.2} 
& \cellcolor{lightpurple2}{63.8} \\

Qwen3-4B {\method} + RL 
& \cellcolor{lightpurple1}{80.8} 
& \cellcolor{lightpurple3}{76.0} 
& \cellcolor{lightpurple5}{49.5} 
& \cellcolor{lightpurple3}{61.1} 
& \cellcolor{lightpurple3}{55.7} 
& \cellcolor{lightpurple1}{71.3} 
& \cellcolor{lightpurple3}{65.7} \\
\midrule
Qwen3-4B-2507-Thinking 
& \cellcolor{lightpurple2}{84.7} 
& \cellcolor{lightpurple4}{79.2} 
& \cellcolor{lightpurple6}{51.5} 
& \cellcolor{lightpurple4}{66.4} 
& \cellcolor{lightpurple2}{62.4} 
& \cellcolor{lightpurple2}{68.0} 
& \cellcolor{lightpurple4}{68.7} \\

- {\method} only 
& \cellcolor{lightpurple6}{89.0} 
& \cellcolor{lightpurple5}{81.2} 
& \cellcolor{lightpurple5}{51.3} 
& \cellcolor{lightpurple2}{70.1} 
& \cellcolor{lightpurple6}{65.5} 
& \cellcolor{lightpurple2}{66.7} 
& \cellcolor{lightpurple5}{70.6} \\
        \bottomrule[1.6pt]
    \end{tabular}}
    \vspace{-0.6em}
\end{table}

Table~\ref{tab:sota} shows the representative results between ``{\method} + RL'' with ``Direct RL'' under the same training cost (full results in Table~\ref{tabapp:sota} in Appendix~\ref{app:more_results}).
The results lead to the following observations:

\ding{182} \textbf{{\method} can raise the upper bound of RL-trained performance.}
RL training on a {\method}-trained model can still yield notable improvements, particularly in mathematics.
For example, adding an RL stage after {\method} for \texttt{Qwen3-4B} boosts AIME25 accuracy from $70.6\%$ to $76.0\%$ ($+5.4\%$).
Similarly, Beyond AIME scores increase by more than $2\%$.
Across different model scales, applying RL after {\method} consistently achieves higher accuracies than ``Direct RL'' under the same compute budget.  
For instance, on \texttt{Qwen3-4B}, the overall accuracy rises from $62.0\%$ (Direct RL) to $65.7\%$ ({\method}+RL).%
These results suggest that incorporating {\method} as an intermediate stage between long CoT distillation and standard RLVR can be beneficial for elevating final performance.

\ding{183} \textbf{{\method} + RL is an effective and efficient strategy for training high-performing LRMs.}
From Figure~\ref{fig:overview} (Left), the total compute cost of all three {\method} stages amounts to less than $20\%$ of standard RL training with $32$\textsc{k} token sequences.  
From Table~\ref{tab:sota}, for \texttt{DS-7B}, {\method}+RL achieves approximately the same overall performance as \texttt{Polaris-7B-Preview}, despite using a maximum response length of $16$\textsc{k}, whereas \texttt{Polaris-7B-Preview} follows a $16$\textsc{k} $\rightarrow$ $24$\textsc{k} $\rightarrow$ $32$\textsc{k} progression during RL.  
Similarly, \texttt{Polaris-4B-Preview} employs a $40$\textsc{k} $\rightarrow$ $48$\textsc{k} $\rightarrow$ $52$\textsc{k} length schedule and consumes approximately $8$K H800 GPU hours, while our {\method}+RL requires only about $1.5$K H800 GPU hours (see Appendix~\ref{app:training}) and achieves superior performance under $48$\textsc{k} testing length.  
Even without a subsequent RL stage, {\method} alone allows \texttt{DS-1.5B} to outperform \texttt{DeepScaleR}, which uses a maximum training length of $24$\textsc{k}.
Using only {\method} (4K$\rightarrow$8K$\rightarrow$16K), \texttt{Qwen3‑4B‑2507} achieves 89\% accuracy on AIME24.
Remarkably, this 4B model outperforms \texttt{Qwen3‑235B-Thinking} in math reasoning and code generation.
These findings suggest that {\method} can serve both as a strong standalone training approach and as an efficient foundation for subsequent RL, producing competitive LRMs with significantly reduced compute requirements.

\subsection{{\method} Improves the Token Efficiency of Distilled Reasoning Models}\label{subsec:token}
\begin{table}[!t]
    \centering
    \caption{Comparison of the Thinking-Free inference mode of {\method} with efficient reasoning baselines across various reasoning tasks.
``\texttt{Avg@k}'' denotes the average accuracy (in \%) over k generations (i.e., \texttt{pass@1}), and “\texttt{Toks}” indicates the average output length in thousands of tokens (K).
Models with ``*'' are trained from \texttt{DeepScaleR-1.5B}, while the remaining are from \texttt{DS-1.5B}.
Darker cell background colors indicate better results.
    }
    \vspace{-1em}
    \label{tab:results_tokens}
    \vspace{2mm}
    \resizebox{0.95\textwidth}{!}{
    \begin{tabular}{l|cc|cc|cc|cc|cc|cc}
    \toprule[1.6pt]
        \multirow{3}{*}{\textbf{Models}} 
        
           & \multicolumn{2}{c|}{\textbf{AIME 24}} 
           & \multicolumn{2}{c|}{\textbf{AIME 25}} 
           & \multicolumn{2}{c|}{\textbf{Beyond AIME}} 
           & \multicolumn{2}{c|}{\textbf{GPQA}} 
           & \multicolumn{2}{c|}{\textbf{Overall}} \\ \cmidrule(lr){2-11}
        & \texttt{avg@32} & \texttt{Toks}
        & \texttt{avg@32} & \texttt{Toks}
        & \texttt{avg@8} & \texttt{Toks}
        & \texttt{avg@8} & \texttt{Toks}
        & \texttt{Avg.} & \texttt{Toks} \\ 
        \midrule
        1) TLMRE-DS-1.5B ($\alpha=0.1$) & 27.6 & 12.9 & \cellcolor{lightred1}{24.8} & 12.5 & 9.2 & 11.9 & 14.8 & 7.5 & 19.1 & 11.2 \\
        2) AdaptThink-1.5B ($\delta=0.05$) & 28.1 & \cellcolor{lightorange1}{8.0} & 22.6 & 7.9 & \cellcolor{lightred3}{10.0} & \cellcolor{lightorange2}{4.8} & 14.8 & 5.1 & 18.9 & \cellcolor{lightorange1}{6.5} \\
        3) AutoThink-DS-1.5B-Stage3 & \cellcolor{lightred3}{30.3} & 10.2 & \cellcolor{lightred3}{25.2} & 9.1 & \cellcolor{lightred1}{9.4} & 8.7 & 17.4 & 7.0 & \cellcolor{lightred1}{20.6} & 8.7 \\
        4) Laser-DE-L4096-1.5B & \cellcolor{lightred3}{30.3} & 8.3 & \cellcolor{lightred2}{24.9} & \cellcolor{lightorange1}{7.4} & \cellcolor{lightred2}{9.7} & 7.4 & \cellcolor{lightred1}{21.1} & \cellcolor{lightorange1}{4.7} & \cellcolor{lightred2}{21.5} & 6.9 \\
        \hline
        5) AutoThink-Stage3$^*$ & \cellcolor{lightred6}{38.9} & 8.7 & \cellcolor{lightred6}{28.9} & 7.7 & \cellcolor{lightred5}{11.6} & 7.8 & \cellcolor{lightred2}{27.3} & 5.4 & \cellcolor{lightred5}{26.7} & 7.4 \\
        6) L1-1.5B-Max$^*$ & 27.2 & \cellcolor{lightorange5}{3.2} & \cellcolor{lightred4}{26.3} & \cellcolor{lightorange5}{2.9} & 9.1 & \cellcolor{lightorange4}{3.1} & \cellcolor{lightred3}{32.4} & \cellcolor{lightorange3}{2.3} & \cellcolor{lightred3}{23.8} & \cellcolor{lightorange4}{2.9} \\
        7) Thinkless-1.5B-RL$^*$ & 28.4 & 11.3 & 24.1 & 11.1 & 8.1 & 11.7 & 20.3 & 12.7 & 20.2 & 11.7 \\
        \hline
        DS-1.5B (Thinking) & \cellcolor{lightred1}{29.6} & 16.7 & 23.0 & 16.5 & 8.7 & 14.4 & 16.3 & 9.8 & 19.4 & 14.3 \\ 
        DS-1.5B (Thinking-Free) & 12.4 & \cellcolor{lightorange2}{5.7} & 10.9 & \cellcolor{lightorange3}{4.4} & 4.4 & \cellcolor{lightorange3}{3.4} & 4.2 & \cellcolor{lightorange5}{0.9} & 8.0 & \cellcolor{lightorange3}{3.6} \\ 
        - {\method} stage 1 & 21.9 & \cellcolor{lightorange6}{1.6} & 15.3 & \cellcolor{lightorange6}{1.4} & 8.7 & \cellcolor{lightorange6}{1.3} & \cellcolor{lightred4}{32.9} & \cellcolor{lightorange6}{0.8} & 19.7 & \cellcolor{lightorange6}{1.3} \\
        - {\method} stage 2 & \cellcolor{lightred4}{31.5} & \cellcolor{lightorange4}{3.4} & 24.2 & \cellcolor{lightorange4}{3.1} & \cellcolor{lightred4}{10.1} & \cellcolor{lightorange5}{2.9} & \cellcolor{lightred5}{35.3} & \cellcolor{lightorange4}{1.6} & \cellcolor{lightred4}{25.3} & \cellcolor{lightorange5}{2.7} \\
        - {\method} stage 3 & \cellcolor{lightred5}{37.5} & \cellcolor{lightorange3}{5.3} & \cellcolor{lightred5}{28.4} & \cellcolor{lightorange2}{5.0} & \cellcolor{lightred6}{12.4} & \cellcolor{lightorange1}{4.9} & \cellcolor{lightred6}{35.6} & \cellcolor{lightorange2}{2.6} & \cellcolor{lightred6}{28.5} & \cellcolor{lightorange2}{4.4} \\
        \bottomrule[1.6pt]
    \end{tabular}}
    \vspace{-0.6em}
\end{table}

We compare the thinking-free inference with other RL-based efficient reasoning baselines in 1.5B size in Table~\ref{tab:results_tokens} (see also Table~\ref{tabapp:results_tokens} in Appendix~\ref{app:more_results}).
We draw the following conclusions:

\ding{182} \textbf{Both accuracy and token efficiency steadily improve after stage~3 of {\method}.}  
For \texttt{DS-1.5B}, thinking-free accuracy on AIME24 increases from $29.6\%$ (initial model) to $37.5\%$ ({\method} stage~3), while the average output length remains substantially shorter than that of the original thinking mode (5.3K vs.\ 16.7K tokens).  
A similar trend is observed for \texttt{Qwen3-4B}, where accuracy improves from $26.9\%$ to $75.1\%$ across stages, with output lengths still far below those of the original thinking model.
These results demonstrate that {\method} naturally produces more token-efficient LRMs, offering an alternative pathway to train models that deliver both high accuracy and reduced output length.

\ding{183} \textbf{Compared with other RL-based token control methods, {\method} achieves the best performance–efficiency trade-off without specialized reward or training designs.}  
In \texttt{DS-1.5B}, both {\method} stage~2 and stage~3 outperform almost all baselines in terms of overall accuracy while maintaining competitive or lower token usage.  
We visualize the overall accuracy–token usage trade-off in Figure~\ref{fig:overview} (Right), where {\method} consistently lies on the Pareto frontier across different stages.  
This observation motivates a rethinking of existing ``token-efficient reasoning RL designs'' that rely heavily on specialized length reward shaping: training with {\method} offers an alternative paradigm in which a strong slow-thinking LRM can be obtained, and a more efficient variant can be realized simply by switching to the thinking-free mode without any additional length-control mechanisms.

\section{Analysis}

\subsection{Why {\method} Leads to Better Thinking-Mode Inference?}\label{subsec:reason}

\begin{figure}[t]
    \centering
        \includegraphics[width=0.45\linewidth]{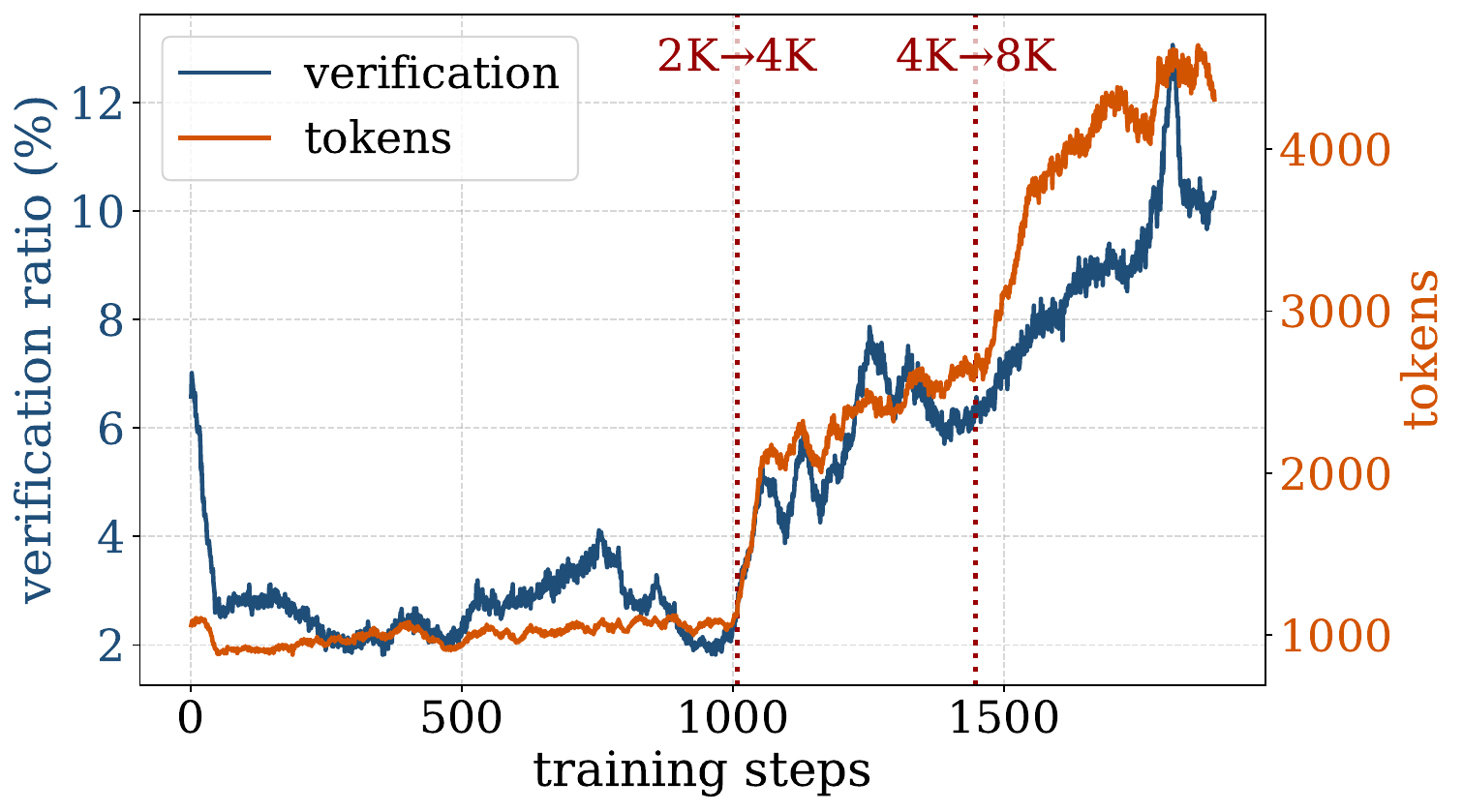}
        \includegraphics[width=0.45\linewidth]{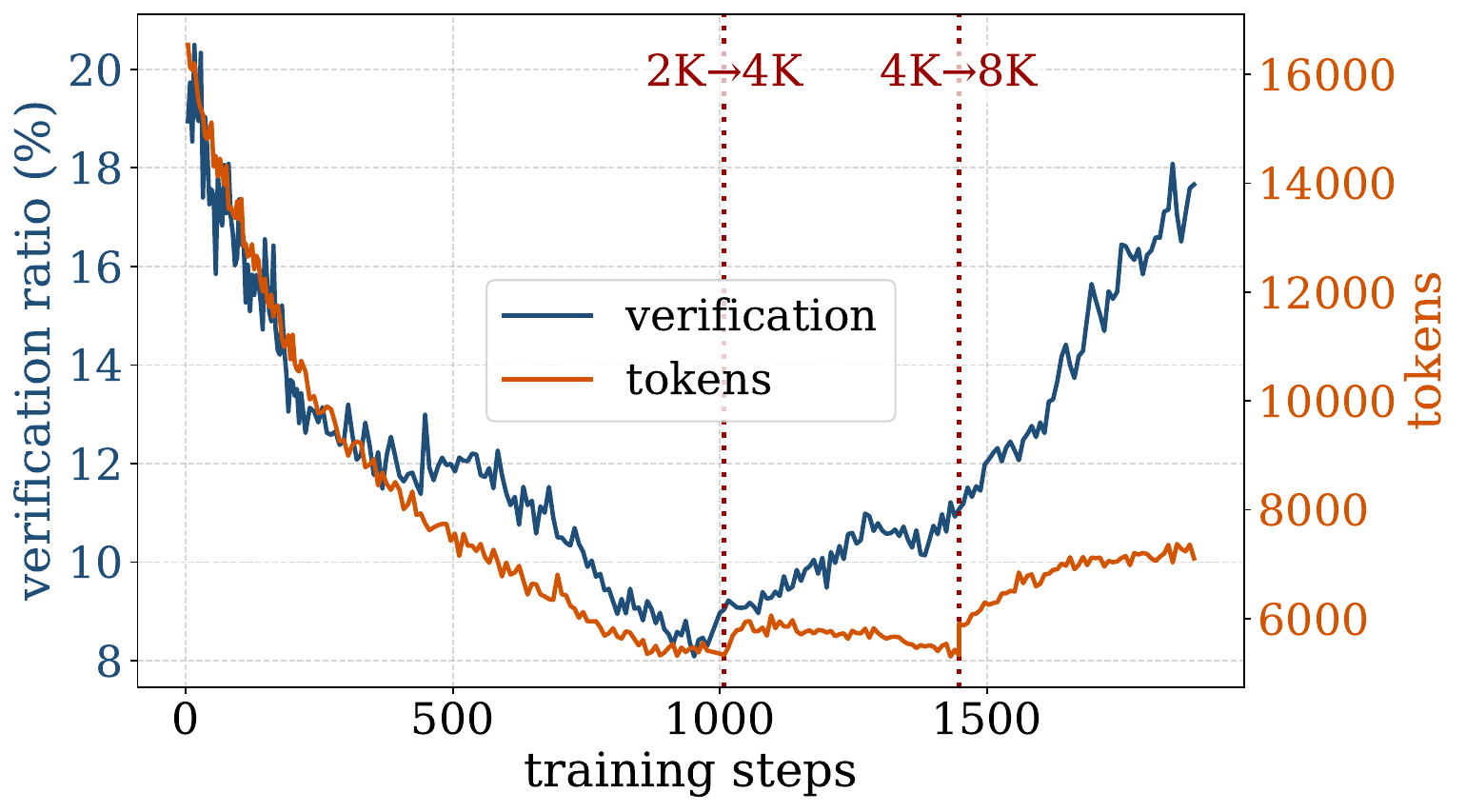}
    \vspace{-1em}
    \caption{Behaviour-Level Analysis of \texttt{DS-1.5B} over the {\method} Training Course. The ratio of verification steps and the average output tokens over training steps on the training set in thinking-free mode (\textbf{Left}) and on AIME25 in thinking mode (\textbf{Right}) in 3 stages of {\method}.}
    \label{fig:behaviour-analysis}
    \vspace{-1em}
\end{figure}

In this section, we delve deeper into why {\method}, which leverages void thinking content in \cref{template:thinking free}, can generalize to enhance reasoning in \cref{template:thinking}. 
We analyze \texttt{DS-1.5B} from two perspectives:

\ding{182} \textbf{Behavioural level:} The learned \textit{verification} behaviours after \texttt{\textless/think\textgreater} during {\method} can generalize to the slow-thinking verification occurring within the \texttt{\textless think\textgreater} and \texttt{\textless/think\textgreater}. 
The blue lines in Figure~\ref{fig:behaviour-analysis} show the ratio of verification steps (i.e., the number of verification steps divided by the total number of steps; see Appendix~\ref{app:analysis} for details) for the training set (thinking-free mode) and AIME 25 dataset (thinking mode). 
We observe that the verification ratio exhibits a similar trend: a rapid drop in Stage~1, followed by steady growth in Stage~2, and a sharp increase in Stage~3. Notably, the sharp decline in Stage~1 resembles an information compression process. In Stages~2 and~3, the model begins to explore more extensively (Figure~\ref{fig:behaviour-analysis} Right), which may explain why {\method} achieves superior performance.
As verification is believed to be vital for slow-thinking reasoning~\citep{setlur2025e3}, the observed generalization of verification behavior suggests a transfer from thinking-free training to inference in thinking mode.

\begin{figure}[t]
    \centering
        \includegraphics[width=0.45\linewidth]{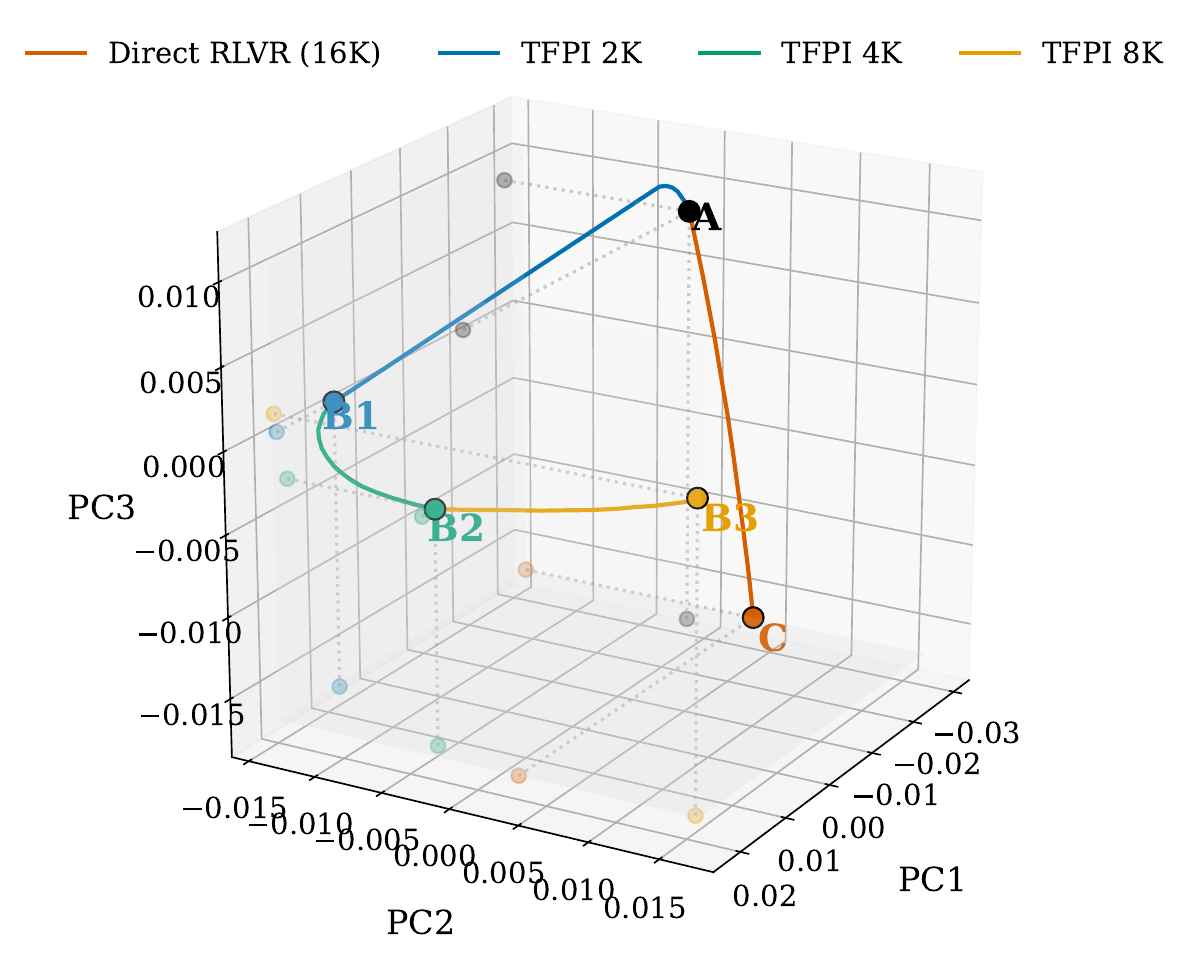}
        \includegraphics[width=0.45\linewidth]{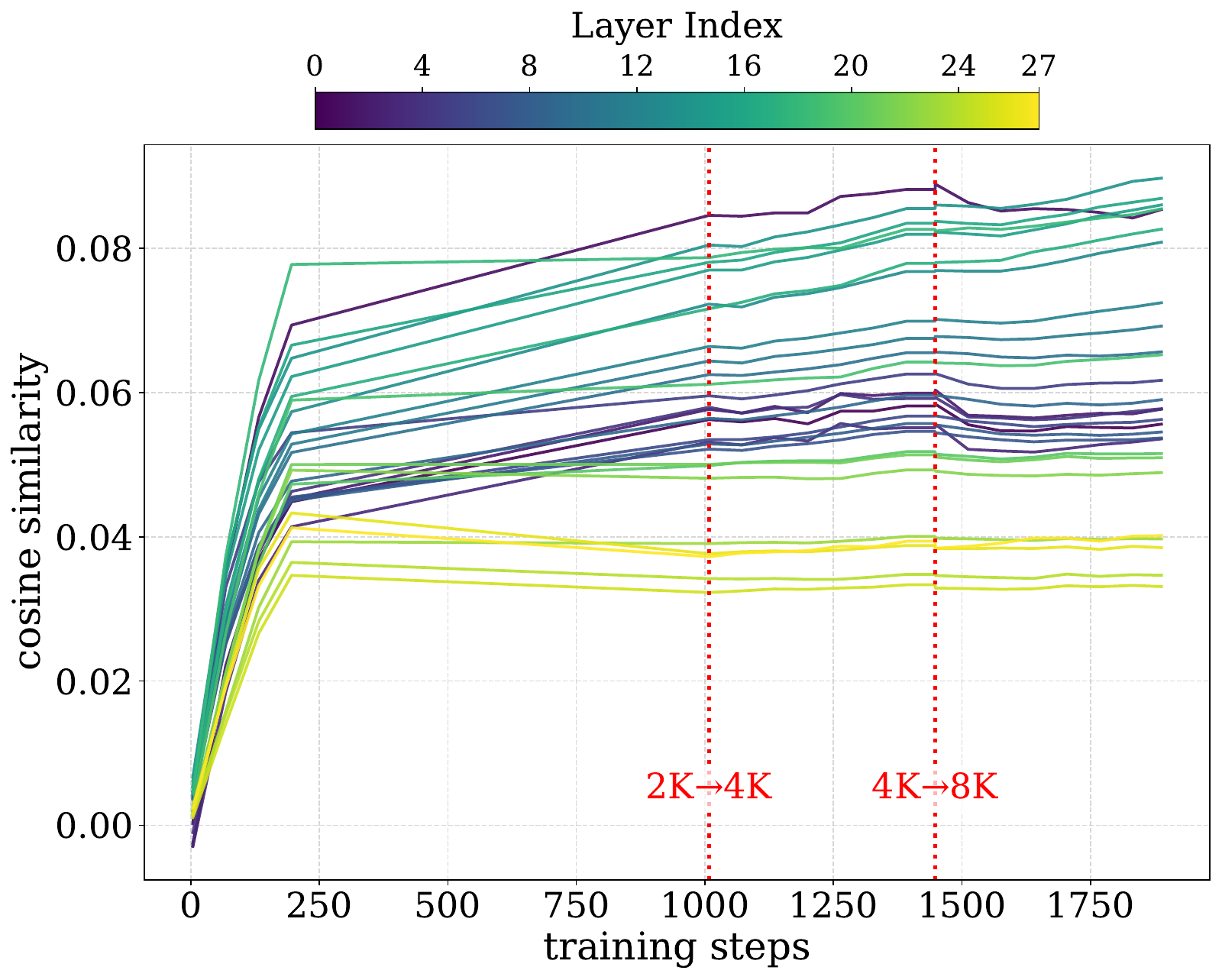}
    \vspace{-1em}
    \caption{
Parameter-Level Analysis. 
\textbf{Left:} PCA projection of model parameters from \texttt{DS-1.5B} to final checkpoints. 
{\method} (blue) starts at \textbf{A}, passes through intermediate points (\textbf{B1}, \textbf{B2}, \textbf{B3}), and ultimately converges near the Direct RL final checkpoint (\textbf{C}). 
\textbf{Right:} Cosine similarity between parameter updates of {\method}-trained checkpoints and (C-A) across layers during training. 
}
    \label{fig:parameter-analysis}
    \vspace{-1em}
\end{figure}

\ding{183} \textbf{Parameter level:} {\method} explores the parameter space more extensively at a faster pace, with its parameter update directions progressively aligning with those of ``Direct RL.'' 
As shown in Figure~\ref{fig:parameter-analysis} (Left), the PCA visualization of the initial model, {\method}-trained checkpoints, and Direct RL checkpoints exhibits distinct trajectories in parameter space. 
The {\method} begins from the initial model (\textbf{A}), moves through intermediate points (\textbf{B1}), (\textbf{B2}), and (\textbf{B3}), and ultimately converges to a region near the Direct RL final checkpoint (\textbf{C}).
This indicates that {\method} traverses a larger and more diverse region of parameter space before reaching a point close to the RL-trained model. 
Such broad exploration may help explain why {\method} can lead to better LRMs.  
Furthermore, Figure~\ref{fig:parameter-analysis} (Right) shows that the cosine similarity of parameter updates between the {\method}-trained checkpoints and the Direct RL final checkpoint steadily increases across nearly all layers throughout training. 
This suggests that during {\method} training, the parameter updates share similarities with those in standard long-CoT training.
\revise{To make our analysis more robust, additional results are given in Appendix~\ref{app:para-analysis}.}

\subsection{About Reasoning Pattern and Rollout Speed}

\begin{figure}[t]
    \centering
        \includegraphics[width=0.45\linewidth]{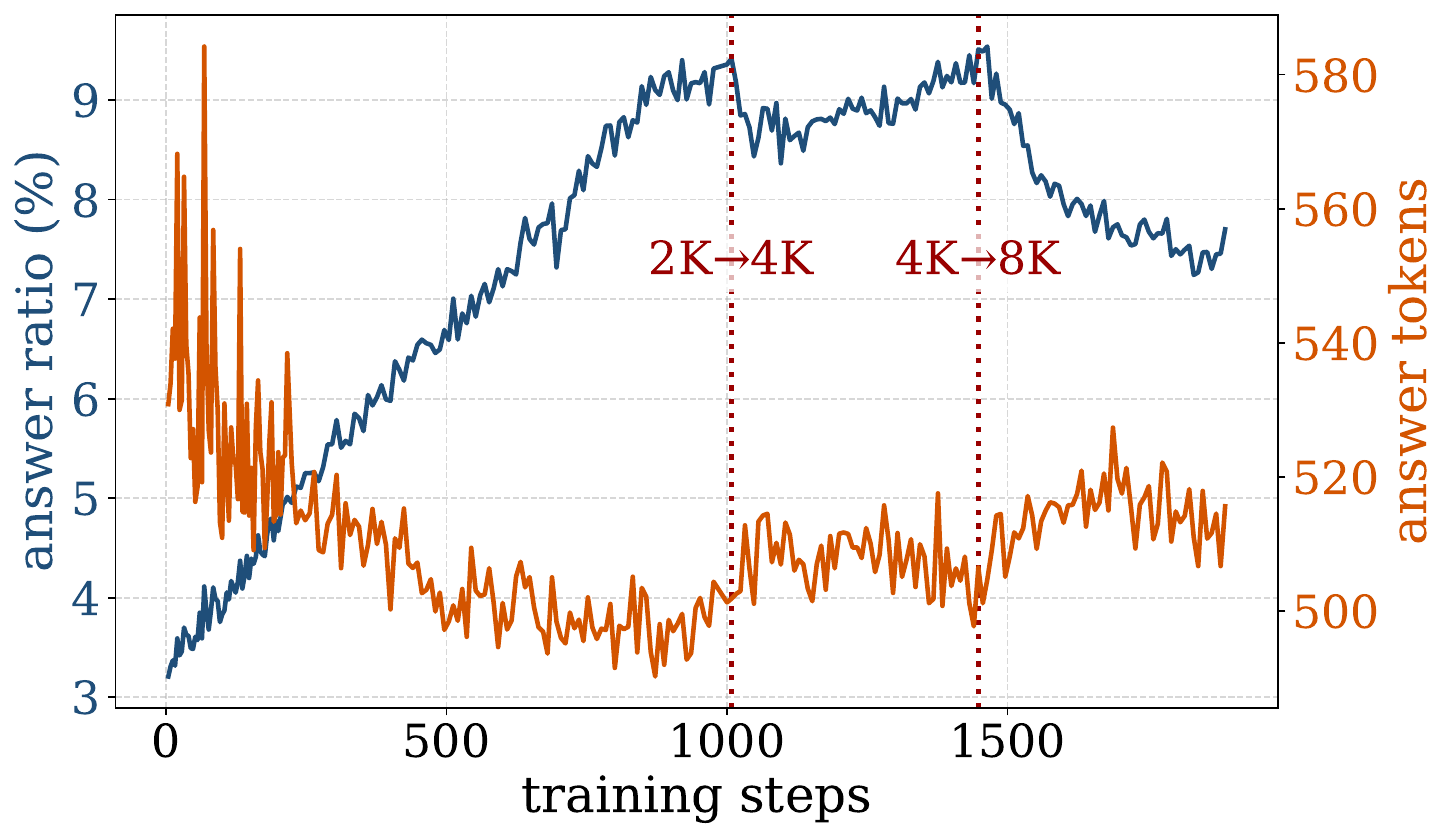}
        \includegraphics[width=0.45\linewidth]{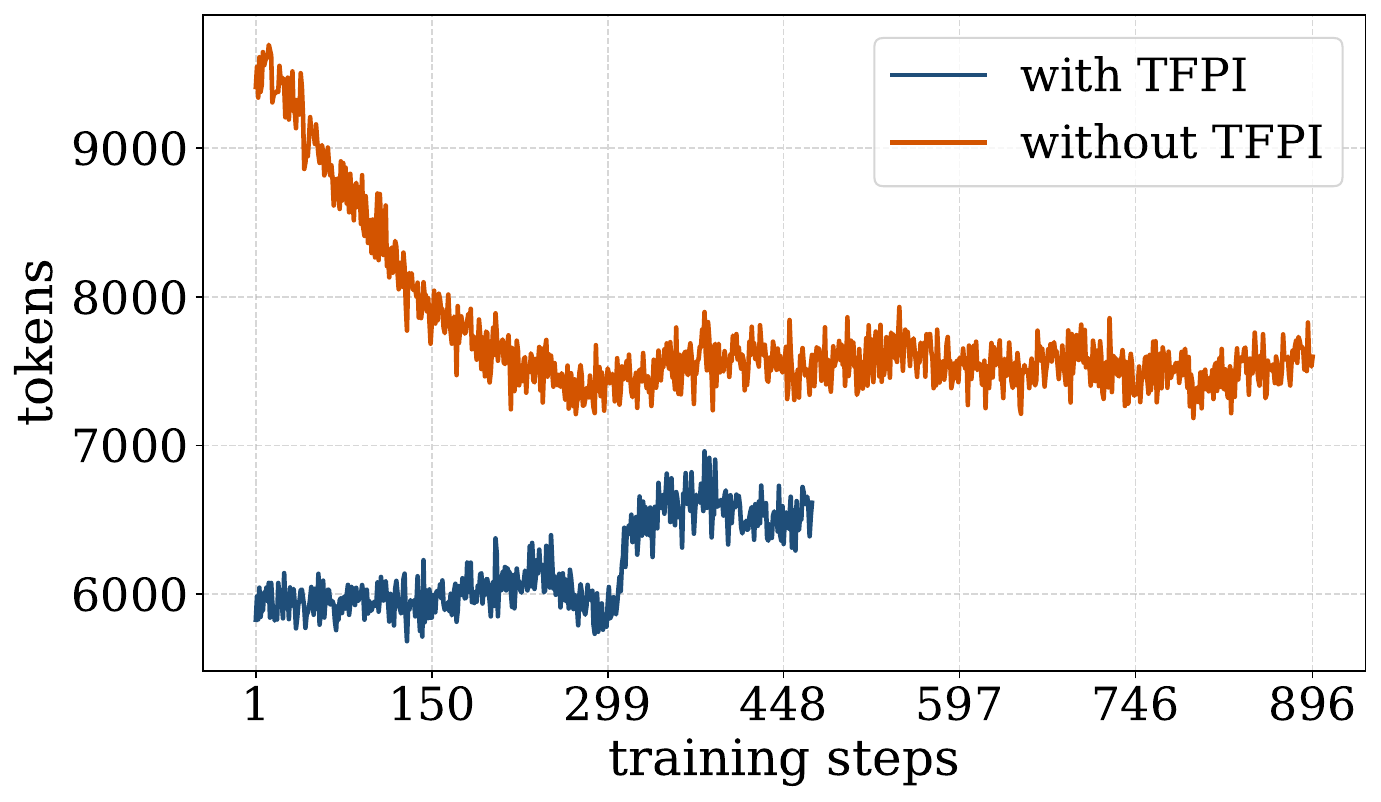}
    \vspace{-1em}
    \caption{\textbf{Left:} For {\method} with \texttt{DS-1.5B} on AIME25 in thinking mode, showing the average number of answer tokens (excluding the thinking part) and the ratio of answer length to total response length over training steps. 
    \textbf{Right:} For long CoT RL with \texttt{DS-1.5B}, showing the average number of tokens during rollout on the training set over training steps, with and without the {\method} stage.}
    \label{fig:analysis}
    \vspace{-0.5em}
\end{figure}

\textbf{{\method} Preserves the Reasoning Pattern of Thinking Mode.} 
In \emph{thinking mode} (\cref{template:thinking}), a response $y$ comprises a long \emph{thinking} section and a concise \emph{answer} $y^\text{ans}$. 
In \emph{\thinkfree{}} mode (\cref{template:thinking free}), the thinking section is omitted, and $y^\text{ans}$ contains a shorter reasoning path. 
While standard RL tends to lengthen the thinking part, {\method} increases the length of $y^\text{ans}$ due to the absence of explicit thinking. 
As shown in Figure~\ref{fig:analysis} (Left), for \texttt{DS-1.5B} trained with {\method} and evaluated in thinking mode, $|y^\text{ans}|$ remains stable at $500$–$580$ tokens, whereas $|y^\text{ans}| / |y|$ rises as the total length $|y|$ decreases. 
This indicates that {\method} preserves the core reasoning pattern of slow-thinking rather than drifting toward an excessively extended ``slow-slow thinking'' behavior.

\textbf{{\method} Speeds Up Rollout for Long-CoT RL Training}.
Another advantage of {\method} is its ability to speed up the rollout stage in standard long-CoT RL training.
As shown in Figure~\ref{fig:analysis} (Right), when directly performing RL from \texttt{DS-1.5B}, the average output tokens during rollout on the training set start at over 9K, decrease to around 7.5K within the first 300 steps, and then fluctuate around 7.5K in the later stages.
In contrast, when RL is performed after the {\method} phase, the average output tokens start at only 6K and the maximum length is below 7K tokens.

\subsection{\revise{Scaling Up Model Size}}\label{subsec:scaleup}

\revise{To assess the effect of {\method} on larger reasoning models, we conducted experiments with \texttt{Qwen3-14B} under the same settings as \texttt{Qwen3-4B}. 
As shown in Table~\ref{tab:furhter}, {\method} still outperforms Direct RL under the same training compute. 
Notably, Direct RL yields results only comparable to those of the initial model (66.9\% vs. 66.8\% on average). 
We hypothesize that Polaris-53K is, in some sense, too easy for \texttt{Qwen3-14B} (with over 78\% rollout accuracy), leaving only approximately 30\% effective data after dynamic sampling (excluding 0/8 and 8/8 prompts). 
Nonetheless, {\method} remains effective in this setting. 
Additionally, {\method} reduces the average number of output tokens on the test set by 23.8\% even in thinking mode (14.1K vs. 18.5K). 
These results highlight the effectiveness of {\method} for larger model sizes, and we believe that with more challenging, higher-quality data, {\method} can further realize its scalability potential.}

\begin{table}[!t]
    \centering
    \caption{\revise{Additional results of {\method}. All models are evaluated in thinking mode.}}
\vspace{-1em}
    \label{tab:furhter}
    \vspace{2mm}
    \resizebox{0.95\textwidth}{!}{
    \begin{tabular}{l|ccc|c|c|c|c}
    \toprule[1.6pt]
        \multirow{1}{*}{\textbf{Models}} 
         & \textbf{AIME 24} 
           & \textbf{AIME 25} 
           & \textbf{Beyond AIME} 
           & \textbf{GPQA}
           & \textbf{LiveCode}
           & \textbf{IFEval}
           & \textbf{Overall} \\ 
        \midrule
        \multicolumn{8}{c}{\texttt{\textbf{{Stage Schedule Sensitivity, \texttt{Qwen3-4B}}}}} \\
        \midrule
        $4\rm{K}\rightarrow8\rm{K}\rightarrow16\rm{K}$ & {\textbf{79.9}} & {70.6} & {\textbf{46.7}} & {58.5} & {\textbf{57.0}} & {\textbf{70.2}} & {\textbf{63.8}} \\
        $8\rm{K}\rightarrow16\rm{K}$ & 78.0 & 70.2 & \textbf{46.7} & \textbf{59.5} & 55.8 & 64.5 & 62.4 \\
        $16\rm{K}$ only & 77.1 & \textbf{71.5} & 45.8 & 56.4 & 54.4 & 66.0 & 61.9 \\
        \midrule
        \multicolumn{8}{c}{\texttt{\textbf{{Multi-Stage Direct RL, \texttt{Qwen3-4B}}}}} \\
        \midrule
        {\method} Stage 3 & {\textbf{79.9}} & {\textbf{70.6}} & {\textbf{46.7}} & {\textbf{58.5}} & {\textbf{57.0}} & {\textbf{70.2}} & {\textbf{63.8}} \\
        Multi-Stage Direct RL & 64.9 & 48.6 & 32.5 & 55.2 & 51.0 & 65.2 & 52.9 \\
        \midrule
        \multicolumn{8}{c}{\texttt{\textbf{{Qwen3-14B}}}} \\
        \midrule
        Initial Model & 81.7 & 73.2 & 50.2 & 65.0 & 63.5 & 67.3 & 66.8 \\ 
- Direct RL & 81.7 & 74.5 & \textbf{50.8} & 64.8 & 61.8 & \textbf{68.2} & 66.9 \\
- {\method} stage 3 & \textbf{83.2} & \textbf{76.0} & 50.5 & \textbf{65.6} & \textbf{63.8} & 67.8 & \textbf{67.8} \\
\midrule
        \multicolumn{8}{c}{\texttt{\textbf{{Qwen3-14B Tokens (K)}}}} \\
        \midrule
        Initial Model & 16.8 & 20.8 & 22.8 & 9.4 & 16.0 & - & 17.1 \\ 
- Direct RL & 17.9 & 22.1 & 24.5 & 10.4 & 17.4 & - & 18.5 \\
- {\method} stage 3 & \textbf{13.8} & \textbf{17.3} & \textbf{18.3} & \textbf{7.3} & \textbf{13.8} & - & \textbf{14.1} \\
        \bottomrule[1.6pt]
    \end{tabular}}
\vspace{-0.5em}
\end{table}

\subsection{\revise{Ablation Study}}\label{subsec:ablation}

\textbf{\revise{Stage Schedule Sensitivity}}.
\revise{To assess sensitivity to the stage schedule, we train \texttt{Qwen3-4B} under alternative schedules ($8\mathrm{K}\!\rightarrow\!16\mathrm{K}$ and 16K-only) with approximately matched compute budgets. 
As shown in Table~\ref{tab:furhter}, all schedules outperform the Direct RL baseline and achieve comparable performance on math benchmarks, indicating that {\method} is robust to the choice of schedule. 
Nevertheless, $4\mathrm{K}\!\rightarrow\!8\mathrm{K}\!\rightarrow\!16\mathrm{K}$ yields the best overall results (63.8\% vs. 62.4\% and 61.9\%). 
We hypothesize that this stems from an implicit data curriculum induced by dynamic sampling: beginning at 4K improves training efficiency and emphasizes easier problems early on.}

\textbf{\revise{Multi-Stage Direct RL}}.
\revise{As discussed in Section~\ref{subsec:thinking-free training} and Figure~\ref{fig:meta-exp} (right), applying Direct RL with a 4K response length causes a substantial performance drop on AIME25 for \texttt{Qwen3-4B}. 
In contrast, {\method} trains stably even at 4K. 
This indicates that, even in a multi-stage setting, the initial stage length should not be too short.
To rule out the potential artifact introduced by multi-stage training, we run Direct RL with the identical multi-stage schedule on \texttt{Qwen3-4B}. 
As shown in Table~\ref{tab:furhter}, Direct RL still incurs a substantial performance decrease under the $4\mathrm{K}\!\rightarrow\!8\mathrm{K}\!\rightarrow\!16\mathrm{K}$ schedule. 
Additional details are provided in Appendix~\ref{app:sb}.}

\section{Related Work}\label{sec:related_work}

\textbf{LRMs \& Efficiency RLVR}. 
RLVR has enabled the development of numerous high-performing large reasoning models (LRMs)~\citep{yang2025qwen3, zeng2025glm4_5, Polaris2025}, inspiring research on model behaviors~\citep{liu2025drgrpo, wang2025beyond8020}, novel algorithms~\citep{liu2025drgrpo, zheng2025gspo}, multimodal extensions~\citep{meng2025mmeureka, xiao2025perception, wang2025papo}, tool integration~\citep{jin2025searchr1, feng2025retool, li2025torl, xue2025simpletir, team2025kimik2}, and other related directions~\citep{xu2025doublechecker, zhang2025verifyrl, chen2025verithinker, zhang2025critiquegrpo}.  
RLVR can be applied directly to base LLMs (``RL zero'')~\citep{zeng2025simplerl, guo2025deepseekr1} or initialized from SFT-distilled long-CoT models, the latter typically yielding stronger results~\citep{deepscaler2025, Polaris2025}.  
A major challenge for RLVR is the cost of training with long contexts, as longer outputs are often necessary for harder tasks~\citep{shrivastava2025gfpo, zeng2025glm4_5}, consuming high computational resources.  
Multi-stage RLVR mitigates this by starting with shorter contexts and gradually extending them~\citep{deepscaler2025, Polaris2025, he2025skyworkreasoner}, while algorithmic approaches modify GRPO to reduce length bias~\citep{yu2025dapo, liu2025drgrpo, wang2025beyond8020}.  
Orthogonal to these strategies, we introduce \method{} as a lightweight stage before RLVR, improving efficiency and strengthening the slow-thinking mode at inference with minimal training cost, thus facilitating more effective subsequent RLVR.

\textbf{Efficient Reasoning}.
To address the issue of overthinking~\citep{chen2024overthinking,sui2025overthinkingsurvey}, considerable efforts have been made, including prompt-based methods~\citep{muennighoff2025s1,yang2025deer,fu2025dynasor,chen2025seal,fu2025deepconf}, SFT-based approaches~\citep{kang2025c3ot,ma2025cot-value,munkhbat2025self-train-efficient,luo2025o1-pruner}, and RL-related designs.
RL-related approaches can be further categorized into length-based reward shaping~\citep{team2025kimik1.5,aggarwal2025L1,arora2025TLMRE,liu2025Laser}, integration of fast and slow thinking~\citep{fang2025thinkless,zhang2025adaptthink,lou2025adacot,tu2025AutoThink,jiang2025LHRMs,zhang2025ASRR}, and thinking budget control~\citep{li2025selfbudgeter,hammoud2025curriculum-grpo,wen2025budgetthinker}.
These methods primarily trade accuracy for efficiency and rely on specialized reward functions or training strategies to encourage more efficient reasoning.
In contrast, our proposed \method{} naturally yields even more efficient LRMs without specialized rewards or training designs.

\section{Conclusion}

After recognizing the benefits of {\thinkfree} for both inference and the training of distilled reasoning models, we introduce {\method}, a cost‑efficient intermediate stage between long‑CoT distillation and standard RL training.
As a strong initialization point, {\method} accelerates RL convergence, enhances attainable performance, and promotes more token‑efficient reasoning without complex reward shaping or elaborate training pipelines.
We further explain the factors behind {\method}'s success from both the behavioral and parameter levels.
Overall, {\method} provides a complementary path for building “token‑efficient” LRMs, offering an effective and efficient alternative to current RL paradigms.

\section*{Acknowledgements}
This work was partially supported by an Area of Excellence project (AoE/E-601/24-N), a Theme-based Research Project (T32-615/24-R), and the Innovation and Technology Commission (ITCPD/179) from the Research Grants Council of the Hong Kong Special Administrative Region, China. We would like to thank all reviewers for their helpful suggestions in improving this paper.

\bibliography{iclr2026_conference}

\begin{thebibliography}{74}
\providecommand{\natexlab}[1]{#1}
\providecommand{\url}[1]{\texttt{#1}}
\expandafter\ifx\csname urlstyle\endcsname\relax
  \providecommand{\doi}[1]{doi: #1}\else
  \providecommand{\doi}{doi: \begingroup \urlstyle{rm}\Url}\fi

\bibitem[Aggarwal \& Welleck(2025)Aggarwal and Welleck]{aggarwal2025L1}
Pranjal Aggarwal and Sean Welleck.
\newblock L1: Controlling how long a reasoning model thinks with reinforcement
  learning.
\newblock \emph{arXiv preprint arXiv:2503.04697}, 2025.

\bibitem[An et~al.(2025)An, Xie, Li, Li, Zhang, Gong, Zhong, Xu, Qiu, Wang, and
  Kong]{Polaris2025}
Chenxin An, Zhihui Xie, Xiaonan Li, Lei Li, Jun Zhang, Shansan Gong, Ming
  Zhong, Jingjing Xu, Xipeng Qiu, Mingxuan Wang, and Lingpeng Kong.
\newblock Polaris: A post-training recipe for scaling reinforcement learning on
  advanced reasoning models, 2025.
\newblock URL \url{https://hkunlp.github.io/blog/2025/Polaris}.

\bibitem[Arora \& Zanette(2025)Arora and Zanette]{arora2025TLMRE}
Daman Arora and Andrea Zanette.
\newblock Training language models to reason efficiently.
\newblock \emph{arXiv preprint arXiv:2502.04463}, 2025.

\bibitem[{ByteDance-Seed}(2025)]{bytedance_seed_2025_beyondaime}
{ByteDance-Seed}.
\newblock Beyondaime: Advancing math reasoning evaluation beyond high school
  olympiads.
\newblock \url{https://huggingface.co/datasets/ByteDance-Seed/BeyondAIME},
  2025.
\newblock Hugging Face repository.

\bibitem[Chen et~al.(2025{\natexlab{a}})Chen, Zhang, Hong, Kundu, and
  Wang]{chen2025seal}
Runjin Chen, Zhenyu Zhang, Junyuan Hong, Souvik Kundu, and Zhangyang Wang.
\newblock Seal: Steerable reasoning calibration of large language models for
  free.
\newblock \emph{arXiv preprint arXiv:2504.07986}, 2025{\natexlab{a}}.

\bibitem[Chen et~al.(2025{\natexlab{b}})Chen, Xu, Liu, Li, Song, Jaiswal,
  Zhang, Hu, Wang, Chen, et~al.]{chen2025gpas}
Tianhao Chen, Xin Xu, Zijing Liu, Pengxiang Li, Xinyuan Song, Ajay~Kumar
  Jaiswal, Fan Zhang, Jishan Hu, Yang Wang, Hao Chen, et~al.
\newblock Gpas: Accelerating convergence of llm pretraining via
  gradient-preserving activation scaling.
\newblock \emph{arXiv preprint arXiv:2506.22049}, 2025{\natexlab{b}}.

\bibitem[Chen et~al.(2024)Chen, Xu, Liang, He, Pang, Yu, Song, Liu, Zhou,
  Zhang, et~al.]{chen2024overthinking}
Xingyu Chen, Jiahao Xu, Tian Liang, Zhiwei He, Jianhui Pang, Dian Yu, Linfeng
  Song, Qiuzhi Liu, Mengfei Zhou, Zhuosheng Zhang, et~al.
\newblock Do not think that much for 2+ 3=? on the overthinking of o1-like
  llms.
\newblock \emph{arXiv preprint arXiv:2412.21187}, 2024.

\bibitem[Chen et~al.(2025{\natexlab{c}})Chen, Ma, Fang, Yu, and
  Wang]{chen2025verithinker}
Zigeng Chen, Xinyin Ma, Gongfan Fang, Ruonan Yu, and Xinchao Wang.
\newblock Verithinker: Learning to verify makes reasoning model efficient.
\newblock \emph{arXiv preprint arXiv:2505.17941}, 2025{\natexlab{c}}.

\bibitem[Fang et~al.(2025)Fang, Ma, and Wang]{fang2025thinkless}
Gongfan Fang, Xinyin Ma, and Xinchao Wang.
\newblock Thinkless: Llm learns when to think.
\newblock \emph{arXiv preprint arXiv:2505.13379}, 2025.

\bibitem[Feng et~al.(2025)Feng, Huang, Qu, Zhang, Qin, Zhong, Jiang, Chi, and
  Zhong]{feng2025retool}
Jiazhan Feng, Shijue Huang, Xingwei Qu, Ge~Zhang, Yujia Qin, Baoquan Zhong,
  Chengquan Jiang, Jinxin Chi, and Wanjun Zhong.
\newblock Retool: Reinforcement learning for strategic tool use in llms.
\newblock \emph{arXiv preprint arXiv:2504.11536}, 2025.

\bibitem[Fu et~al.(2025{\natexlab{a}})Fu, Chen, Zhuang, Fu, Stoica, and
  Zhang]{fu2025dynasor}
Yichao Fu, Junda Chen, Yonghao Zhuang, Zheyu Fu, Ion Stoica, and Hao Zhang.
\newblock Reasoning without self-doubt: More efficient chain-of-thought through
  certainty probing.
\newblock In \emph{ICLR 2025 Workshop on Foundation Models in the Wild},
  2025{\natexlab{a}}.

\bibitem[Fu et~al.(2025{\natexlab{b}})Fu, Wang, Tian, and Zhao]{fu2025deepconf}
Yichao Fu, Xuewei Wang, Yuandong Tian, and Jiawei Zhao.
\newblock Deep think with confidence.
\newblock \emph{arXiv preprint arXiv:2508.15260}, 2025{\natexlab{b}}.

\bibitem[Guo et~al.(2025)Guo, Yang, Zhang, Song, Zhang, Xu, Zhu, Ma, Wang, Bi,
  et~al.]{guo2025deepseekr1}
Daya Guo, Dejian Yang, Haowei Zhang, Junxiao Song, Ruoyu Zhang, Runxin Xu,
  Qihao Zhu, Shirong Ma, Peiyi Wang, Xiao Bi, et~al.
\newblock Deepseek-r1: Incentivizing reasoning capability in llms via
  reinforcement learning.
\newblock \emph{arXiv preprint arXiv:2501.12948}, 2025.

\bibitem[Hammoud et~al.(2025)Hammoud, Alhamoud, Hammoud, Bou-Zeid, Ghassemi,
  and Ghanem]{hammoud2025curriculum-grpo}
Hasan Abed Al~Kader Hammoud, Kumail Alhamoud, Abed Hammoud, Elie Bou-Zeid,
  Marzyeh Ghassemi, and Bernard Ghanem.
\newblock Train long, think short: Curriculum learning for efficient reasoning.
\newblock \emph{arXiv preprint arXiv:2508.08940}, 2025.

\bibitem[He et~al.(2025)He, Liu, Liu, Yan, Wang, Cheng, Zhang, Zhang, Xu, Shen,
  et~al.]{he2025skyworkreasoner}
Jujie He, Jiacai Liu, Chris~Yuhao Liu, Rui Yan, Chaojie Wang, Peng Cheng,
  Xiaoyu Zhang, Fuxiang Zhang, Jiacheng Xu, Wei Shen, et~al.
\newblock Skywork open reasoner 1 technical report.
\newblock \emph{arXiv preprint arXiv:2505.22312}, 2025.

\bibitem[Hu et~al.(2025)Hu, Petty, Shi, Merrill, and
  Linzen]{hu2025prepretaining}
Michael~Y Hu, Jackson Petty, Chuan Shi, William Merrill, and Tal Linzen.
\newblock Between circuits and chomsky: Pre-pretraining on formal languages
  imparts linguistic biases.
\newblock \emph{arXiv preprint arXiv:2502.19249}, 2025.

\bibitem[Huang et~al.(2024)Huang, Wang, Xia, Li, Zou, Xu, Fan, Ye, Chern, Ye,
  et~al.]{huang2024olympicarena}
Zhen Huang, Zengzhi Wang, Shijie Xia, Xuefeng Li, Haoyang Zou, Ruijie Xu,
  Run-Ze Fan, Lyumanshan Ye, Ethan Chern, Yixin Ye, et~al.
\newblock Olympicarena: Benchmarking multi-discipline cognitive reasoning for
  superintelligent ai.
\newblock \emph{Advances in Neural Information Processing Systems},
  37:\penalty0 19209--19253, 2024.

\bibitem[Jaech et~al.(2024)Jaech, Kalai, Lerer, Richardson, El-Kishky, Low,
  Helyar, Madry, Beutel, Carney, et~al.]{jaech2024openaio1}
Aaron Jaech, Adam Kalai, Adam Lerer, Adam Richardson, Ahmed El-Kishky, Aiden
  Low, Alec Helyar, Aleksander Madry, Alex Beutel, Alex Carney, et~al.
\newblock Openai o1 system card.
\newblock \emph{arXiv preprint arXiv:2412.16720}, 2024.

\bibitem[Jain et~al.(2024)Jain, Han, Gu, Li, Yan, Zhang, Wang, Solar-Lezama,
  Sen, and Stoica]{jain2024livecodebench}
Naman Jain, King Han, Alex Gu, Wen-Ding Li, Fanjia Yan, Tianjun Zhang, Sida
  Wang, Armando Solar-Lezama, Koushik Sen, and Ion Stoica.
\newblock Livecodebench: Holistic and contamination free evaluation of large
  language models for code.
\newblock \emph{arXiv preprint arXiv:2403.07974}, 2024.

\bibitem[Jiang et~al.(2025)Jiang, Wu, Huang, Dong, Chi, Dong, Zhang, Lv, Cui,
  and Wei]{jiang2025LHRMs}
Lingjie Jiang, Xun Wu, Shaohan Huang, Qingxiu Dong, Zewen Chi, Li~Dong,
  Xingxing Zhang, Tengchao Lv, Lei Cui, and Furu Wei.
\newblock Think only when you need with large hybrid-reasoning models.
\newblock \emph{arXiv preprint arXiv:2505.14631}, 2025.

\bibitem[Jin et~al.(2025)Jin, Zeng, Yue, Yoon, Arik, Wang, Zamani, and
  Han]{jin2025searchr1}
Bowen Jin, Hansi Zeng, Zhenrui Yue, Jinsung Yoon, Sercan Arik, Dong Wang, Hamed
  Zamani, and Jiawei Han.
\newblock Search-r1: Training llms to reason and leverage search engines with
  reinforcement learning.
\newblock \emph{arXiv preprint arXiv:2503.09516}, 2025.

\bibitem[Kang et~al.(2025)Kang, Sun, Chen, and Zou]{kang2025c3ot}
Yu~Kang, Xianghui Sun, Liangyu Chen, and Wei Zou.
\newblock C3ot: Generating shorter chain-of-thought without compromising
  effectiveness.
\newblock In \emph{Proceedings of the AAAI Conference on Artificial
  Intelligence}, volume~39, pp.\  24312--24320, 2025.

\bibitem[Kwon et~al.(2023)Kwon, Li, Zhuang, Sheng, Zheng, Yu, Gonzalez, Zhang,
  and Stoica]{kwon2023vllm}
Woosuk Kwon, Zhuohan Li, Siyuan Zhuang, Ying Sheng, Lianmin Zheng, Cody~Hao Yu,
  Joseph~E. Gonzalez, Hao Zhang, and Ion Stoica.
\newblock Efficient memory management for large language model serving with
  pagedattention.
\newblock In \emph{Proceedings of the ACM SIGOPS 29th Symposium on Operating
  Systems Principles}, 2023.

\bibitem[Li et~al.(2025{\natexlab{a}})Li, Zou, and Liu]{li2025torl}
Xuefeng Li, Haoyang Zou, and Pengfei Liu.
\newblock Torl: Scaling tool-integrated rl.
\newblock \emph{arXiv preprint arXiv:2503.23383}, 2025{\natexlab{a}}.

\bibitem[Li et~al.(2025{\natexlab{b}})Li, Dong, Ma, Zhang, and
  Sui]{li2025selfbudgeter}
Zheng Li, Qingxiu Dong, Jingyuan Ma, Di~Zhang, and Zhifang Sui.
\newblock Selfbudgeter: Adaptive token allocation for efficient llm reasoning.
\newblock \emph{arXiv preprint arXiv:2505.11274}, 2025{\natexlab{b}}.

\bibitem[Liu et~al.(2025{\natexlab{a}})Liu, Zhou, Deng, Huang, Liu, Deng,
  Zhang, and He]{liu2025Laser}
Wei Liu, Ruochen Zhou, Yiyun Deng, Yuzhen Huang, Junteng Liu, Yuntian Deng,
  Yizhe Zhang, and Junxian He.
\newblock Learn to reason efficiently with adaptive length-based reward
  shaping.
\newblock \emph{arXiv preprint arXiv:2505.15612}, 2025{\natexlab{a}}.

\bibitem[Liu et~al.(2025{\natexlab{b}})Liu, Chen, Li, Qi, Pang, Du, Lee, and
  Lin]{liu2025drgrpo}
Zichen Liu, Changyu Chen, Wenjun Li, Penghui Qi, Tianyu Pang, Chao Du, Wee~Sun
  Lee, and Min Lin.
\newblock Understanding r1-zero-like training: A critical perspective.
\newblock \emph{arXiv preprint arXiv:2503.20783}, 2025{\natexlab{b}}.

\bibitem[Lou et~al.(2025)Lou, Sun, Liang, Qu, Shen, Wang, Li, Yang, and
  Wu]{lou2025adacot}
Chenwei Lou, Zewei Sun, Xinnian Liang, Meng Qu, Wei Shen, Wenqi Wang, Yuntao
  Li, Qingping Yang, and Shuangzhi Wu.
\newblock Adacot: Pareto-optimal adaptive chain-of-thought triggering via
  reinforcement learning.
\newblock \emph{arXiv preprint arXiv:2505.11896}, 2025.

\bibitem[Luo et~al.(2025{\natexlab{a}})Luo, Shen, He, Wang, Liu, Li, Tan, Cao,
  and Tao]{luo2025o1-pruner}
Haotian Luo, Li~Shen, Haiying He, Yibo Wang, Shiwei Liu, Wei Li, Naiqiang Tan,
  Xiaochun Cao, and Dacheng Tao.
\newblock O1-pruner: Length-harmonizing fine-tuning for o1-like reasoning
  pruning.
\newblock \emph{arXiv preprint arXiv:2501.12570}, 2025{\natexlab{a}}.

\bibitem[Luo et~al.(2025{\natexlab{b}})Luo, Tan, Wong, Shi, Tang, Roongta, Cai,
  Luo, Li, Popa, and Stoica]{deepscaler2025}
Michael Luo, Sijun Tan, Justin Wong, Xiaoxiang Shi, William~Y. Tang, Manan
  Roongta, Colin Cai, Jeffrey Luo, Li~Erran Li, Raluca~Ada Popa, and Ion
  Stoica.
\newblock Deepscaler: Surpassing o1-preview with a 1.5b model by scaling rl,
  2025{\natexlab{b}}.
\newblock Notion Blog.

\bibitem[Ma et~al.(2025)Ma, Wan, Yu, Fang, and Wang]{ma2025cot-value}
Xinyin Ma, Guangnian Wan, Runpeng Yu, Gongfan Fang, and Xinchao Wang.
\newblock Cot-valve: Length-compressible chain-of-thought tuning.
\newblock \emph{arXiv preprint arXiv:2502.09601}, 2025.

\bibitem[Meng et~al.(2025)Meng, Du, Liu, Zhou, Lu, Fu, Han, Shi, Wang, He,
  et~al.]{meng2025mmeureka}
Fanqing Meng, Lingxiao Du, Zongkai Liu, Zhixiang Zhou, Quanfeng Lu, Daocheng
  Fu, Tiancheng Han, Botian Shi, Wenhai Wang, Junjun He, et~al.
\newblock Mm-eureka: Exploring the frontiers of multimodal reasoning with
  rule-based reinforcement learning.
\newblock \emph{arXiv preprint arXiv:2503.07365}, 2025.

\bibitem[Morris et~al.(2023)Morris, Sohl-Dickstein, Fiedel, Warkentin, Dafoe,
  Faust, Farabet, and Legg]{morris2023levels}
Meredith~Ringel Morris, Jascha Sohl-Dickstein, Noah Fiedel, Tris Warkentin,
  Allan Dafoe, Aleksandra Faust, Clement Farabet, and Shane Legg.
\newblock Levels of agi for operationalizing progress on the path to agi.
\newblock \emph{arXiv preprint arXiv:2311.02462}, 2023.

\bibitem[Muennighoff et~al.(2025)Muennighoff, Yang, Shi, Li, Fei-Fei,
  Hajishirzi, Zettlemoyer, Liang, Cand{\`e}s, and Hashimoto]{muennighoff2025s1}
Niklas Muennighoff, Zitong Yang, Weijia Shi, Xiang~Lisa Li, Li~Fei-Fei,
  Hannaneh Hajishirzi, Luke Zettlemoyer, Percy Liang, Emmanuel Cand{\`e}s, and
  Tatsunori Hashimoto.
\newblock s1: Simple test-time scaling.
\newblock \emph{arXiv preprint arXiv:2501.19393}, 2025.

\bibitem[Munkhbat et~al.(2025)Munkhbat, Ho, Kim, Yang, Kim, and
  Yun]{munkhbat2025self-train-efficient}
Tergel Munkhbat, Namgyu Ho, Seo~Hyun Kim, Yongjin Yang, Yujin Kim, and Se-Young
  Yun.
\newblock Self-training elicits concise reasoning in large language models.
\newblock \emph{arXiv preprint arXiv:2502.20122}, 2025.

\bibitem[Phan et~al.(2025)Phan, Gatti, Han, Li, Hu, Zhang, Shi, Choi, Agrawal,
  Chopra, et~al.]{phan2025humanity}
Long Phan, Alice Gatti, Ziwen Han, Nathaniel Li, Josephina Hu, Hugh Zhang, Sean
  Shi, Michael Choi, Anish Agrawal, Arnav Chopra, et~al.
\newblock Humanity's last exam.
\newblock \emph{ArXiv preprint}, abs/2501.14249, 2025.
\newblock URL \url{https://arxiv.org/abs/2501.14249}.

\bibitem[Rein et~al.(2024)Rein, Hou, Stickland, Petty, Pang, Dirani, Michael,
  and Bowman]{rein2024gpqa}
David Rein, Betty~Li Hou, Asa~Cooper Stickland, Jackson Petty, Richard~Yuanzhe
  Pang, Julien Dirani, Julian Michael, and Samuel~R Bowman.
\newblock Gpqa: A graduate-level google-proof q\&a benchmark.
\newblock In \emph{First Conference on Language Modeling}, 2024.

\bibitem[Schulman et~al.(2017)Schulman, Wolski, Dhariwal, Radford, and
  Klimov]{schulman2017ppo}
John Schulman, Filip Wolski, Prafulla Dhariwal, Alec Radford, and Oleg Klimov.
\newblock Proximal policy optimization algorithms.
\newblock \emph{arXiv preprint arXiv:1707.06347}, 2017.

\bibitem[Seed et~al.(2025)Seed, Chen, Fan, Liu, Liu, Lin, Wang, Wang, Wei, Xu,
  et~al.]{seed2025seed1.6thinking}
ByteDance Seed, Jiaze Chen, Tiantian Fan, Xin Liu, Lingjun Liu, Zhiqi Lin,
  Mingxuan Wang, Chengyi Wang, Xiangpeng Wei, Wenyuan Xu, et~al.
\newblock Seed1. 5-thinking: Advancing superb reasoning models with
  reinforcement learning.
\newblock \emph{arXiv preprint arXiv:2504.13914}, 2025.

\bibitem[Setlur et~al.(2025)Setlur, Yang, Snell, Greer, Wu, Smith, Simchowitz,
  and Kumar]{setlur2025e3}
Amrith Setlur, Matthew~YR Yang, Charlie Snell, Jeremy Greer, Ian Wu, Virginia
  Smith, Max Simchowitz, and Aviral Kumar.
\newblock e3: Learning to explore enables extrapolation of test-time compute
  for llms.
\newblock \emph{arXiv preprint arXiv:2506.09026}, 2025.

\bibitem[Shao et~al.(2024)Shao, Wang, Zhu, Xu, Song, Bi, Zhang, Zhang, Li, Wu,
  et~al.]{shao2024grpo}
Zhihong Shao, Peiyi Wang, Qihao Zhu, Runxin Xu, Junxiao Song, Xiao Bi, Haowei
  Zhang, Mingchuan Zhang, YK~Li, Yang Wu, et~al.
\newblock Deepseekmath: Pushing the limits of mathematical reasoning in open
  language models.
\newblock \emph{arXiv preprint arXiv:2402.03300}, 2024.

\bibitem[Sheng et~al.(2024)Sheng, Zhang, Ye, Wu, Zhang, Zhang, Peng, Lin, and
  Wu]{sheng2024verl}
Guangming Sheng, Chi Zhang, Zilingfeng Ye, Xibin Wu, Wang Zhang, Ru~Zhang,
  Yanghua Peng, Haibin Lin, and Chuan Wu.
\newblock Hybridflow: A flexible and efficient rlhf framework.
\newblock \emph{arXiv preprint arXiv: 2409.19256}, 2024.

\bibitem[Shrivastava et~al.(2025)Shrivastava, Awadallah, Balachandran, Garg,
  Behl, and Papailiopoulos]{shrivastava2025gfpo}
Vaishnavi Shrivastava, Ahmed Awadallah, Vidhisha Balachandran, Shivam Garg,
  Harkirat Behl, and Dimitris Papailiopoulos.
\newblock Sample more to think less: Group filtered policy optimization for
  concise reasoning.
\newblock \emph{arXiv preprint arXiv:2508.09726}, 2025.

\bibitem[Sui et~al.(2025)Sui, Chuang, Wang, Zhang, Zhang, Yuan, Liu, Wen,
  Zhong, Chen, et~al.]{sui2025overthinkingsurvey}
Yang Sui, Yu-Neng Chuang, Guanchu Wang, Jiamu Zhang, Tianyi Zhang, Jiayi Yuan,
  Hongyi Liu, Andrew Wen, Shaochen Zhong, Hanjie Chen, et~al.
\newblock Stop overthinking: A survey on efficient reasoning for large language
  models.
\newblock \emph{arXiv preprint arXiv:2503.16419}, 2025.

\bibitem[Team et~al.(2025{\natexlab{a}})Team, Bai, Bao, Chen, Chen, Chen, Chen,
  Chen, Chen, Chen, et~al.]{team2025kimik2}
Kimi Team, Yifan Bai, Yiping Bao, Guanduo Chen, Jiahao Chen, Ningxin Chen,
  Ruijue Chen, Yanru Chen, Yuankun Chen, Yutian Chen, et~al.
\newblock Kimi k2: Open agentic intelligence.
\newblock \emph{arXiv preprint arXiv:2507.20534}, 2025{\natexlab{a}}.

\bibitem[Team et~al.(2025{\natexlab{b}})Team, Du, Gao, Xing, Jiang, Chen, Li,
  Xiao, Du, Liao, et~al.]{team2025kimik1.5}
Kimi Team, Angang Du, Bofei Gao, Bowei Xing, Changjiu Jiang, Cheng Chen, Cheng
  Li, Chenjun Xiao, Chenzhuang Du, Chonghua Liao, et~al.
\newblock Kimi k1. 5: Scaling reinforcement learning with llms.
\newblock \emph{arXiv preprint arXiv:2501.12599}, 2025{\natexlab{b}}.

\bibitem[Tong et~al.(2024)Tong, Zhang, Wang, Wu, and He]{dartmath2024tong}
Yuxuan Tong, Xiwen Zhang, Rui Wang, Ruidong Wu, and Junxian He.
\newblock Dart-math: Difficulty-aware rejection tuning for mathematical
  problem-solving.
\newblock \emph{ArXiv preprint}, abs/2407.13690, 2024.
\newblock URL \url{https://arxiv.org/abs/2407.13690}.

\bibitem[Tu et~al.(2025)Tu, Lin, Zhang, Tian, Li, Lan, and
  Zhao]{tu2025AutoThink}
Songjun Tu, Jiahao Lin, Qichao Zhang, Xiangyu Tian, Linjing Li, Xiangyuan Lan,
  and Dongbin Zhao.
\newblock Learning when to think: Shaping adaptive reasoning in r1-style models
  via multi-stage rl.
\newblock \emph{arXiv preprint arXiv:2505.10832}, 2025.

\bibitem[Wang et~al.(2025{\natexlab{a}})Wang, Yu, Gao, Zheng, Liu, Lu, Dang,
  Chen, Yang, Zhang, et~al.]{wang2025beyond8020}
Shenzhi Wang, Le~Yu, Chang Gao, Chujie Zheng, Shixuan Liu, Rui Lu, Kai Dang,
  Xionghui Chen, Jianxin Yang, Zhenru Zhang, et~al.
\newblock Beyond the 80/20 rule: High-entropy minority tokens drive effective
  reinforcement learning for llm reasoning.
\newblock \emph{arXiv preprint arXiv:2506.01939}, 2025{\natexlab{a}}.

\bibitem[Wang et~al.(2025{\natexlab{b}})Wang, Zhou, Li, and
  Liu]{wang2025octothinker}
Zengzhi Wang, Fan Zhou, Xuefeng Li, and Pengfei Liu.
\newblock Octothinker: Mid-training incentivizes reinforcement learning
  scaling.
\newblock \emph{arXiv preprint arXiv:2506.20512}, 2025{\natexlab{b}}.

\bibitem[Wang et~al.(2025{\natexlab{c}})Wang, Guo, Stoica, Xu, Wang, Ha, Chen,
  Chen, Yan, Huang, et~al.]{wang2025papo}
Zhenhailong Wang, Xuehang Guo, Sofia Stoica, Haiyang Xu, Hongru Wang,
  Hyeonjeong Ha, Xiusi Chen, Yangyi Chen, Ming Yan, Fei Huang, et~al.
\newblock Perception-aware policy optimization for multimodal reasoning.
\newblock \emph{arXiv preprint arXiv:2507.06448}, 2025{\natexlab{c}}.

\bibitem[Wei et~al.(2022)Wei, Wang, Schuurmans, Bosma, Ichter, Xia, Chi, Le,
  and Zhou]{CoT2022Wei}
Jason Wei, Xuezhi Wang, Dale Schuurmans, Maarten Bosma, Brian Ichter, Fei Xia,
  Ed~H. Chi, Quoc~V. Le, and Denny Zhou.
\newblock Chain-of-thought prompting elicits reasoning in large language
  models.
\newblock In Sanmi Koyejo, S.~Mohamed, A.~Agarwal, Danielle Belgrave, K.~Cho,
  and A.~Oh (eds.), \emph{Advances in Neural Information Processing Systems 35:
  Annual Conference on Neural Information Processing Systems 2022, NeurIPS
  2022, New Orleans, LA, USA, November 28 - December 9, 2022}, 2022.
\newblock URL
  \url{http://papers.nips.cc/paper\_files/paper/2022/hash/9d5609613524ecf4f15af0f7b31abca4-Abstract-Conference.html}.

\bibitem[Wen et~al.(2025)Wen, Wu, Sun, Zhang, Chen, Wang, Liu, Zhang, and
  Li]{wen2025budgetthinker}
Hao Wen, Xinrui Wu, Yi~Sun, Feifei Zhang, Liye Chen, Jie Wang, Yunxin Liu,
  Ya-Qin Zhang, and Yuanchun Li.
\newblock Budgetthinker: Empowering budget-aware llm reasoning with control
  tokens.
\newblock \emph{arXiv preprint arXiv:2508.17196}, 2025.

\bibitem[Xiao et~al.(2025)Xiao, Xu, Huang, Gao, Liu, Liu, and
  Chen]{xiao2025perception}
Tong Xiao, Xin Xu, Zhenya Huang, Hongyu Gao, Quan Liu, Qi~Liu, and Enhong Chen.
\newblock Advancing multimodal reasoning capabilities of multimodal large
  language models via visual perception reward.
\newblock \emph{arXiv preprint arXiv:2506.07218}, 2025.

\bibitem[Xu et~al.(2024)Xu, Xiao, Chao, Huang, Yang, and Wang]{xu2024egsm}
Xin Xu, Tong Xiao, Zitong Chao, Zhenya Huang, Can Yang, and Yang Wang.
\newblock Can llms solve longer math word problems better?
\newblock \emph{ArXiv preprint}, abs/2405.14804, 2024.
\newblock URL \url{https://arxiv.org/abs/2405.14804}.

\bibitem[Xu et~al.(2025{\natexlab{a}})Xu, Chen, Zhang, Liu, Li, Jaiswal, Yan,
  Hu, Wang, Chen, et~al.]{xu2025doublechecker}
Xin Xu, Tianhao Chen, Fan Zhang, Wanlong Liu, Pengxiang Li, Ajay~Kumar Jaiswal,
  Yuchen Yan, Jishan Hu, Yang Wang, Hao Chen, et~al.
\newblock Double-checker: Enhancing reasoning of slow-thinking llms via
  self-critical fine-tuning.
\newblock \emph{arXiv preprint arXiv:2506.21285}, 2025{\natexlab{a}}.

\bibitem[Xu et~al.(2025{\natexlab{b}})Xu, Xu, Xiao, Chen, Yan, Zhang, Diao,
  Yang, and Wang]{xu2025ugphysics}
Xin Xu, Qiyun Xu, Tong Xiao, Tianhao Chen, Yuchen Yan, Jiaxin Zhang, Shizhe
  Diao, Can Yang, and Yang Wang.
\newblock Ugphysics: A comprehensive benchmark for undergraduate physics
  reasoning with large language models.
\newblock \emph{arXiv preprint arXiv:2502.00334}, 2025{\natexlab{b}}.

\bibitem[Xu et~al.(2025{\natexlab{c}})Xu, Zhang, Chen, Chao, Hu, and
  Yang]{xu2025ugmathbench}
Xin Xu, Jiaxin Zhang, Tianhao Chen, Zitong Chao, Jishan Hu, and Can Yang.
\newblock Ugmathbench: A diverse and dynamic benchmark for undergraduate-level
  mathematical reasoning with large language models.
\newblock \emph{arXiv preprint arXiv:2501.13766}, 2025{\natexlab{c}}.

\bibitem[Xue et~al.(2025)Xue, Zheng, Liu, Li, Ma, and An]{xue2025simpletir}
Zhenghai Xue, Longtao Zheng, Qian Liu, Yingru Li, Zejun Ma, and Bo~An.
\newblock Simpletir: End-to-end reinforcement learning for multi-turn
  tool-integrated reasoning.
\newblock \url{https://simpletir.notion.site/report}, 2025.
\newblock Notion Blog.

\bibitem[Yang et~al.(2024)Yang, Zhang, Hui, Gao, Yu, Li, Liu, Tu, Zhou, Lin,
  et~al.]{yang2024qwen25math}
An~Yang, Beichen Zhang, Binyuan Hui, Bofei Gao, Bowen Yu, Chengpeng Li,
  Dayiheng Liu, Jianhong Tu, Jingren Zhou, Junyang Lin, et~al.
\newblock Qwen2. 5-math technical report: Toward mathematical expert model via
  self-improvement.
\newblock \emph{arXiv preprint arXiv:2409.12122}, 2024.

\bibitem[Yang et~al.(2025{\natexlab{a}})Yang, Li, Yang, Zhang, Hui, Zheng, Yu,
  Gao, Huang, Lv, et~al.]{yang2025qwen3}
An~Yang, Anfeng Li, Baosong Yang, Beichen Zhang, Binyuan Hui, Bo~Zheng, Bowen
  Yu, Chang Gao, Chengen Huang, Chenxu Lv, et~al.
\newblock Qwen3 technical report.
\newblock \emph{arXiv preprint arXiv:2505.09388}, 2025{\natexlab{a}}.

\bibitem[Yang et~al.(2025{\natexlab{b}})Yang, Si, Dai, Yao, Zheng, Chen, Lin,
  and Wang]{yang2025pi}
Chenxu Yang, Qingyi Si, Mz~Dai, Dingyu Yao, Mingyu Zheng, Minghui Chen, Zheng
  Lin, and Weiping Wang.
\newblock Test-time prompt intervention.
\newblock \emph{arXiv preprint arXiv:2508.02511}, 2025{\natexlab{b}}.

\bibitem[Yang et~al.(2025{\natexlab{c}})Yang, Si, Duan, Zhu, Zhu, Li, Lin, Cao,
  and Wang]{yang2025deer}
Chenxu Yang, Qingyi Si, Yongjie Duan, Zheliang Zhu, Chenyu Zhu, Qiaowei Li,
  Zheng Lin, Li~Cao, and Weiping Wang.
\newblock Dynamic early exit in reasoning models.
\newblock \emph{arXiv preprint arXiv:2504.15895}, 2025{\natexlab{c}}.

\bibitem[Ye et~al.(2025)Ye, Huang, Xiao, Chern, Xia, and Liu]{ye2025limo}
Yixin Ye, Zhen Huang, Yang Xiao, Ethan Chern, Shijie Xia, and Pengfei Liu.
\newblock Limo: Less is more for reasoning.
\newblock \emph{arXiv preprint arXiv:2502.03387}, 2025.

\bibitem[Yu et~al.(2023)Yu, Jiang, Shi, Yu, Liu, Zhang, Kwok, Li, Weller, and
  Liu]{metamath2023yu}
Longhui Yu, Weisen Jiang, Han Shi, Jincheng Yu, Zhengying Liu, Yu~Zhang,
  James~T Kwok, Zhenguo Li, Adrian Weller, and Weiyang Liu.
\newblock Metamath: Bootstrap your own mathematical questions for large
  language models.
\newblock \emph{ArXiv preprint}, abs/2309.12284, 2023.
\newblock URL \url{https://arxiv.org/abs/2309.12284}.

\bibitem[Yu et~al.(2025)Yu, Zhang, Zhu, Yuan, Zuo, Yue, Dai, Fan, Liu, Liu,
  et~al.]{yu2025dapo}
Qiying Yu, Zheng Zhang, Ruofei Zhu, Yufeng Yuan, Xiaochen Zuo, Yu~Yue, Weinan
  Dai, Tiantian Fan, Gaohong Liu, Lingjun Liu, et~al.
\newblock Dapo: An open-source llm reinforcement learning system at scale.
\newblock \emph{arXiv preprint arXiv:2503.14476}, 2025.

\bibitem[Zeng et~al.(2025{\natexlab{a}})Zeng, Lv, Zheng, Hou, Chen, Xie, Wang,
  Yin, Zeng, Zhang, et~al.]{zeng2025glm4_5}
Aohan Zeng, Xin Lv, Qinkai Zheng, Zhenyu Hou, Bin Chen, Chengxing Xie, Cunxiang
  Wang, Da~Yin, Hao Zeng, Jiajie Zhang, et~al.
\newblock Glm-4.5: Agentic, reasoning, and coding (arc) foundation models.
\newblock \emph{arXiv preprint arXiv:2508.06471}, 2025{\natexlab{a}}.

\bibitem[Zeng et~al.(2025{\natexlab{b}})Zeng, Huang, Liu, Liu, He, Ma, and
  He]{zeng2025simplerl}
Weihao Zeng, Yuzhen Huang, Qian Liu, Wei Liu, Keqing He, Zejun Ma, and Junxian
  He.
\newblock Simplerl-zoo: Investigating and taming zero reinforcement learning
  for open base models in the wild.
\newblock \emph{arXiv preprint arXiv:2503.18892}, 2025{\natexlab{b}}.

\bibitem[Zhang et~al.(2025{\natexlab{a}})Zhang, Xu, Wang, Cui, Liu, and
  An]{zhang2025verifyrl}
Fuxiang Zhang, Jiacheng Xu, Chaojie Wang, Ce~Cui, Yang Liu, and Bo~An.
\newblock Incentivizing llms to self-verify their answers.
\newblock \emph{arXiv preprint arXiv:2506.01369}, 2025{\natexlab{a}}.

\bibitem[Zhang et~al.(2025{\natexlab{b}})Zhang, Lin, Hou, Feng, and
  Li]{zhang2025adaptthink}
Jiajie Zhang, Nianyi Lin, Lei Hou, Ling Feng, and Juanzi Li.
\newblock Adaptthink: Reasoning models can learn when to think.
\newblock \emph{arXiv preprint arXiv:2505.13417}, 2025{\natexlab{b}}.

\bibitem[Zhang et~al.(2025{\natexlab{c}})Zhang, Sun, Zhang, Feng, Lu, Yang, and
  Meng]{zhang2025critiquegrpo}
Xiaoying Zhang, Hao Sun, Yipeng Zhang, Kaituo Feng, Chaochao Lu, Chao Yang, and
  Helen Meng.
\newblock Critique-grpo: Advancing llm reasoning with natural language and
  numerical feedback.
\newblock \emph{arXiv preprint arXiv:2506.03106}, 2025{\natexlab{c}}.

\bibitem[Zhang et~al.(2025{\natexlab{d}})Zhang, Ruan, Ma, Zhu, Zhao, Li, Chen,
  Zeng, and Cai]{zhang2025ASRR}
Xiaoyun Zhang, Jingqing Ruan, Xing Ma, Yawen Zhu, Haodong Zhao, Hao Li,
  Jiansong Chen, Ke~Zeng, and Xunliang Cai.
\newblock When to continue thinking: Adaptive thinking mode switching for
  efficient reasoning.
\newblock \emph{arXiv preprint arXiv:2505.15400}, 2025{\natexlab{d}}.

\bibitem[Zheng et~al.(2025)Zheng, Liu, Li, Chen, Yu, Gao, Dang, Liu, Men, Yang,
  et~al.]{zheng2025gspo}
Chujie Zheng, Shixuan Liu, Mingze Li, Xiong-Hui Chen, Bowen Yu, Chang Gao, Kai
  Dang, Yuqiong Liu, Rui Men, An~Yang, et~al.
\newblock Group sequence policy optimization.
\newblock \emph{arXiv preprint arXiv:2507.18071}, 2025.

\bibitem[Zhou et~al.(2023)Zhou, Lu, Mishra, Brahma, Basu, Luan, Zhou, and
  Hou]{zhou2023ifeval}
Jeffrey Zhou, Tianjian Lu, Swaroop Mishra, Siddhartha Brahma, Sujoy Basu,
  Yi~Luan, Denny Zhou, and Le~Hou.
\newblock Instruction-following evaluation for large language models.
\newblock \emph{arXiv preprint arXiv:2311.07911}, 2023.
\newblock \doi{10.48550/arXiv.2311.07911}.
\newblock URL \url{https://arxiv.org/abs/2311.07911}.

\end{thebibliography}
\bibliographystyle{iclr2026_conference}

\newpage
\appendix
\section{Background and Preliminary}

\subsection{\revise{RLVR Algorithms}}\label{app:preliminary}

\textbf{Proximal Policy Optimization (PPO)}.
PPO~\citep{schulman2017ppo} constrains the policy update within a proximal region of the old policy $\pi_{\theta_\text{old}}$ through the clipping mechanism.
Specifically, PPO employs the following objective (we omit the KL regularization term hereinafter for brevity):
{\small
\begin{align}
\mathcal{J}_\text{PPO}(\theta) = \mathbb{E}_{ x \sim \mathcal{D},\, y \sim \pi_{\theta_\text{old}}( \cdot | x) }
\left[ \frac{1}{|y|} \sum_{t=1}^{|y|} 
\min \left( r_{t}(\theta) \widehat{A}_{t},  \, \mathrm{clip} \left( r_{t}(\theta), 1 - {\varepsilon}, 1 + {\varepsilon}\right) \widehat{A}_{t} \right)
\right],
\end{align}
}with the importance ratio of the token $y_t$ is defined as
$
r_{t}(\theta) = \frac{ \pi_{\theta} (y_{t} | x, y_{<t}) }{ \pi_{\theta_\text{old}} (y_{t} | x,y_{<t})}
$,
the advantage $\widehat{A}_{t}$ of $y_t$ is estimated by a value model, and $\varepsilon$ is the clipping range of importance ratios.

\textbf{Group Relative Policy Optimization (GRPO)}.
GRPO~\citep{shao2024grpo} bypasses the need for the value model by computing the relative advantage of each response within a group of responses to the same query.
Specifically, GRPO optimizes $\mathcal{J}_\text{GRPO}(\theta) = \mathbb{E}_{ x \sim \mathcal{D}} \left[ \mathcal{J}_\text{GRPO}(\theta, x) \right],$ where:
{\small
\begin{align}
\label{equ:grpo}
\mathcal{J}_\text{GRPO}(\theta, x) = 
\left[ \frac{1}{G} \sum_{i=1}^{G} \frac{1}{|y_i|} \sum_{t=1}^{|y_i|} 
\min \left( r_{i,t}(\theta) \widehat{A}_{i,t},  \, \mathrm{clip} \left( r_{i,t}(\theta), 1 - {\varepsilon}, 1 + {\varepsilon}\right) \widehat{A}_{i,t} \right)
\right].
\end{align}
}Here ${\{y_i\}_{i=1}^G \sim \pi_{\theta_\text{old}}( \cdot | x) }$, $G$ is the number of generated responses to each query $x$ (i.e., the group size), and $r_{i,t}(\theta), \,\widehat{A}_{i,t}$ is computed as in \cref{eq: adv-ratio}.

Numerous variants have been proposed to improve GRPO.
For example, DAPO~\citep{yu2025dapo} introduces token-level normalization and dynamic sampling; Dr GRPO~\citep{liu2025drgrpo} removes length bias to prevent incorrect responses from growing longer over time; and \citet{wang2025beyond8020} train selectively on forking tokens (see also Section~\ref{sec:related_work}).
Our \method{} is orthogonal to these RLVR algorithms.
That is to say, any RLVR algorithm can be applied to our proposed {\method} stage.
\textbf{To isolate the effect of RLVR algorithms, we employ DAPO as our RLVR algorithm in all experiments for fair comparison.}

\subsection{\thinkfree{} Operation}\label{app:think-free operation}

In Section~\ref{subsec:thing-free reasoning}, we have shown chat templates for both the original query $x$ and its thinking-free version $x'=\thinkfree(x)$ for Qwen models.
Here, we showcase one additional example under the DeepSeek~\citep{guo2025deepseekr1} template as below.

\begin{tcolorbox}[colback=rliableblue!10!white,colframe=black,boxrule=1pt,boxsep=2pt,top=3pt,bottom=3pt,left=2pt,right=2pt]
\begin{template}[\textbf{\emph{Thinking Mode (DeepSeek)}}]
\label{template:thinking-ds}
Please reason step by step, and put your final answer within \textbackslash \textbackslash boxed\{\}.$<$\textbar User\textbar$>$\textbf{\{question (x)\}}$<$\textbar Assistant \textbar$>$\textbackslash n\looseness=-1
\end{template}
\vspace{0.05em}
\begin{template}[\textbf{\emph{Thinking-Free Mode (DeepSeek)}}]
\label{template:thinking free-ds}
Please reason step by step, and put your final answer within \textbackslash \textbackslash boxed\{\}.$<$\textbar User\textbar$>$\textbf{\{question (x)\}}$<$\textbar Assistant\textbar$>$\textbackslash n\looseness=-1 \textcolor{red}{\textless think\textgreater \textbackslash n\looseness=-1 \textbackslash n\looseness=-1\textless/think\textgreater}
\end{template}
\end{tcolorbox}

\section{Meta Experiments}\label{app:meta-exp}

\subsection{Token Consumption of \thinkfree{}}\label{app:meta-exp-token}

We provide the detailed experimental setup for the meta-experiment in Section~\ref{subsec:thing-free reasoning} in this appendix.

For \texttt{DS-1.5B}, we use the decoding parameters suggested by \citet{guo2025deepseekr1} in thinking mode: temperature = 0.6, top-$p$ = 0.95, and top-$k$ = $-1$.  
For \texttt{Qwen3-4B}, we adopt the recommended settings from \citet{yang2025qwen3} in thinking mode: temperature = 0.6, top-$p$ = 0.95, and top-$k$ = $20$.  
For \thinkfree{}, we set temperature = 0.7, top-$p$ = 0.8, and use both top-$k$ = $-1$ and top-$k$ = $20$.  
The maximum output length is fixed at 32K tokens for both modes.  
For evaluation, we sample 32 generations per query on AIME 2025 and report the average number of output tokens.
The parameters are given in Table~\ref{tabapp:meta-token-setup} and the results are shown in Figure~\ref{fig:meta-exp} (Left).

\begin{table}
    \centering
    \caption{Decoding Parameters of Meta-Experiment in Section~\ref{subsec:thing-free reasoning}}
    \begin{tabular}{l|cc}
    \toprule
         & Thinking Mode  & Thinking-Free Mode \\
    \midrule
        \texttt{DS-1.5B} & $T = 0.6$, top-$p$ = 0.95, and top-$k$ = $-1$ & $T = 0.7$, top-$p$ = 0.8, and top-$k$ = $20$ \\
        \texttt{Qwen3-4B} & $T = 0.6$, top-$p$ = 0.95, and top-$k$ = $20$ & $T = 0.7$, top-$p$ = 0.8, and top-$k$ = $20$ \\
    \bottomrule
    \end{tabular}
    \label{tabapp:meta-token-setup}
\end{table}

\subsection{Detailed Setup of \thinkfree{} Training}\label{app:meta-exp-train}

We provide the detailed experimental setup for the meta-experiment discussed in Section~\ref{subsec:thinking-free training} in this appendix. 

For training, we use the DAPO~\citep{yu2025dapo} algorithm, with configurations identical to those in Appendix~\ref{app:training}, except that the maximum output length is set to 4K. 
For evaluation, we set the maximum output length to 48K and perform testing in thinking mode as described in~\cref{template:thinking}. All other evaluation parameters follow Appendix~\ref{app:evaluation}.
Note that the initial \texttt{avg@32} value in Figure~\ref{fig:meta-exp} (Right) is higher than the value reported in~\citep{yang2025qwen3} (68.2 vs.\ 65.6), because we adopt the RoPE scaling method described in Polaris~\citep{Polaris2025}.

\section{Experimental Details}\label{app:exp-setup}

In this appendix, we provide the details of our main experiments in Section~\ref{sec:exp}.

\subsection{Training Details}\label{app:training}

We build on the VeRL codebase~\citep{sheng2024verl}, with the RLVR loss following the DAPO recipe~\citep{yu2025dapo}. Specifically, the RLVR loss is defined in \cref{eq: dapo objective}. For our {\method}, it becomes
\begin{equation}
\mathbb{E}_{x \sim \mathcal{D}} \left[ \mathcal{J}_{\text{DAPO}}(\theta, x') \right], 
\quad x' = \thinkfree(x).
\end{equation}

For fair comparison, we use identical hyperparameters across methods. For \textit{clip-higher}, we set $\varepsilon_{\text{low}} = 0.2$ and $\varepsilon_{\text{high}} = 0.28$. Training is performed with a batch size and mini-batch size of 256, a learning rate of $10^{-6}$, and no warm-up scheduling. Both KL divergence loss and entropy loss are excluded.

For rollout, we use temperature $= 1$, $\text{topp} = 1$, and $\text{topk} = -1$. We generate 8 rollouts per problem. Experiments are conducted on \texttt{DS-1.5B}, \texttt{Qwen3-4B}, and \texttt{DS-7B}, with Polaris-53K~\citep{Polaris2025} as the training dataset.
\textbf{In principle, we could apply dataset filtering at each training stage to accelerate training}~\citet{Polaris2025}. \textbf{However, for fairness, we deliberately use the full training set for all experiments.}

For Direct RLVR, the maximum output length is set to 16K for \texttt{DS-1.5B} and \texttt{DS-7B}, and 32K for \texttt{Qwen3-4B}. For {\method}, we adopt a multi-stage training strategy~\citep{Polaris2025,deepscaler2025}:
\begin{itemize}
    \item \texttt{DS-1.5B} and \texttt{DS-7B}: $2\text{K} \rightarrow 4\text{K} \rightarrow 8\text{K}$.
    \item \texttt{Qwen3-4B}: $4\text{K} \rightarrow 8\text{K} \rightarrow 16\text{K}$.
\end{itemize}

Our experiments are conducted with 32 H20 GPUs.
A summary of the number of training steps and training time is provided in Table~\ref{tabapp:train}.
\revise{To provide a comprehensive view of training compute, we also report the number of processed tokens and the average MFU in Table~\ref{tabapp:tokens}. 
Notably, our {method} achieves a higher MFU and therefore can generate more tokens than ``Direct RL'' within the same wall-clock time. 
This arises because shorter responses during RL training reduce rollout-stage ``bubble'', yielding a practical efficiency advantage.}

\begin{table}
    \centering
    \caption{Training Steps and Time of Main Experiments. ``kh'' denotes one thousand H20 Hours.}
    \resizebox{\linewidth}{!}{%
        \begin{tabular}{l|ccc|ccc}
            \toprule
             & \texttt{DS-1.5B} & \texttt{Qwen3-4B} & \texttt{DS-7B} & \texttt{DS-1.5B} & \texttt{Qwen3-4B} & \texttt{DS-7B} \\
             \midrule
            \method{} Stage 1 & 2K, 1K steps & 4K, 100 steps & 2K, 1K steps & 0.84 kh & 0.35 kh & 1.73 kh \\
            \method{} Stage 2 & 4K, 440 steps & 8K, 56 steps & 4K, 232 steps & 0.62 kh & 0.52 kh & 0.82 kh \\
            \method{} Stage 3 & 8K, 440 steps & 16K, 64 steps & 8K, 144 steps & 1.21 kh & 0.91 kh & 0.94 kh \\
            \method{} Total & - & - & - & 2.66 kh & 1.79 kh & 3.50 kh \\
            Direct RLVR & 16K, 456 steps & 32K, 20 steps & 16K, 280 steps & 2.67 kh & 1.9 kh & 3.52 kh \\
            \midrule
            \method{}: RLVR & 16K, 472 steps & 32K, 192 steps & 536 steps & 2.49 kh & 13.5 kh & 6.55 kh \\
            \method{} + RLVR Total  & - & - & - & 5.16 kh & 15.4 kh & 10 kh \\
            Direct RLVR Total& 16K, 896 steps & 32K, 216 steps  & 16K, 820 steps & 5.17 kh & 15.7 kh & 10.1 kh \\
             \bottomrule
        \end{tabular}
    }
    \label{tabapp:train}
\end{table}

\begin{table}
    \centering
    \caption{\revise{Total processed tokens and the average MFU (in \%) of Main Experiments.}}
    \resizebox{\linewidth}{!}{%
        \begin{tabular}{l|ccc|ccc|}
            \toprule
             & \texttt{DS-1.5B} & \texttt{Qwen3-4B} & \texttt{DS-7B} & \texttt{DS-1.5B} & \texttt{Qwen3-4B} & \texttt{DS-7B} \\
             \midrule
            \method{} Stage 1 & $2.37 \times 10^{9}$ & $5.00 \times 10^8$ & $2.75 \times 10^9$ & 69.27 & 66.28 & 73.73 \\
            \method{} Stage 2 & $2.26 \times 10^{9}$ & $5.02 \times 10^8$ & $1.20 \times 10^9$ & 69.54 & 66.67 & 73.79 \\
            \method{} Stage 3 & $3.92 \times 10^{9}$ & $4.94 \times 10^8$ & $1.24 \times 10^9$ & 68.66 &  66.76 & 73.76 \\
            \method{} Total & $8.55 \times 10^{9}$ & $ 1.50 \times 10^9$ & $5.19 \times 10^{9}$ & - & - & - \\
            Direct RLVR & $7.48 \times 10^9 $ & $6.48 \times 10^8$ & $4.77\times 10^9$ & 68.42 & 61.57 & 72.83\\
            \midrule
            \method{}: RLVR & $6.26 \times 10^9$ & $5.80 \times 10^9$ & $7.98 \times 10^9$ & 66.20 & 60.08 & 72.58 \\
            \method{} + RLVR Total & $1.48 \times 10^{10}$ & $7.30 \times 10^9$ & $1.27 \times 10^{10}$ & - & - & - \\
            Direct RLVR Total & $1.44 \times 10^{10}$ & $7.36 \times 10^9$ & $1.38 \times 10^{10}$ & 68.11 & 62.24 & 72.89 \\
             \bottomrule
        \end{tabular}
    }
    \label{tabapp:tokens}
\end{table}

\subsection{Baselines}\label{app:baselins}

To evaluate the efficacy of {\method}, we compare {\method} with direct RLVR training from an SFT-distilled LRM (``Direct RL'' for short) under the same total training compute, i.e., the combined compute of the three {\method} stages equals that of direct RLVR (Table~\ref{tab:main_result}).  
To examine the effect of inserting a {\method} stage before RLVR, we apply {\method} to an SFT-distilled model (``{\method} + RL''), continue with standard RLVR, and compare the results with ``Direct RL'' using approximately the same training compute (Table~\ref{tab:sota}).  
We also include several high-performing LRMs of the same model size from previous works for reference, including Polaris~\citep{Polaris2025}, DeepScaleR~\citep{deepscaler2025}, Skywork-OR1~\citep{he2025skyworkreasoner}, and AReal-RL.  
For both Table~\ref{tab:main_result} and Table~\ref{tab:sota}, all models are evaluated in thinking mode (see also Appendix~\ref{app:evaluation}).  
To assess the impact of {\method} on reasoning efficiency, we compare our {\method}-trained model with various RL-based efficient reasoning baselines, including TLMRE~\citep{arora2025TLMRE}, AdaptThink~\citep{zhang2025adaptthink}, AutoThink~\citep{tu2025AutoThink}, Laser~\citep{liu2025Laser}, L1Max~\citep{aggarwal2025L1}, and ThinkLess~\citep{fang2025thinkless} (Table~\ref{tab:results_tokens}).
The training and testing settings of these baselines, as reported in their original papers, are summarized in Table~\ref{tabapp:baselines}.  
\textbf{For fair comparison, we standardize the testing parameters to \texttt{topp} = 0.95, \texttt{topk} = $-1$, and $T$ = 0.6 with a 32K maximum length}, following the recommendations of DeepSeek-R1~\citep{guo2025deepseekr1}.

\begin{table}[ht!]
\centering
\caption{
Training and evaluation details of efficient reasoning baselines from original papers.
We unify the evaluation setting for fair comparison in Table~\ref{tab:results_tokens}.
Details are provided in Appendix~\ref{app:baselins} and~\ref{app:evaluation}.
}
\label{tabapp:baselines}
\resizebox{\linewidth}{!}{
\begin{tabular}{lcccc}
\toprule
\textbf{Model} & \textbf{Training Data} & \textbf{Training Length} & \textbf{Test Length} & \textbf{Evaluation Details} \\
\midrule
\multicolumn{5}{c}{\texttt{\textbf{DeepSeek-Distill-Qwen-1.5B}}} \\
\midrule
TLMRE-DS-1.5B ($\alpha = 0.1$) & 3.2K examples from NuminaMath & -- & 32K & 10 generations for AIME24 \\
AdaptThink-1.5B ($\delta = 0.05$) & DeepScaleR-40K & 16K & 16K & $T=0.6$, 16 generations for AIME24 \\
AutoThink-DS-1.5B-Stage3 & DeepScaleR-40K & 8K $\rightarrow$ 16K $\rightarrow$ 24K & -- & $T=0.6$, 16 generations \\
Laser-DE-L4096-1.5B & DeepScaleR-40K & -- & 32K & 16 generations for AIME24 \\
\midrule
\multicolumn{5}{c}{\texttt{\textbf{DeepscaleR-1.5B-Preview}}} \\
\midrule
AutoThink-Stage3 & DeepScaleR-40K & 8K $\rightarrow$ 16K $\rightarrow$ 24K & -- & $T=0.6$, 16 generations \\
L1-1.5B-Max & DeepScaleR-40K & 4K & 8K & -- \\
Thinkless-1.5B-RL & DeepScaleR-40K & 24K & -- & -- \\
\bottomrule
\end{tabular}
}
\end{table}

\subsection{Evaluation Details}\label{app:evaluation}

To comprehensively evaluate model capabilities, we employ a diverse set of benchmarks covering mathematical reasoning, multi-task reasoning, code generation, and instruction-following:  

\begin{enumerate}
    \item  \textbf{Mathematical reasoning:} We evaluate on AIME24, AIME25, and BeyondAIME~\citep{bytedance_seed_2025_beyondaime}.  
For AIME24 and AIME25, we report \texttt{pass@1} accuracy using 32 samples per problem (\texttt{avg@32}); for BeyondAIME, we report \texttt{avg@8}.  
    \item \textbf{Multi-task reasoning:} We evaluate on GPQA-Diamond~\citep{rein2024gpqa} and report \texttt{pass@1} with 8 samples per problem.  
    \item \textbf{Code generation:} We assess coding ability on LiveCodeBench~\citep{jain2024livecodebench} (2024-08–2025-01 subset, aligned with DeepSeek-R1~\citep{guo2025deepseekr1}), reporting \texttt{pass@1} with 8 samples per problem.  
    \item \textbf{Instruction-following:} We evaluate on IFEval~\citep{zhou2023ifeval} and report \texttt{pass@1} of the strict prompt accuracy with 4 samples per problem.  
\end{enumerate}

All evaluation codes are adapted from the DeepscaleR~\citep{deepscaler2025} codebase, where vLLM~\citep{kwon2023vllm} is leveraged to accelerate inference.
For IFEval, we use the same codes provided by the official paper~\citep{zhou2023ifeval}.

We provide our decoding parameters as follows:  
\begin{itemize}
    \item \textbf{Table~\ref{tab:main_result}}: We set the temperature to 0.6 and \texttt{topp} = 0.95. For LRMs trained from \texttt{DS-1.5B} and \texttt{DS-7B}, we use \texttt{topk} = $-1$ with a maximum sequence length of 32K tokens. For LRMs trained from \texttt{Qwen3-4B}, we use \texttt{topk} = $20$ with a maximum sequence length of 48K tokens, applying RoPE scaling as proposed by \citet{Polaris2025}.
    \item \textbf{Table~\ref{tab:sota}}: We set the temperature to 0.6 and \texttt{topp} = 0.95. For LRMs initialized from \texttt{DS-1.5B} and \texttt{DS-7B}, we use \texttt{topk} = $-1$ with a maximum sequence length of 32K tokens. For LRMs initialized from \texttt{Qwen3-4B}, we use \texttt{topk} = $20$ with a maximum sequence length of 48K tokens, again applying RoPE scaling as proposed by \citet{Polaris2025}. For models marked with ``*'', we report the results from their original publications (see Appendix~\ref{app:source}).  
    \item \textbf{Table~\ref{tab:results_tokens}}: For efficient reasoning baselines listed in Table~\ref{tabapp:baselines} and the thinking mode of initial models, we set \texttt{topp} = 0.95, \texttt{topk} = $-1$, and $T$ = 0.6, with a maximum sequence length of 32K tokens (48K for \texttt{Qwen3-4B}). For the thinking-free mode of initial models and our {\method}, we set \texttt{topp} = 0.8, \texttt{topk} = $20$, and $T$ = 0.7 with a maximum sequence length of 32K tokens, following \citet{yang2025qwen3}.
\end{itemize}

\subsection{Source of Some Results in Table~\ref{tab:sota}}\label{app:source}

Results for LRMs marked with ``*'' are taken directly from the Seed-1.5-Thinking report~\citep{seed2025seed1.6thinking} and the corresponding \href{https://huggingface.co/Qwen/Qwen3-235B-A22B-Thinking-2507}{Hugging Face page of Qwen3-2507}.
Note that the LiveCodeBench test set subsets of these results, and the metric of IFEval may differ from those in our experiments; their results are included for reference only.

\section{Analysis Details}\label{app:analysis}

As \textit{verification} is an important indicator of slow-thinking capabilities~\citep{setlur2025e3}, we conduct experiments in Section~\ref{subsec:reason} to examine how \textit{verification} can generalize to slow-thinking, even when trained with {\method} (thinking-free mode). Following \citet{yang2025pi}, for the experiments in Figure~\ref{fig:behaviour-analysis}, we segmented the reasoning trajectories using `\textbackslash n\textbackslash n' as delimiters and classified each step according to whether it contained verification-related phrases such as ``wait'', ``let me verify'', ``let me check'', ``checking'', ``verifying'', or ``double-check''.

\section{The Use of Large Language Models}

We only used LLMs (specifically GPT-5) for language refinement during the writing process of this work; all ideas, experimental designs, and analyses were conducted by the authors. The text polished by GPT-5 was subsequently reviewed and verified by us to ensure the quality of the content.

\section{More Results}

\subsection{\revise{Additional Results of Parameter-Level Analysis}}\label{app:para-analysis}

\begin{figure}
    \centering
    \includegraphics[width=0.5\linewidth]{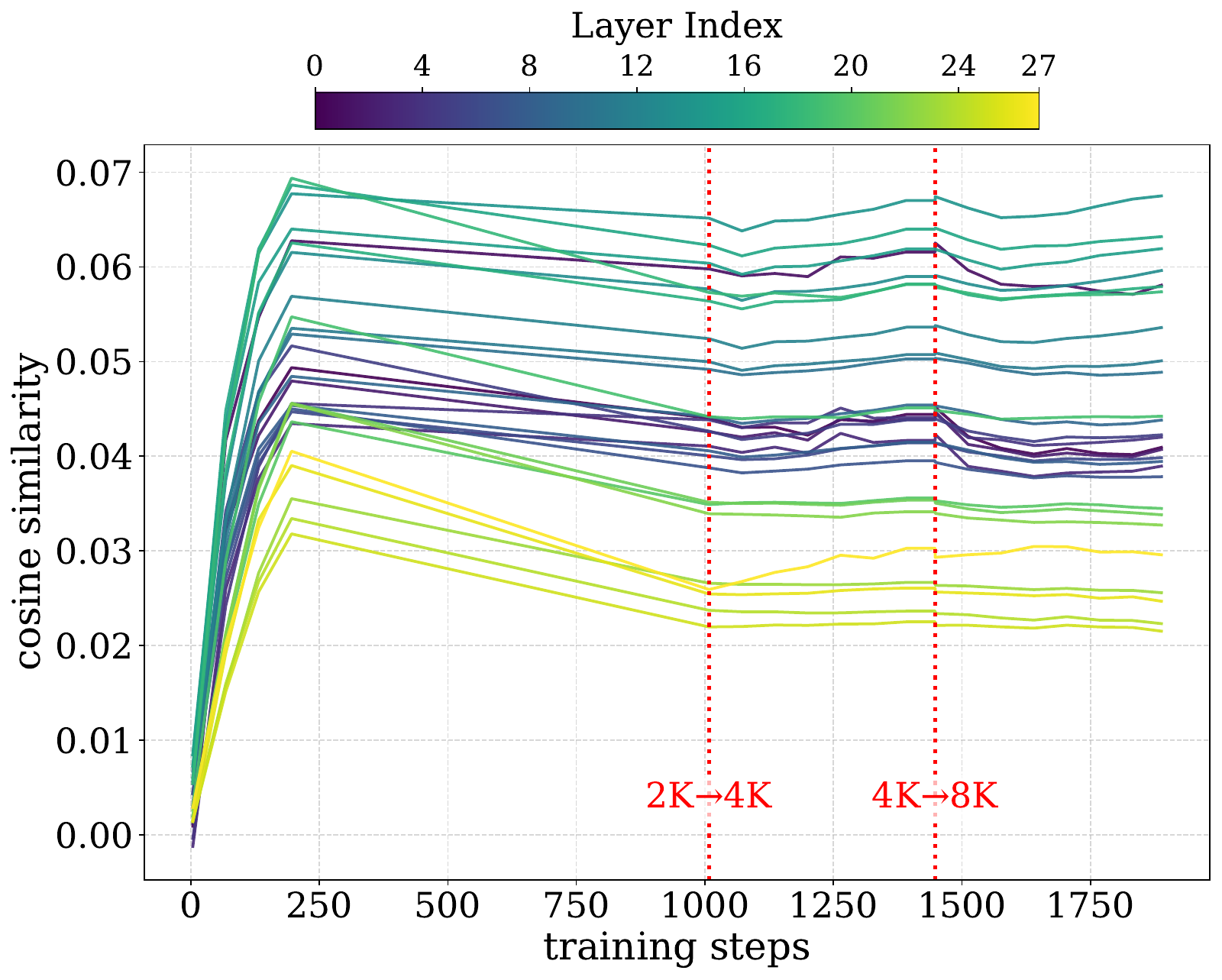}
    \caption{\revise{Cosine similarity between parameter updates of {\method}-trained checkpoints and an early ``Direct RL'' update of \texttt{DS-1.5B} across layers during training.}}
    \label{fig:cos-sim-align}
\end{figure}

\revise{
Figure~\ref{fig:parameter-analysis} (right) reports the cosine similarity of parameter-update vectors between {\method}-trained checkpoints and the final Direct RL checkpoint.
For completeness, Figure~\ref{fig:cos-sim-align} presents the cosine similarity between {\method}-trained checkpoints and an early Direct RL checkpoint (step 456) from the \texttt{DS-1.5B} run used in Table~\ref{tab:main_result}.
We observe stronger alignment with the final Direct RL checkpoint than with the early-stage checkpoint.
}

\subsection{Complete Results of Table~\ref{tab:sota} and Table~\ref{tab:results_tokens}}
\label{app:more_results}

Due to the page limit, we present only representative results in Sections~\ref{subsec:sota} and~\ref{subsec:token}, with the complete results provided in Tables~\ref{tabapp:sota} and~\ref{tabapp:results_tokens}, respectively.

\begin{table}[!t]
    \centering
    \caption{Results (\%) of RL after {\method} (``{\method}+RL'') vs. ``Direct RL'' across different benchmarks.  
``\texttt{Avg@k}'' denotes the average accuracy (\%) over $k$ random generations (i.e., \texttt{pass@1}).  
For LRMs marked with ``*'', results are taken from the corresponding reports (see Appendix~\ref{app:source}); all other results are from our own runs.  
All models are evaluated in thinking mode.  
The total training compute for ``{\method}+RL'' is matched to that of ``Direct RL'' for fair comparison.}
    \vspace{-1em}
    \label{tabapp:sota}
    \vspace{2mm}
    \resizebox{\textwidth}{!}{
    \begin{tabular}{l|ccc|c|c|c|c}
    \toprule[1.6pt]
        \multirow{3}{*}{\textbf{Models}} 
        & \multicolumn{3}{c|}{\textbf{Mathematics}} 
        & \textbf{Multi-Task} 
        & \textbf{Code} 
        & \textbf{Instruction} 
        & \textbf{Overall} \\ \cmidrule(lr){2-8}
        ~ & \textbf{AIME 24} 
           & \textbf{AIME 25} 
           & \textbf{Beyond AIME} 
           & \textbf{GPQA}
           & \textbf{LiveCode}
           & \textbf{IFEval}
           & \textbf{Overall} \\ 
        & \texttt{Avg@32} & \texttt{Avg@32} & \texttt{Avg@8} & \texttt{Avg@8} & \texttt{Avg@8} & \texttt{Avg@4} & \texttt{Avg.} \\ 
        \midrule
        DeepSeek R1$^*$ & 79.8 & 65.0 & 42.4 & 71.5 & 64.3 & 86.1 & 68.2 \\
        Seed-1.5-Thinking$^*$ & 86.7 & 74.0 & 48.0 & 77.3 & 64.9 & 87.4 & 73.0 \\
        Claude4 Opus Thinking$^*$ & - & 75.5 & - & 79.6 & 48.9 & 89.7 & - \\
        Qwen3-235B-Thinking$^*$ & 85.7 & 81.5 & - & 71.1 & 55.7 & 83.4 & -\\
        Qwen3-8B & 75.3 & 67.0 & 43.2 & 61.7 & 51.7 & 66.0 & 61.7 \\
        DS-R1-0528-Qwen3-8B & 82.8 & 74.6 & 50.6 & 61.4 & 59.0 & 73.9 & 67.0 \\
        \midrule
        \multicolumn{8}{c}{\texttt{\textbf{1.5B Size}}} \\
        \midrule
        DeepScaleR-1.5B & 37.8 & 31.6 & 13.1 & 19.1 & 21.9 & 40.5 & 27.3 \\
        \hline
        DS-1.5B Direct RL & 37.2 & 28.5 & 12.9 & 24.6 & 19.5 & 39.5 & 27.0 \\
        DS-1.5B {\method} stage 3 & 40.1 & 30.8 & 13.8 & 29.6 & 19.9 & 40.8 & 29.2 \\
        DS-1.5B {\method} + RL & 42.3 & 32.9 & 15.1 & 27.8 & 20.9 & 41.0 & 30.0 \\
        \midrule
        \multicolumn{8}{c}{\texttt{\textbf{4B Size}}} \\
        \midrule
        Polaris-4B-Preview & 73.2 & 70.7 & 39.9 & 54.9 & 39.7 & 63.9 & 57.0 \\
        \hline
        Qwen3-4B Direct RL & 78.8 & 71.5 & 46.4 & 56.2 & 54.3 & 65.1 & 62.0 \\
        Qwen3-4B {\method} stage 3 & 79.9 & 70.6 & 46.7 & 58.5 & 57.0 & 70.2 & 63.8 \\
        Qwen3-4B {\method} + RL & 80.8 & 76.0 & 49.5 & 61.1 & 55.7 & 71.3 & 65.7 \\
        \hline
        
        Qwen3-4B-2507-Thinking & 84.7 & 79.2 & 51.5 & 66.4 & 62.4 & 68.0 & 68.7 \\
        - {\method} only & 89.0 & 81.2 & 52.5 & 70.1 & 65.5 & 66.7 & 70.8 \\
        \midrule
        \multicolumn{8}{c}{\texttt{\textbf{7B Size}}} \\
        \midrule
        AReal-boba-RL-7B & 61.5 & 46.1 & 30.7 & 35.4 & 34.2 & 57.5 & 44.2 \\
        Skywork-OR1-7B-Preview & 61.2 & 46.4 & 31.2 & 35.2 & 43.3 & 55.4 & 45.5 \\
        Polaris-7B-Preview & 70.6 & 48.8 & 37.0 & 34.1 & 43.9 & 55.6 & 48.3 \\
        \hline
         DS-7B Direct RL & 62.3 & 47.9 & 30.1 & 36.9 & 42.8 & 57.1 & 46.2 \\
         DS-7B {\method} stage 3 & 62.0 & 44.6 & 31.1 & 46.8 & 42.1 & 60.2 & 47.8 \\
         DS-7B {\method} + RL & 65.3 & 47.1 & 33.2 & 45.9 & 43.3 & 57.3 & 48.7 \\
        \bottomrule[1.6pt]
    \end{tabular}}
    \vspace{-0.5em}
\end{table}

\begin{table}[!t]
    \centering
    \caption{Comparison of the Thinking-Free inference mode of {\method} with efficient reasoning baselines across various reasoning tasks.
``\texttt{Avg@k}'' denotes the average accuracy (in \%) over k generations (i.e., \texttt{pass@1}), and “\texttt{Toks}” indicates the average output length in thousands of tokens (K).
    }
    \vspace{-1em}
    \label{tabapp:results_tokens}
    \vspace{2mm}
    \resizebox{\textwidth}{!}{
    \begin{tabular}{l|cc|cc|cc|cc|cc|cc}
    \toprule[1.6pt]
        \multirow{3}{*}{\textbf{Models}} 
        
           & \multicolumn{2}{c|}{\textbf{AIME 24}} 
           & \multicolumn{2}{c|}{\textbf{AIME 25}} 
           & \multicolumn{2}{c|}{\textbf{Beyond AIME}} 
           & \multicolumn{2}{c|}{\textbf{GPQA}} 
           & \multicolumn{2}{c|}{\textbf{Overall}} \\ \cmidrule(lr){2-11}
        & \texttt{avg@32} & \texttt{Toks}
        & \texttt{avg@32} & \texttt{Toks}
        & \texttt{avg@8} & \texttt{Toks}
        & \texttt{avg@8} & \texttt{Toks}
        & \texttt{Avg.} & \texttt{Toks} \\ 
        \midrule
        \multicolumn{11}{c}{\texttt{\textbf{DeepSeek-Distill-Qwen-1.5B}}} \\
        \midrule
        TLMRE-DS-1.5B ($\alpha=0.1$) & 27.6 & 12.9 & 24.8 & 12.5 & 9.2 & 11.9 & 14.8 & 7.5 & 19.1 & 11.2 \\
        AdaptThink-1.5B ($\delta=0.05$) & 28.1 & 8.0 & 22.6 & 7.9 & 10.0 & 4.8 & 14.8 & 5.1 & 18.9 & 6.5 \\
        AutoThink-DS-1.5B-Stage3 & 30.3 & 10.2 & 25.2 & 9.1 & 9.4 & 8.7 & 17.4 & 7.0 & 20.6 & 8.7 \\
        Laser-DE-L4096-1.5B & 30.3 & 8.3 & 24.9 & 7.4 & 9.7 & 7.4 & 21.1 & 4.7 & 21.5 & 6.9 \\
        \hline
        Initial Model (Thinking) & 29.6 & 16.7 & 23.0 & 16.5 & 8.7 & 14.4 & 16.3 & 9.8 & 19.4 & 14.3 \\ 
        Initial Model (Thinking-Free) & 12.4 & 5.7 & 10.9 & 4.4 & 4.4 & 3.4 & 4.2 & 0.9 & 8.0 & 3.6 \\ 
        - {\method} stage 1 & 21.9 & 1.6 & 15.3 & 1.4 & 8.7 & 1.3 & 32.9 & 0.8 & 19.7 & 1.3 \\
        - {\method} stage 2 & 31.5 & 3.4 & 24.2 & 3.1 & 10.1 & 2.9 & 35.3 & 1.6 & 25.3 & 2.7 \\
        - {\method} stage 3 & 37.5 & 5.3 & 28.4 & 5.0 & 12.4 & 4.9 & 35.6 & 2.6 & 28.5 & 4.4 \\
        \midrule
        \multicolumn{11}{c}{\texttt{\textbf{DeepscaleR-1.5B-Preview}}} \\
        \midrule
        AutoThink-Stage3 & 38.9 & 8.7 & 28.9 & 7.7 & 11.6 & 7.8 & 27.3 & 5.4 & 26.7 & 7.4 \\
        L1-1.5B-Max & 27.2 & 3.2 & 26.3 & 2.9 & 9.1 & 3.1 & 32.4 & 2.3 & 23.8 & 2.9 \\
        Thinkless-1.5B-RL & 28.4 & 11.3 & 24.1 & 11.1 & 8.1 & 11.7 & 20.3 & 12.7 & 20.2 & 11.7 \\
        \midrule
        \multicolumn{11}{c}{\texttt{\textbf{Qwen3-4B}}} \\
        \midrule
        Initial Model (Thinking) & 73.6 & 18.0 & 68.3 & 22.3 & 43.4 & 23.5 & 56.8 & 10.7 & 60.5 & 18.6 \\ 
        Initial Model (Thinking-Free) & 26.9 & 7.2 & 20.2 & 5.5 & 11.0 & 4.2 & 45.0 & 2.3 & 25.8 & 4.8 \\ 
        - {\method} stage 1 & 43.9 & 3.7 & 31.1 & 3.4 & 23.2 & 3.4 & 47.8 & 1.5 & 36.5 & 3.0 \\
        - {\method} stage 2 & 63.7 & 7.2 & 52.8 & 8.0 & 33.0 & 7.0 & 50.9 & 2.3 & 50.1 & 6.1 \\
        - {\method} stage 3 & 75.1 & 10.5 & 66.9 & 12.7 & 42.1 & 12.7 & 55.9 & 3.7 & 60.0 & 9.9\\
        Qwen3-4B-Instruct-2507 & 60.4 & 7.9 & 46.1 & 7.4 & 33.0 & 7.3 & 64.4 & 4.8 & 51.0 & 6.9 \\
        \bottomrule[1.6pt]
    \end{tabular}}
    \vspace{-0.5em}
\end{table}

\subsection{\revise{Thinking vs. Thinking-Free Inference}}\label{app:thinking_vs_thinkfree}

\revise{
To compare thinking and thinking-free modes, Table~\ref{tabapp:think_vs_thinkfree} reports accuracy and average output token counts across training stages of {\method} for \texttt{Qwen3-4B}.
The accuracy gap between the two modes narrows from Stage~1 to Stage~3. 
In Stage~3, the thinking mode achieves an average accuracy of 63.9\%, whereas the thinking-free mode achieves 60.0\%. Meanwhile, the thinking-free mode produces substantially fewer tokens on average (9.9k vs.\ 15.3k). 
One may choose the thinking-free mode to trade a modest decrease in accuracy for improved efficiency, or switch between modes to balance accuracy and cost.
}

\begin{table}[!t]
    \centering
    \caption{\revise{Comparison of the Thinking-Free inference mode with the Thinking mode across various reasoning tasks after {\method}.
``\texttt{Avg@k}'' denotes the average accuracy (in \%) over k generations (i.e., \texttt{pass@1}), and “\texttt{Toks}” indicates the average output length in thousands of tokens (K).}
    }
    \vspace{-1em}
    \label{tabapp:think_vs_thinkfree}
    \vspace{2mm}
    \resizebox{\textwidth}{!}{
    \begin{tabular}{l|cc|cc|cc|cc|cc|cc}
    \toprule[1.6pt]
        \multirow{3}{*}{\textbf{Models}} 
        
           & \multicolumn{2}{c|}{\textbf{AIME 24}} 
           & \multicolumn{2}{c|}{\textbf{AIME 25}} 
           & \multicolumn{2}{c|}{\textbf{Beyond AIME}} 
           & \multicolumn{2}{c|}{\textbf{GPQA}} 
           & \multicolumn{2}{c|}{\textbf{Overall}} \\ \cmidrule(lr){2-11}
        & \texttt{avg@32} & \texttt{Toks}
        & \texttt{avg@32} & \texttt{Toks}
        & \texttt{avg@8} & \texttt{Toks}
        & \texttt{avg@8} & \texttt{Toks}
        & \texttt{Avg.} & \texttt{Toks} \\ 
        \midrule
        \multicolumn{11}{c}{\texttt{\textbf{Qwen3-4B}}} \\
        \midrule
        Initial Model (Thinking) & 73.6 & 18.0 & 68.3 & 22.3 & 43.4 & 23.5 & 56.8 & 10.7 & 60.5 & 18.6 \\ 
        Initial Model (Thinking-Free) & 26.9 & 7.2 & 20.2 & 5.5 & 11.0 & 4.2 & 45.0 & 2.3 & 25.8 & 4.8 \\ 
        \hline
        - {\method} stage 1 (Thinking) & 75.2 & 13.8 & 67.8 & 17.1 & 42.4 & 16.8 & 57.9 & 7.5 & 60.8 & 13.8 \\
        - {\method} stage 1 (Thinking-Free) & 43.9 & 3.7 & 31.1 & 3.4 & 23.2 & 3.4 & 47.8 & 1.5 & 36.5 & 3.0 \\
        \hline
        - {\method} stage 2 (Thinking) & 76.0 & 13.4 & 68.2 & 17.0 & 44.7 & 16.7 & 57.8 & 7.7 & 61.7 & 13.7 \\
        - {\method} stage 2 & 63.7 & 7.2 & 52.8 & 8.0 & 33.0 & 7.0 & 50.9 & 2.3 & 50.1 & 6.1 \\
        \hline
        - {\method} stage 3 (Thinking) & 79.9 & 15.1 & 70.6 & 18.7 & 46.7 & 18.8 & 58.5 & 8.6 & 63.9 & 15.3 \\
        - {\method} stage 3 & 75.1 & 10.5 & 66.9 & 12.7 & 42.1 & 12.7 & 55.9 & 3.7 & 60.0 & 9.9\\
        \hline
        Qwen3-4B-Instruct-2507 & 60.4 & 7.9 & 46.1 & 7.4 & 33.0 & 7.3 & 64.4 & 4.8 & 51.0 & 6.9 \\
        \bottomrule[1.6pt]
    \end{tabular}}
    \vspace{-0.5em}
\end{table}

\subsection{\revise{About Decoding Parameters}}\label{app:decoding_para}

\revise{
The decoding parameters used in our main-text experiments are detailed in Appendices~\ref{app:meta-exp} and~\ref{app:evaluation}. 
Following the official Hugging Face recommendations for each model, we adopt different settings for the thinking and thinking-free modes. 
To assess robustness, we additionally report results under a unified configuration: temperature = 0.6, top-p = 0.95, top-k = 20, and a maximum output length of 32k tokens. 
As shown in Tables~\ref{tabapp:meta-exp}, \ref{tabapp:main_result_v1}, \ref{tabapp:sota_v1}, and \ref{tabapp:results_tokens_v1}, all of our conclusions remain unchanged under this unified setting.
}

\revise{
Polaris~\citep{Polaris2025} reports that longer-context inference with RoPE scaling improves performance on harder questions and evaluates at 90K tokens.
We find that 90K is prohibitively slow, and 48K suffices based on the clip ratio analysis (Table~\ref{tabapp:clip_ratio}). 
As shown in Table~\ref{tabapp:sota_v1}, constraining the maximum output length to 32K slightly degrades the performance of the thinking mode for models RL-tuned from \texttt{Qwen3-4B}. 
We therefore recommend a 48K context length for \texttt{Qwen3-4B}-based RL-tuned models.
}

\begin{table}
    \centering
    \begin{tabular}{c|cccccc}
    \toprule
       Response Length  & 16K & 32K & 48K & 64K & 80K & 90K \\
       \midrule
        \texttt{Avg@32} & 42.6 & 64.1 & 68.3 & 68.5 & 68.5 & 68.5 \\
        Clip Ratio & 55.7 & 25.7 & 5.2 & 1.2 & 0.6 & 0.0 \\
    \bottomrule
    \end{tabular}
    \caption{AIME25 results for \texttt{Qwen3-4B}: \texttt{Avg@32} and clip ratio across maximum output lengths.}
    \label{tabapp:clip_ratio}
\end{table}

\begin{table}
    \centering
    \caption{\revise{Re-evaluate the meta-experiment in Figure~\ref{fig:meta-exp} (Left) under a unified evaluation setting (temperature = 0.6, top-p = 0.95, top-k = 20). ``\# Tokens'' denotes the average number of tokens on AIME25 and ``$\Delta$'' is the percentage of reduction in token consumption.}}
    \label{tabapp:meta-exp}
    \begin{tabular}{l|ccc}
        \toprule
        Model & \# Tokens (Thinking) & \# Tokens (Thinking-Free) & $\Delta$ \\
        \midrule
       \texttt{DS-1.5B} (original setting)  & 16.5K & 4.4K & -73\% \\
       Unified Setting & 16.4K & 3.6K & -78\% \\
       \midrule
       \texttt{Qwen3-4B} (original setting)  & 17.4K & 5.1K & -71\% \\
       Unified Setting & 17.4K & 4.7K & -73\% \\
       \bottomrule
    \end{tabular}
\end{table}

\begin{table}[!t]
    \centering
    \caption{\revise{Results of {\method} vs. direct RL across different benchmarks.  
``\texttt{Avg@k}'' denotes the average accuracy (\%) over $k$ random generations (i.e., \texttt{pass@1}).  
All models are evaluated in thinking mode.  
The total training compute for the 3 stages of {\method} equals that of ``Direct RL'' for fair comparison.
This table corresponds to Table~\ref{tab:main_result} under a unified decoding setting.}}
\vspace{-1em}
    \label{tabapp:main_result_v1}
    \vspace{2mm}
    \resizebox{\textwidth}{!}{
    \begin{tabular}{l|ccc|c|c|c|c}
    \toprule[1.6pt]
        \multirow{3}{*}{\textbf{Models}} 
        & \multicolumn{3}{c|}{\textbf{Mathematics}} 
        & \textbf{Multi-Task} 
        & \textbf{Code} 
        & \textbf{Instruction} 
        & \textbf{Overall} \\ \cmidrule(lr){2-8}
        ~ & \textbf{AIME 24} 
           & \textbf{AIME 25} 
           & \textbf{Beyond AIME} 
           & \textbf{GPQA}
           & \textbf{LiveCode}
           & \textbf{IFEval}
           & \textbf{Overall} \\ 
        & \texttt{Avg@32} & \texttt{Avg@32} & \texttt{Avg@8} & \texttt{Avg@8} & \texttt{Avg@8} & \texttt{Avg@4} & \texttt{Avg.} \\ 
        \midrule
        \multicolumn{8}{c}{\texttt{\textbf{{DeepSeek-Distill-Qwen-1.5B}}}} \\
        \midrule
        Initial Model & 29.1 & 25.5 & 9.0 & 16.3 & 17.6 & 40.1 & 22.9  \\ 
- Direct RL & 35.5 & 27.9 & 12.5 & 20.2 & 18.7 & 38.1 & 25.5 \\
- {\method} stage 1 & 31.8 & 26.7 & 12.5 & 30 & 18.9 & 39.5 & 26.6 \\
- {\method} stage 2 & 35.1 & 27.6 & 13.4 & 29.7 & 18.9 & 44.0 & 28.1  \\
- {\method} stage 3 & 36.1 & 30.1 & 14.9 & 28.5 & 21.0 & 41.8 & 28.7 \\
        \midrule
        \multicolumn{8}{c}{\texttt{\textbf{{Qwen3-4B}}}} \\
        \midrule
        Initial Model  & 71.6 & 64.1 & 41.5 & 56.6 & 54.9 & 64.9 & 58.9 \\
Direct RL  & 73.4 & 62.2 & 41.1 & 56.2 & 52.1 & 66.0 & 58.5 \\
{\method} stage 1 & 74.7 & 67.4 & 42.1 & 57.9 & 55.3 & 66.0 & 60.6 \\
{\method} stage 2 & 75.6 & 67.5 & 44.6 & 57.8 & 54.6 & 64.8 & 60.8 \\
{\method} stage 3 & 78.7 & 69.5 & 46.1 & 58.5 & 56.8 & 70.2 & 63.3 \\
        \midrule
        \multicolumn{8}{c}{\texttt{\textbf{{DeepSeek-Distill-Qwen-7B}}}} \\
        \midrule
        Initial Model & 53.3 & 40.5 & 25.0 & 36.8 & 37.6 & 55.6 & 41.5 \\ 
- Direct RL &  56.8 & 41.6 & 26.7 & 37.4 & 39.6 & 58.2 & 43.4 \\
- {\method} stage 1 & 58.8 & 39.6 & 28.8 & 49.4 & 40.1 & 56.6 & 45.5 \\
- {\method} stage 2 & 61.7 & 42.9 & 31.5 & 47.5 & 42.0 & 57.5 & 47.2 \\
- {\method} stage 3 & 62.6 & 43.2 & 30.7 & 47.1 & 42.4 & 59.3 & 47.5 \\
        \bottomrule[1.6pt]
    \end{tabular}}
\vspace{-0.5em}
\end{table}

\begin{table}[!t]
    \centering
    \caption{\revise{Results (\%) of RL after {\method} (``{\method}+RL'') vs. ``Direct RL'' across different benchmarks.  
``\texttt{Avg@k}'' denotes the average accuracy (\%) over $k$ random generations (i.e., \texttt{pass@1}).  
For LRMs marked with ``*'', results are taken from the corresponding reports (see Appendix~\ref{app:source}); 
All models are evaluated in thinking mode.  
The total training compute for ``{\method}+RL'' is matched to that of ``Direct RL'' for fair comparison.
This table corresponds to Table~\ref{tab:sota} under a unified decoding setting.}}
    \vspace{-1em}
    \label{tabapp:sota_v1}
    \vspace{2mm}
    \resizebox{\textwidth}{!}{
    \begin{tabular}{l|ccc|c|c|c|c}
    \toprule[1.6pt]
        \multirow{3}{*}{\textbf{Models}} 
        & \multicolumn{3}{c|}{\textbf{Mathematics}} 
        & \textbf{Multi-Task} 
        & \textbf{Code} 
        & \textbf{Instruction} 
        & \textbf{Overall} \\ \cmidrule(lr){2-8}
        ~ & \textbf{AIME 24} 
           & \textbf{AIME 25} 
           & \textbf{Beyond AIME} 
           & \textbf{GPQA}
           & \textbf{LiveCode}
           & \textbf{IFEval}
           & \textbf{Overall} \\ 
        & \texttt{Avg@32} & \texttt{Avg@32} & \texttt{Avg@8} & \texttt{Avg@8} & \texttt{Avg@8} & \texttt{Avg@4} & \texttt{Avg.} \\ 
        \midrule
        DeepSeek R1$^*$ 
& {79.8} 
& {65.0} 
& {42.4} 
& {71.5} 
& {64.3} 
& {86.1} 
& {68.2} \\

Seed-1.5-Thinking$^*$ 
& {86.7} 
& {74.0} 
& {48.0} 
& {77.3} 
& {64.9} 
& {87.4} 
& {73.0} \\

Claude4 Opus Thinking$^*$ 
& - 
& {75.5} 
& - 
& {79.6} 
& {48.9} 
& {89.7} 
& - \\

Qwen3-235B-Thinking$^*$ 
& {85.7} 
& {81.5} 
& - 
& {71.1} 
& {55.7} 
& {83.4} 
& - \\

\midrule
Qwen3-4B Direct RL 
& 75.8 & 65.8 & 42.1 & 56.0 & 53.9 & 65.1 & 59.8 \\

Qwen3-4B {\method} stage 3 
& 78.7 & 69.5 & 46.1 & 58.5 & 56.8 & 70.2 & 63.3 \\

Qwen3-4B {\method} + RL 
& 77.9 & 70.8 & 47.5 & 60.8 & 54.5 & 71.3 & 63.8 \\

        \bottomrule[1.6pt]
    \end{tabular}}
    \vspace{-0.5em}
\end{table}

\begin{table}[!t]
    \centering
    \caption{\revise{Comparison of the Thinking-Free inference mode of {\method} with efficient reasoning baselines across various reasoning tasks.
``\texttt{Avg@k}'' denotes the average accuracy (in \%) over k generations (i.e., \texttt{pass@1}), and “\texttt{Toks}” indicates the average output length in thousands of tokens (K).
Models with ``*'' are trained from \texttt{DeepScaleR-1.5B}, while the remaining are from \texttt{DS-1.5B}.
This table corresponds to Table~\ref{tab:results_tokens} under a unified decoding setting.}
    }
    \vspace{-1em}
    \label{tabapp:results_tokens_v1}
    \vspace{2mm}
    \resizebox{\textwidth}{!}{
    \begin{tabular}{l|cc|cc|cc|cc|cc|cc}
    \toprule[1.6pt]
        \multirow{3}{*}{\textbf{Models}} 
        
           & \multicolumn{2}{c|}{\textbf{AIME 24}} 
           & \multicolumn{2}{c|}{\textbf{AIME 25}} 
           & \multicolumn{2}{c|}{\textbf{Beyond AIME}} 
           & \multicolumn{2}{c|}{\textbf{GPQA}} 
           & \multicolumn{2}{c|}{\textbf{Overall}} \\ \cmidrule(lr){2-11}
        & \texttt{avg@32} & \texttt{Toks}
        & \texttt{avg@32} & \texttt{Toks}
        & \texttt{avg@8} & \texttt{Toks}
        & \texttt{avg@8} & \texttt{Toks}
        & \texttt{Avg.} & \texttt{Toks} \\ 
        \midrule
        1) TLMRE-DS-1.5B ($\alpha=0.1$) & 27.6 & 12.9 & {24.8} & 12.5 & 9.2 & 11.9 & 14.8 & 7.5 & 19.1 & 11.2 \\
        2) AdaptThink-1.5B ($\delta=0.05$) & 28.1 & {8.0} & 22.6 & 7.9 & {10.0} & {4.8} & 14.8 & 5.1 & 18.9 & {6.5} \\
        3) AutoThink-DS-1.5B-Stage3 & {30.3} & 10.2 & {25.2} & 9.1 & {9.4} & 8.7 & 17.4 & 7.0 & {20.6} & 8.7 \\
        4) Laser-DE-L4096-1.5B & {30.3} & 8.3 & {24.9} & {7.4} & {9.7} & 7.4 & {21.1} & {4.7} & {21.5} & 6.9 \\
        \hline
        5) AutoThink-Stage3$^*$ & {38.9} & 8.7 & {28.9} & 7.7 & {11.6} & 7.8 & {27.3} & 5.4 & {26.7} & 7.4 \\
        6) L1-1.5B-Max$^*$ & 27.2 & {3.2} & {26.3} & {2.9} & 9.1 & {3.1} & {32.4} & {2.3} & {23.8} & {2.9} \\
        7) Thinkless-1.5B-RL$^*$ & 28.4 & 11.3 & 24.1 & 11.1 & 8.1 & 11.7 & 20.3 & 12.7 & 20.2 & 11.7 \\
        \hline
        DS-1.5B (Thinking) & 29.1 & 16.5 & 25.5 & 16.4 & 9.0 & 17.3 & 16.3 & 9.8 & 20.0 & 15.0 \\ 
        DS-1.5B (Thinking-Free) & 13.3 & 5.1 & 11.0 & 3.6 & 4.0 & 2.6 & 6.2 & 0.9 & 8.6 & 3.0 \\ 
        - {\method} stage 1 & 23.9 & 1.5 & 16.2 & 1.4 & 9.1 & 1.2 & 36.4 & 0.8 & 21.4 & 1.2 \\
        - {\method} stage 2 & 28.6 & 3.4 & 25.0 & 3.0 & 10.6 & 2.9 & 35.7 & 1.6 & 25.0 & 2.7 \\
        - {\method} stage 3 & 37.7 & 5.2 & 28.4 & 5.1 & 14.5 & 4.9 & 34.1 & 2.6 & 28.7 & 4.4  \\
        \bottomrule[1.6pt]
    \end{tabular}}
    \vspace{-0.5em}
\end{table}

\subsection{\revise{About Training Temperature}}\label{app:training_temp}

\begin{figure}
    \centering
    \includegraphics[width=0.45\linewidth]{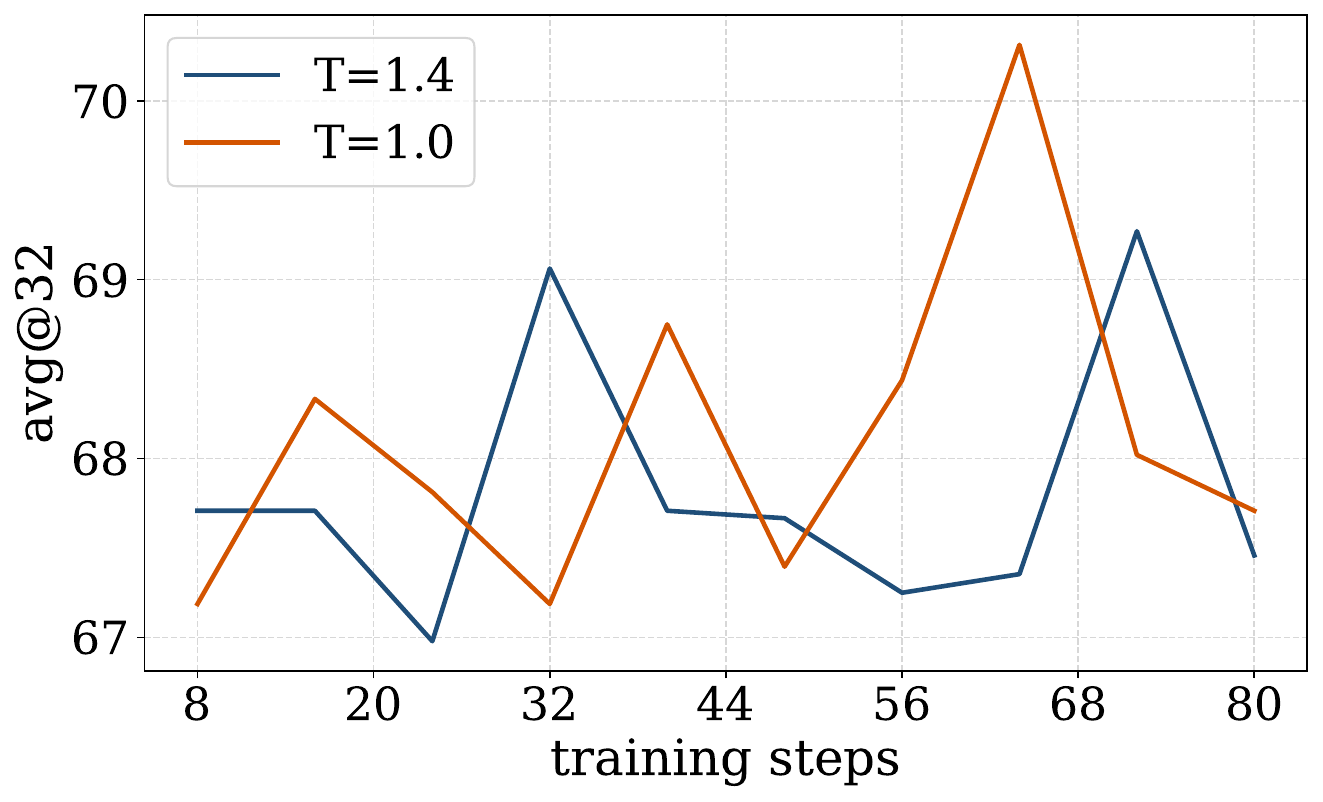}
    \includegraphics[width=0.45\linewidth]{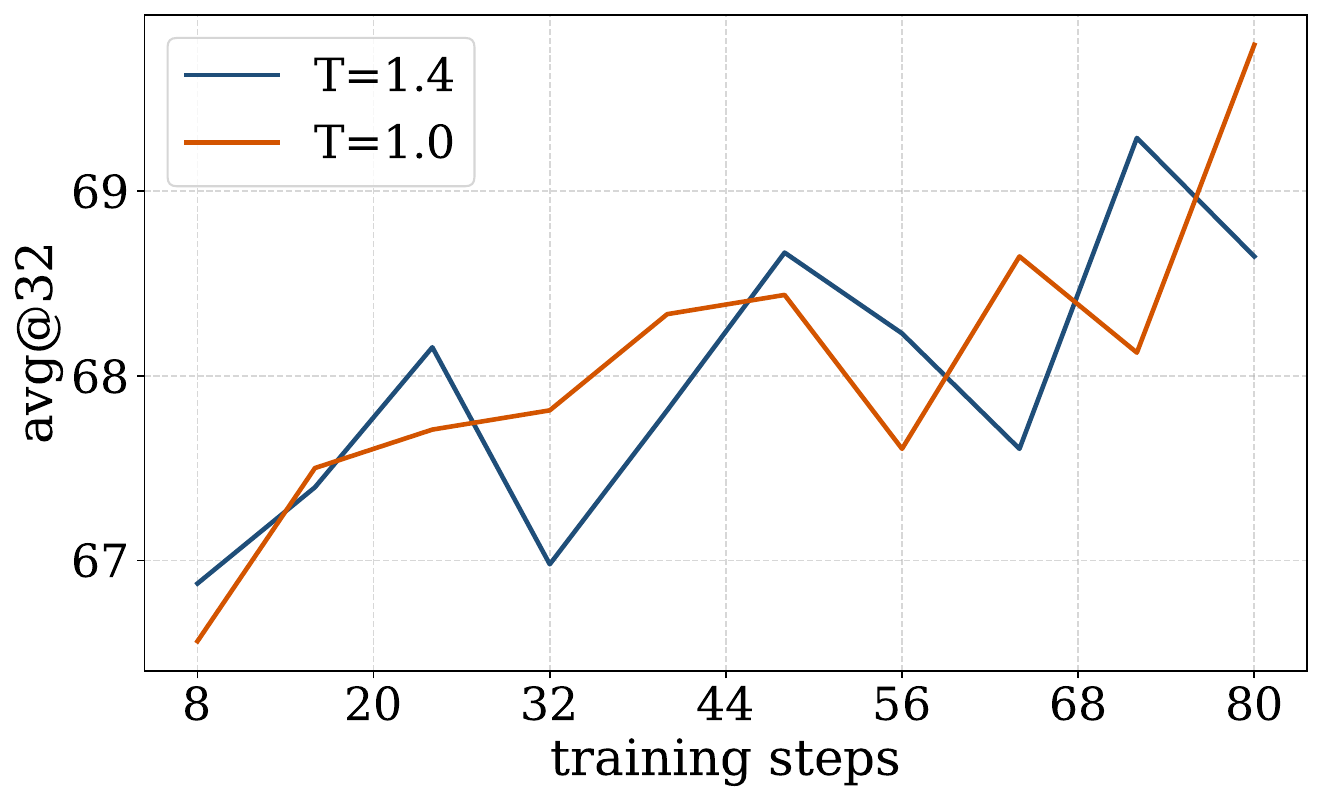}
    \caption{\revise{\texttt{AVG@32} (\%) on AIME25 for \texttt{Qwen3-4B} across training steps under different training temperatures (1.4 vs. 1.0). Left: Stage 1 of {\method}; Right: Direct RL.}}
    \label{fig:temp_ablation}
\end{figure}

\revise{
In our main experiments, we set the training temperature to 1.0 (Appendix~\ref{app:training}). 
Motivated by Polaris\citep{Polaris2025}, which reports an ``optimal'' temperature of 1.4 for \texttt{Qwen3-4B}, we conduct an ablation on training temperature using \texttt{Qwen3-4B}. 
As shown in Figure\ref{fig:temp_ablation}, performance at 1.4 is comparable to 1.0 for both {\method} Stage 1 and Direct RL (measured by \texttt{AVG@32} on AIME25), suggesting that our results are robust to this choice of training temperature.
}

\subsection{\revise{About the Stage Boundary and Stage Length}}\label{app:stage_boundary}

\begin{figure}
    \centering
    \includegraphics[width=0.45\linewidth]{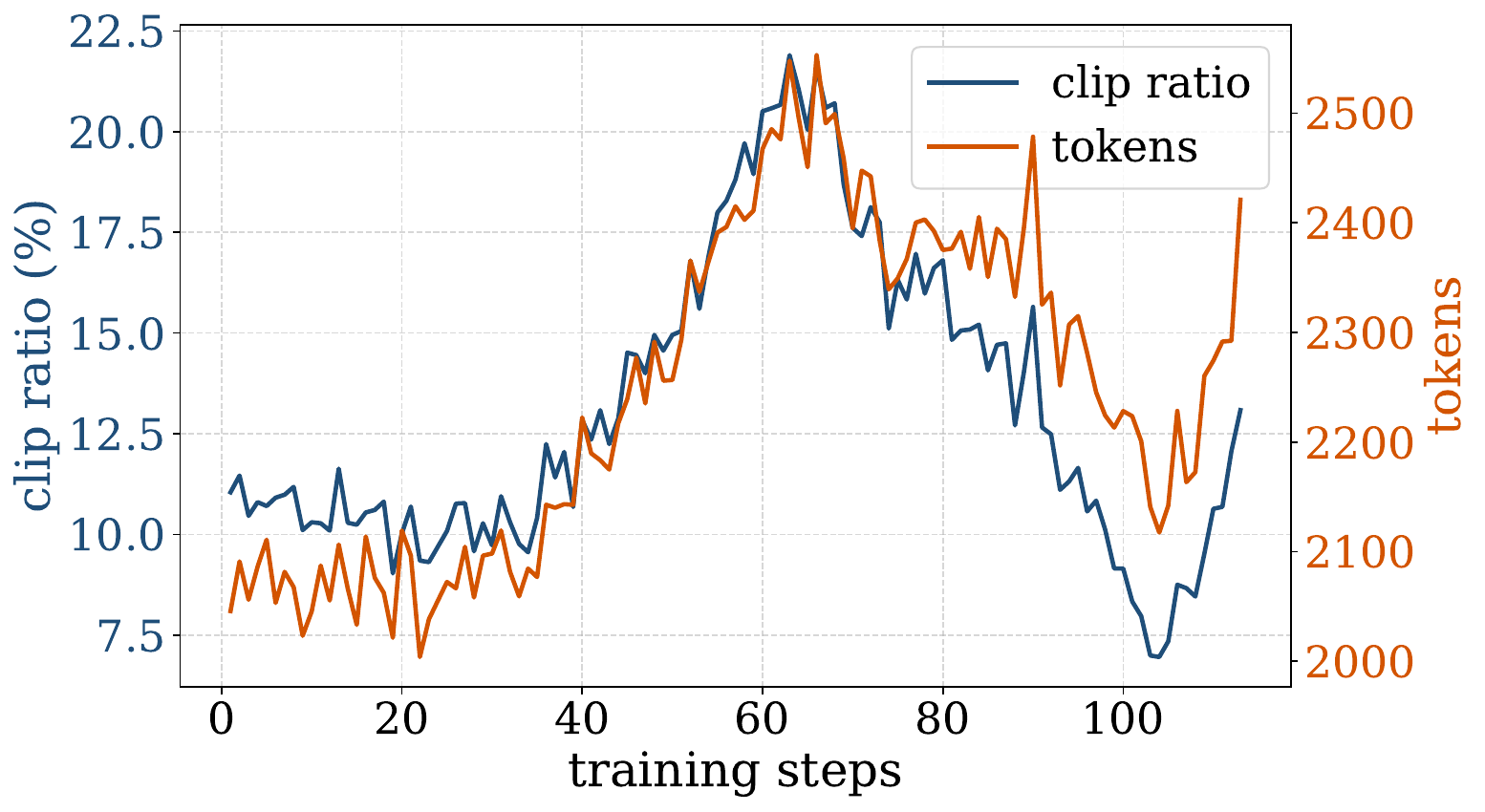}
    \includegraphics[width=0.38\linewidth]{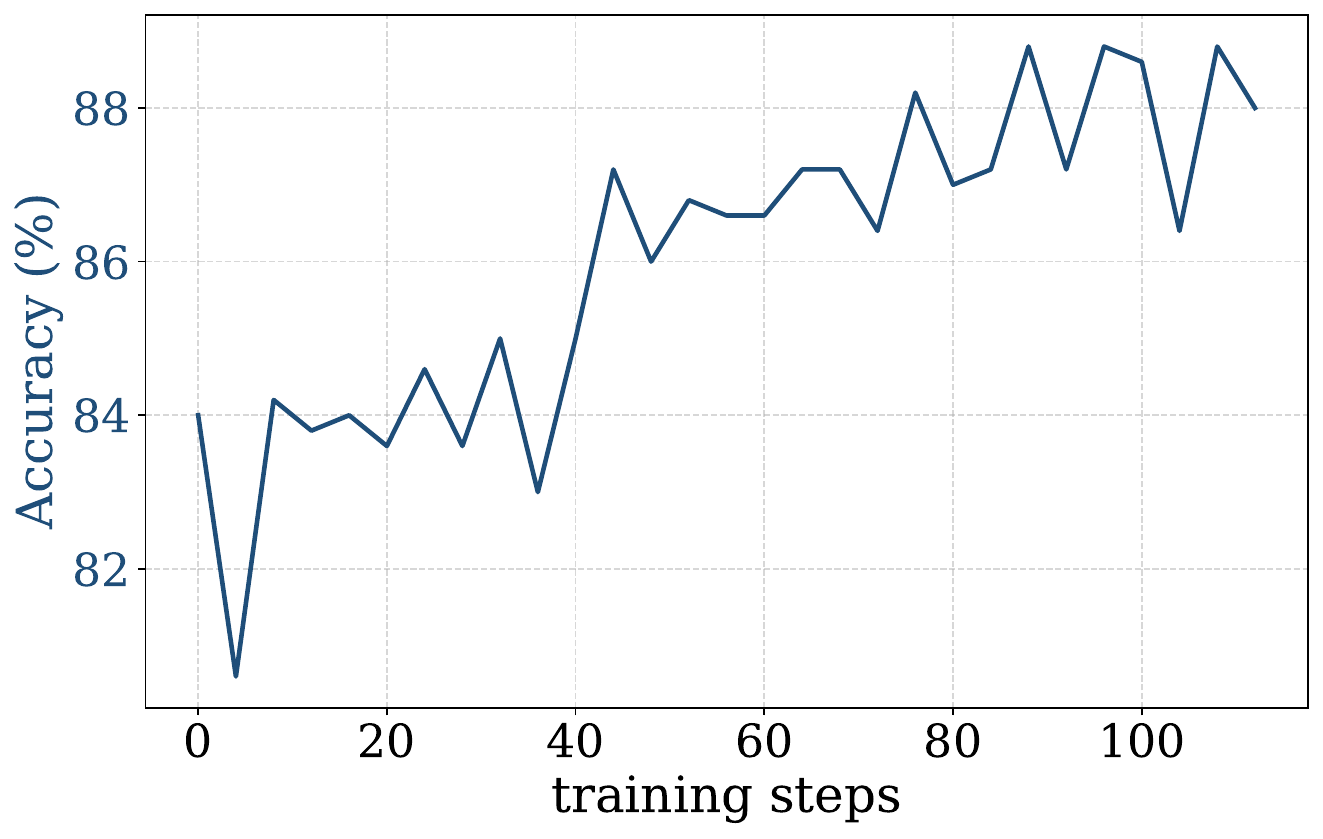}
    \caption{\revise{\texttt{Qwen3-4B} {\method} Stage 1. Left: clip ratio and average output tokens over training steps; Right: MATH500 validation accuracy. Step 100 marks a turning point.}}
    \label{fig:stage_boundary}
\end{figure}

\revise{
Following prior multi-stage training practice~\citep{Polaris2025,deepscaler2025}, we increase the training context length when the clip ratio shows a renewed inflection and validation performance begins to plateau.
Figure~\ref{fig:stage_boundary} illustrates this behavior for {\method} Stage 1 on \texttt{Qwen3-4B}; step 100 is a natural transition point.
}

\revise{
Training context length trades off training throughput and effective learning signal: longer contexts slow training, whereas shorter contexts constrain learning on harder questions. 
We therefore begin with a shorter length and expand as training stabilizes. 
To set stage lengths, we prioritize effective signal: with 8-sample rollouts, 0/8 or 8/8 correctness contributes no learning signal~\citep{yu2025dapo}, so for a candidate length L, we compute the share of questions with 1/8–7/8 correctness as a proxy for effective training mass. 
We choose L to preserve throughput while maintaining substantial effective mass at each stage. For \texttt{Qwen3-4B}, L=4K yields 28K/53K questions in the 1/8–7/8 band, whereas L=8K yields 29K/53K; thus we start at 4K for efficiency and move to 8K once validation gains plateau.
Note that the effective sets across stages are not disjoint; for instance, the same question may score 3/8 at 4K and 6/8 at 8K.
In all experiments, we train on the full Polaris-53K dataset to ensure fairness. However, for better efficiency, one can curate stage-specific subsets based on the effective-signal criterion by filtering out questions that are 0/8 or 8/8 correct under the current context length.
}

\subsection{\revise{Multi-Stage Direct RL ($4\rm{K}\rightarrow8\rm{K}\rightarrow16\rm{K}$)}}\label{app:sb}

\revise{Figure~\ref{fig:meta-exp} (right) shows that applying Direct RL with a 4K response length causes a substantial performance drop on AIME25 for \texttt{Qwen3-4B}. 
In contrast, {\method} trains stably even at 4K, suggesting that—even in a multi-stage setting—the initial stage length should not be too short. 
We posit that enabling multi-stage training with short response lengths is a key benefit of {\method}. 
To rule out the possibility that this effect is an artifact of staging, we run Direct RL with the identical multi-stage schedule on \texttt{Qwen3-4B}. 
The complete results are reported in Table~\ref{tabapp:uwvx}, with training details provided in Table~\ref{tabapp:uwvx_train}.}

\begin{table}[!t]
    \centering
    \caption{\revise{{\method} v.s. Multi-Stage Direct RL.
All models are evaluated in thinking mode.}}
\vspace{-1em}
    \label{tabapp:uwvx}
    \vspace{2mm}
    \resizebox{\textwidth}{!}{
    \begin{tabular}{l|ccc|c|c|c|c}
    \toprule[1.6pt]
        \multirow{3}{*}{\textbf{Models}} 
        & \multicolumn{3}{c|}{\textbf{Mathematics}} 
        & \textbf{Multi-Task} 
        & \textbf{Code} 
        & \textbf{Instruction} 
        & \textbf{Overall} \\ \cmidrule(lr){2-8}
        ~ & \textbf{AIME 24} 
           & \textbf{AIME 25} 
           & \textbf{Beyond AIME} 
           & \textbf{GPQA}
           & \textbf{LiveCode}
           & \textbf{IFEval}
           & \textbf{Overall} \\ 
        & \texttt{Avg@32} & \texttt{Avg@32} & \texttt{Avg@8} & \texttt{Avg@8} & \texttt{Avg@8} & \texttt{Avg@4} & \texttt{Avg.} \\ 
        \midrule
        Initial Model     & {73.6} & {68.3} & {43.4} & {56.8} & {54.9} & {64.9} & {60.3} \\
        \midrule
{\method} stage 1 & {75.2} & {67.8} & {42.4} & {57.9} & {55.3} & {66.0} & {60.8} \\
Multi-Stage Direct RL Stage 1 & 49.2 & 36.0 & 22.6 & 51.4 & 44.0 & 64.5 & 44.6 \\
\midrule
{\method} stage 2 & {76.0} & {68.2} & {44.7} & {57.8} & {54.8} & {64.8} & {61.0} \\
Multi-Stage Direct RL Stage 2 & 58.4 & 42.3 & 28.2 & 52.9 & 48.4 & 63.2 & 48.9 \\
\midrule
{\method} stage 3 & {79.9} & {70.6} & {46.7} & {58.5} & {57.0} & {70.2} & {63.8} \\
Multi-Stage Direct RL Stage 3 & 64.9 & 48.6 & 32.5 & 55.2 & 51.0 & 65.2 & 52.9 \\
        \bottomrule[1.6pt]
    \end{tabular}}
\vspace{-0.5em}
\end{table}

\begin{table}
    \centering
    \caption{\revise{Training Steps, Time, and MFU of {\method} v.s. Direct RL with {\texttt{Qwen3-4B}}.}}
    \label{tabapp:uwvx_train}
    {%
        \begin{tabular}{l|cccccc}
            \toprule
              & \texttt{Training Steps}  & Training Time (kh) & MFU (\%) \\
             \midrule
            \method{} Stage 1 & 4K, 100 steps & 0.35 & 66.28 \\
            Multi-Stage Direct RL Stage 1 & 4K, 100 steps & 0.45 & 66.82 \\
            \midrule
            \method{} Stage 2 & 8K, 56 steps & 0.52 & 66.67 \\
            Multi-Stage Direct RL Stage 2 & 8K, 56 steps & 0.51 & 66.31 \\
            \midrule
            \method{} Stage 3 & 16K, 64 steps & 0.91 & 66.76 \\
            Multi-Stage Direct RL Stage 3 & 16K, 64 steps & 0.97 & 65.99\\
            \midrule
            \method{} Total & -  & 1.79 & - \\
            Multi-Stage Direct RL Total & - & 1.93 & - \\
             \bottomrule
        \end{tabular}
    }
\end{table}

\section{Reproducibility Statement}

To ensure reproducibility, we provide a detailed description of our method in Section~\ref{subsec:tpfi definition} and present all experimental setups in Section~\ref{subsec:exp_setting} and Appendix~\ref{app:exp-setup}, including training configurations and evaluation protocols.
All evaluation datasets, the training set, the training framework, testing codes, as well as the efficient reasoning baselines used in our experiments, are publicly available and properly referenced (see Appendix~\ref{app:exp-setup}). 
We will also release our codes and training checkpoints upon acceptance for reproducibility.

\end{document}